%% file: total.tex
\documentclass{LMCS}

\usepackage{pstricks,pst-plot,pst-node}
\usepackage{enumerate,graphicx,hyperref}

\theoremstyle{plain}\newtheorem{hypothesis}{Hypothesis}
\theoremstyle{plain}\newtheorem{proposition}{Proposition}
\theoremstyle{plain}\newtheorem{lemma}{Lemma}

\newcommand{\widthfig}[4]{
\begin{figure}[htb]
\begin{center}
  \includegraphics[width=#1cm]{#2}
\end{center}
\vspace*{-0.3cm}
\caption{\label{#2}#3}
\vspace*{#4}
\end{figure}}

\newcommand{\cost}{cost}
\newcommand{\ASratio}{AsymRatio}

\newcommand{\chaff}{Chaff}
\newcommand{\minisat}{MiniSat}
\newcommand{\zchaff}{ZChaff}

\newcommand{\pp}{\mbox{\small ++}}

\newcommand{\domain}{{\mathcal D}}
\newcommand{\size}{{\mathcal S}}
\newcommand{\ppp}{{\mathcal P}}

\newcommand{\ifTR}[1]{}
\newcommand{\ifLMCS}[1]{#1}

\def\doi{3 (1:6) 2007}
\lmcsheading%
{\doi}
{1--41}
{}
{}
{Mar.~13, 2006}
{Feb.~26, 2007}
{}

\begin{document}

\title[Goal Asymmetry and DPLL Proofs in Planning]{Structure and
  Problem Hardness:\\ Goal Asymmetry and DPLL Proofs in SAT-Based
  Planning}

\author[J.~Hoffmann]{J\"org Hoffmann\rsuper a}
\address{{\lsuper a}Digital Enterprise Research Institute, Innsbruck, Austria}
\email{joerg.hoffmann@deri.org}

\author[C.~Gomes]{Carla Gomes\rsuper b}
\address{{\lsuper{b,c}}Cornell University, Ithaca, NY, USA}
\email{\{gomes,selman\}@cs.cornell.edu}
\thanks{{\lsuper b}Research supported by the
  Intelligent Information Systems Institute, Cornell University (AFOSR
  grant F49620-01-1-0076).}

\author[B.~Selman]{Bart Selman\rsuper c}


\begin{abstract}
  In Verification and in (optimal) AI Planning, a successful method is
  to formulate the application as boolean satisfiability (SAT), and
  solve it with state-of-the-art DPLL-based procedures. There is a
  lack of understanding of why this works so well. Focussing on the
  Planning context, we identify a form of {\em problem structure}
  concerned with the symmetrical or asymmetrical nature of the cost of
  achieving the individual planning goals. We quantify this sort of
  structure with a simple numeric parameter called $\ASratio$, ranging
  between $0$ and $1$. We run experiments in 10 benchmark domains from
  the International Planning Competitions since 2000; we show that
  $\ASratio$ is a good indicator of SAT solver performance in 8 of
  these domains. We then examine carefully crafted {\em synthetic
  planning domains} that allow control of the amount of structure, and
  that are clean enough for a rigorous analysis of the combinatorial
  search space.  The domains are parameterized by size, and by the
  amount of structure. The CNFs we examine are unsatisfiable, encoding
  one planning step less than the length of the optimal plan. We prove
  upper and lower bounds on the size of the best possible DPLL
  refutations, under different settings of the amount of structure, as
  a function of size.  We also identify the best possible sets of
  branching variables (backdoors).  With {\em minimum} $\ASratio$, we
  prove exponential lower bounds, and identify minimal backdoors of
  size linear in the number of variables. With {\em maximum}
  $\ASratio$, we identify {\em logarithmic} DPLL refutations (and
  backdoors), showing a doubly exponential gap between the two
  structural extreme cases. The reasons for this behavior -- the proof
  arguments -- illuminate the prototypical patterns of structure
  causing the empirical behavior observed in the competition
  benchmarks.
\end{abstract}

\maketitle
\vfill\eject

\input{introduction}

\input{related}

\input{prelim}

\input{real}

\input{sd}

\input{empirical}

\input{strawman}

\input{conclusion}


\noindent {\bf Acknowledgements.} We thank Ashish Sabharwal for advice
on proving resolution lower bounds. We thank Rina Dechter for
discussions on the relation between backdoors and cutsets. We thank
Toby Walsh for discussions on the intuition behind $\ASratio$, and for
the suggestion to construct a red herring. We thank Jussi Rintanen for
providing the code of his random instance generator, and we thank
Martin G\"utlein for implementing the enumeration of CNF variable sets
under exploitation of problem symmetries. We finally thank the
anonymous reviewers for their comments, which contributed a lot to
improve the paper.


\ifTR{
\newpage
\appendix

\input{proofs}

\input{cutsets}

}

\end{document}

%% file: introduction.tex
\section{Introduction}
\label{introduction}


There has been a long interest in a better understanding of what makes
combinatorial problems from SAT and CSP hard or easy. The most
successful work in this area involves random instance distributions
with phase transition characterizations (e.g.,
\cite{cheeseman:etal:ijcai-91,Hogg96}).
However, the link of these
results to more structured instances is less direct. A random
unsatisfiable 3-SAT instance from the phase transition region with
1,000 variables is beyond the reach of any current solver. On the
other hand, many unsatisfiable formulas from Verification and AI
Planning contain well over 100,000 variables and can be proved
unsatisfiable within a few minutes (e.g., with \chaff\
\cite{moskewicz:etal:dac-01}). This raises the question as to whether
one can obtain general measures of structure in SAT encodings, and use
them to characterize typical case complexity. Herein, we address this
question in the context of AI Planning.  We view this as an entry
point to similar studies in other areas.


In Planning, methods are developed to automatically solve the
reachability problem in declaratively specified transition systems.
That is, given some formalism to describe system states and
transitions (``actions''), the task is to find a solution (``plan''):
a sequence of transitions leading from some given start (``initial'')
state to a state that satisfies some given non-temporal formula (the
``goal''). Herein, we consider the wide-spread formalism known as
``STRIPS Planning'' \cite{fikes:nilsson:ai-71}. This is a very simple
formal framework making use of only Boolean state variables
(``facts''), conjunctions of positive atoms, and atomic transition
effects; details and notations are given later (Section~\ref{prelim}).


We focus on showing infeasibility, or, in terms of Planning, on
  proving optimality of plans. We consider the difficulty of showing
  the non-existence of a plan with one step less than the shortest
  possible -- {\em optimal} -- plan. SAT-based search for a plan
  \cite{kautz:selman:ecai-92,kautz:selman:aaai-96,kautz:selman:ijcai-99}
  works by iteratively incrementing a plan length bound $b$, and
  testing in each iteration a formula that is satisfiable iff there
  exists a plan with $b$ steps.  So, our focus is on the last
  unsuccessful iteration in a SAT-based plan search, where the
  optimality of the plan (found later) is proved. This focus is
  relevant since proving optimality is precisely what SAT solvers are
  best for, at the time of writing, in the area of Planning. On the
  one hand, SAT-based planning won the 1st prize for optimal planners
  in the 2004 International Planning Competition
  \cite{hoffmann:edelkamp:jair-05} as well as the 2006 International
  Planning Competition.  On the other hand, finding potentially
  sub-optimal plans is currently done much faster based on heuristic
  search (with non-admissible heuristics), e.g.
  \cite{bonet:geffner:ai-01,hoffmann:nebel:jair-01,gerevini:etal:jair-03,helmert:icaps-04,wah:chen:ijait-04}.


In our work, we first formulate an intuition about what makes DPLL
\cite{davis60,dll62} search hard or easy in Planning. We design a
numeric measure of that sort of problem structure, and we show
empirically that the measure is a good indicator of search performance
in many domains. We also perform a case study: We design synthetic
domains that capture the problem structure in a clean form, and we
analyze DPLL behavior in detail, within these domains.

\subsection{Goal Asymmetry}

In STRIPS Planning, the goal formula is a conjunction of goal facts,
where each fact requires a Boolean variable to be true. The goal facts
are commonly referred to as {\em goals}, and their conjunction is
referred to as a {\em set} of goals. In most if not all benchmark
domains that appear in the literature, the individual goal facts
correspond quite naturally to individual ``sub-problems'' of the
problem instance ({\em task}) to be solved.  The sub-problems
typically interact with each other -- and this is the starting point
of our investigation. 

Given tasks with the same optimal plan length, our intuitions are
these. (1) Proving plan optimality is hard if the optimal plan length
arises from complex interactions between many sub-problems.  (2)
Proving plan optimality is easy if the optimal plan length arises
mostly from a single sub-problem. To formalize these intuitions, we
take a view based on sub-problem ``cost'', serving to express both
intuitions with the same notion, and offering the possibility to
interpolate between (1) and (2). By ``cost'', here, we mean the number
of steps needed to solve a (sub-)problem.  We distinguish a {\em
symmetrical case} -- where the individual sub-problems are all
(symmetrically) ``cheap'' -- and an {\em asymmetrical case} -- where a
single sub-problem (asymmetrically) ``dominates the overall
cost''.\footnote{Please note that this use of the word ``symmetrical''
has nothing to do with the wide-spread notion of ``problem
symmetries'' in the sense of problem permutations; the cost of the
goals has no implications on whether or not parts of the problem can
be permuted.}

The asymmetrical case obviously corresponds to intuition (2) above.
The symmetrical case corresponds to intuition (1) because of the
assumption that the number of steps needed to solve the overall task
is the same in both cases. If each single sub-problem is cheap, but
their conjunction is costly, then that cost must be the result of some
sort of ``competition for a resource'' -- an interaction between the
sub-problems. One can interpolate between the symmetrical and
asymmetrical cases by measuring to what extent any single sub-problem
dominates the overall cost.

It remains to define what ``dominating the overall cost''
means. Herein, we choose a simple maximization and normalization
operation. We select the most costly goal and divide that by the cost
of achieving the conjunction of all goals. For a conjunction $C$ of
facts let $\cost(C)$ be the length of a shortest plan achieving
$C$. Then, for a task with goal set $G$, $\ASratio$ is defined as
$max_{g \in G} \cost(g)/\cost(\bigwedge_{g \in G} g)$.  $\ASratio$
ranges between $0$ and $1$. Values close to $0$ correspond to the
symmetrical case, values close to $1$ correspond to the asymmetrical
case.  With the intuitions explained above, $\ASratio$ should be
thought of as an indirect measure of the degree of sub-problem
interactions. That is, we interpret the value of $\ASratio$ as an {\em
effect} of those interactions. An important open question is whether
more direct measures -- syntactic definitions of the {\em causes} of
sub-problem interactions -- can be found. Starting points for
investigations in this direction may be existing investigations of
``sub-goals'' and their dependencies
\cite{bundy:etal:ai-96,hoffmann:etal:jair-04,hoffmann:jair-05}; we
outline these investigations, and their possible relevance, in
Section~\ref{conclusion}.

Note that $\ASratio$ is a rather simplistic parameter. Particularly,
the use of maximization over individual costs, disregarding any
interactions between the facts, can be harmful. The maximally costly
goal may be independent of the other goals. In such a case, while
taking many steps to achieve, the goal is not the main reason for the
length of the optimal plan. An example for this is given in
Section~\ref{strawman}; we henceforth refer to this as the ``red
herring''.

\vspace{-0.0cm}
\begin{figure}
  \begin{center}
    \begin{tabular}{cc}\vspace{0.3cm}
      \includegraphics[width=5.8cm]{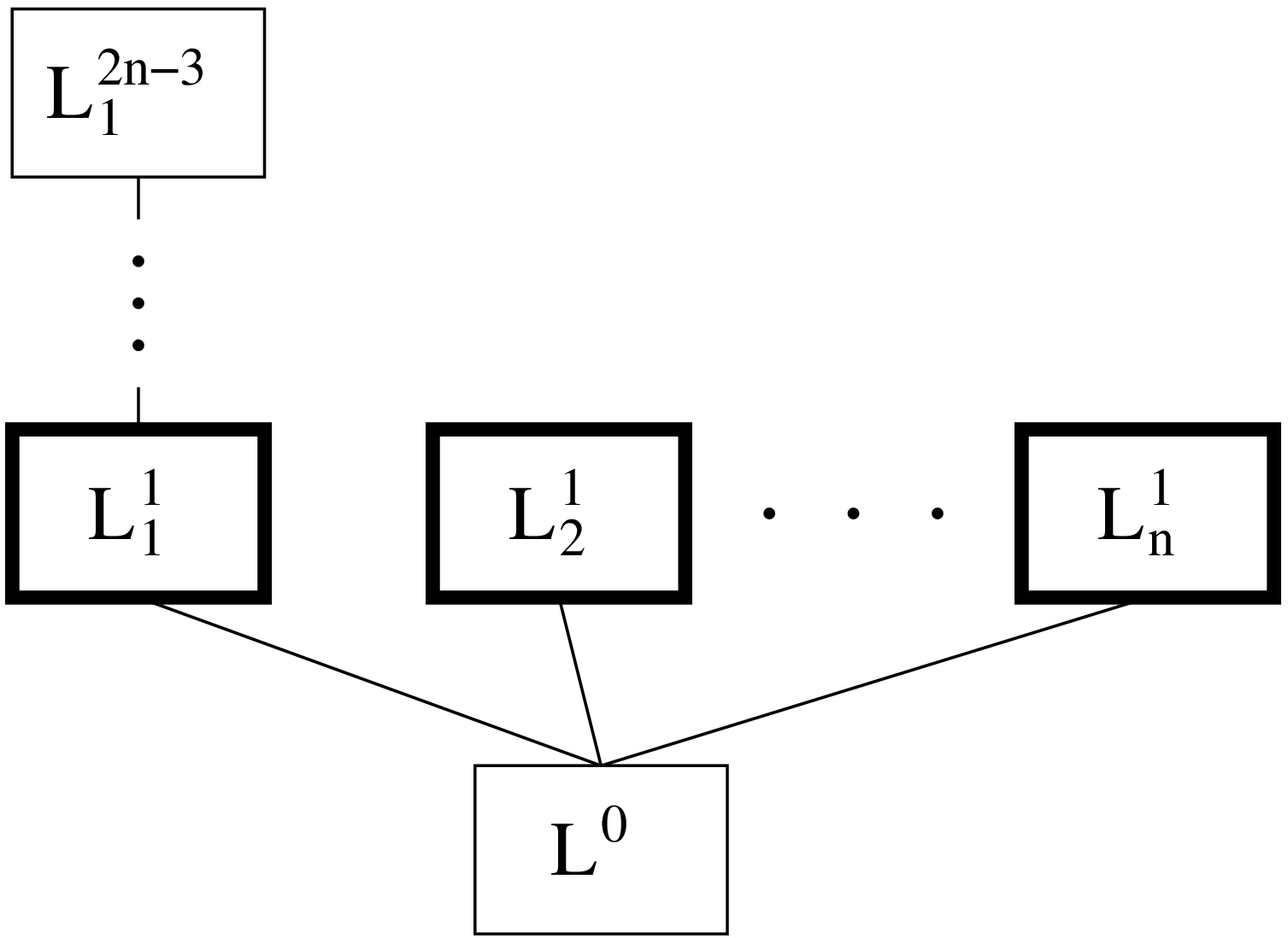} \hspace{0.3cm} & \hspace{0.3cm} \includegraphics[width=5.8cm]{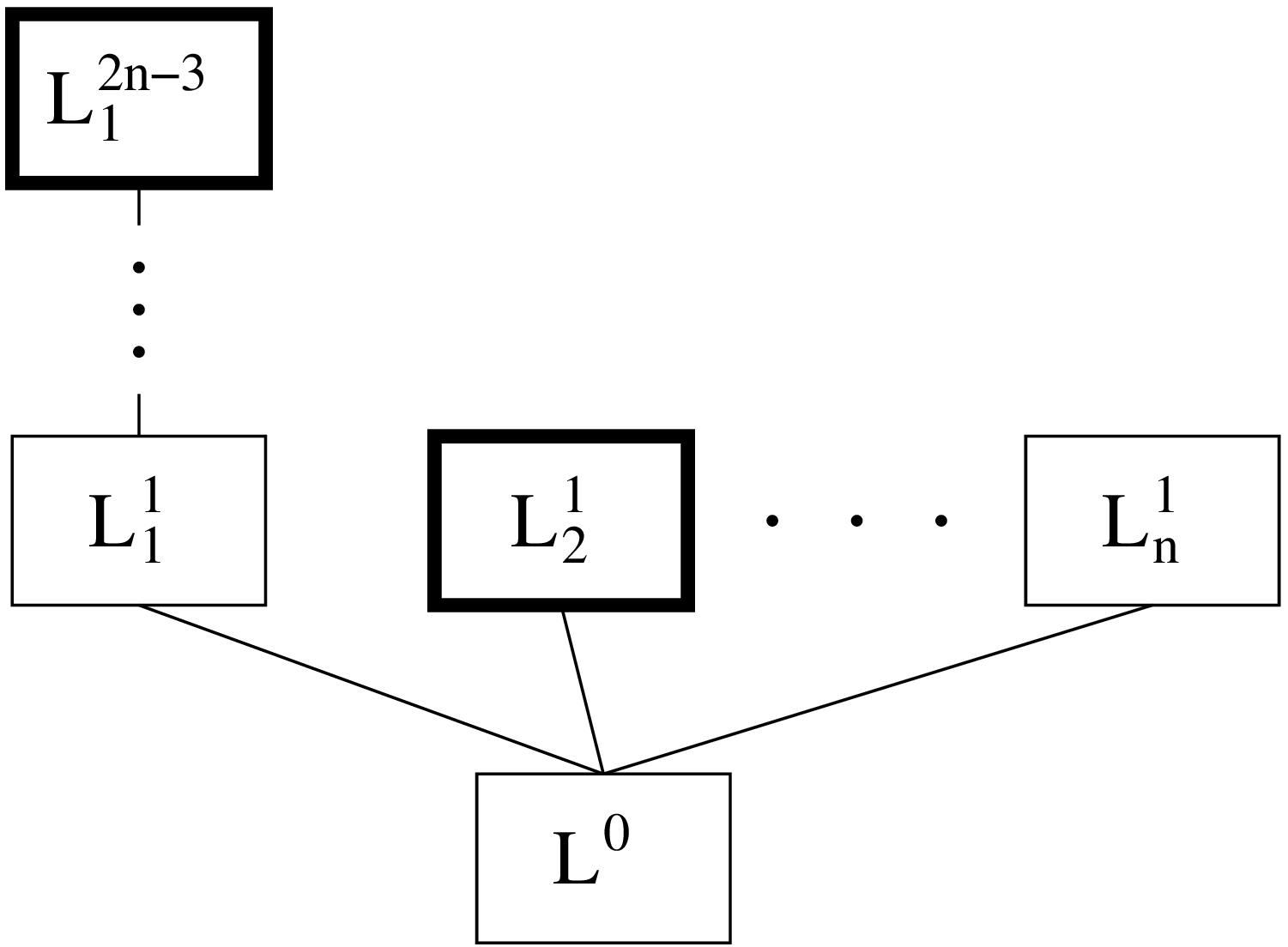}\\
       (a) goals symmetrical case & (b) goals asymmetrical case\\
    \end{tabular}
  \end{center}
  \vspace{-0.0cm}
  \caption{\label{mapintro}The MAP domain. Goal nodes indicated in bold face.}
  \vspace{-0.5cm}
\end{figure}

\begin{sloppypar}
  To make the above concrete, Figure~\ref{mapintro} provides a sketch
  of one of our synthetic domains, called {\em MAP}. One moves in an
  undirected graph and must visit a subset of the nodes. The domain
  has two parameters: the size, $n$, and the amount of structure,
  $k$. The number of nodes in the graph is $3n-3$. The value of $k$
  changes the set of goal nodes. One always starts in the bottom node,
  $L^0$. If $k$ is $1$, then all goal nodes have distance $1$ from the
  start node. For every increase of $k$ by $2$, a single one of the
  goal nodes wanders two steps (edges) further away, and one of the
  other goal nodes is skipped. As a result, the overall plan length is
  $2n-1$ independently of $k$. Our formulas encode the unsolvable task
  of finding a plan with at most $2n-2$ steps; they have $\Theta(n^2)$
  variables ($\Theta(n)$ graph nodes at $\Theta(n)$ time steps). We
  have $\ASratio = k/(2n-1)$.  In particular, for increasing $n$,
  $\ASratio$ converges to $0$ for $k=1$ -- symmetrical case -- and it
  converges to $1$ for $k=2n-3$ -- asymmetrical case.\footnote{When
  setting $k=2n-1$, the formula contains an empty clause.} Note how
  all the goal nodes interact in the symmetrical case, competing for
  the time steps. Note also that the outlier goal node is not
  independent of the other goals, but interacts with them in the same
  competition for the time steps. In our red herring example,
  Section~\ref{strawman}, the main idea is to make the outlier goal
  node independent by placing it on a separate map, and allowing to
  move in parallel on both maps. Note finally that, in
  Figure~\ref{maproadmap} (a), the graph nodes $L_2^1, \dots, L_n^1$
  can be permuted. This is an artefact of the abstract nature of
  MAP. In some experiments in competition examples, we did not find a
  low $\ASratio$ to be connected with a high amount of problem
  symmetries (see also Sections~\ref{sd:map}).
\end{sloppypar}

Beside MAP, we constructed two domains called SBW and SPH. SBW is a
block stacking domain. There are $n$ blocks, and $k$ controls the
amount of stacking restrictions. In the symmetrical case, there are no
restrictions. In the asymmetrical case, one particular stack of blocks
takes many steps to build. SPH is a non-planning domain, namely a
structured version of the pigeon hole problem. There are $n+1$ pigeons
and $n$ holes. The parameter $k$ controls how many holes one
particular ``bad'' pigeon needs, and how many ``good'' pigeons there
are, which can share a hole with the ``bad'' one. The total number of
holes needed remains $n+1$, independently of $k$. The symmetrical case
is the standard pigeon hole. In the asymmetrical case, the bad pigeon
needs $n-1$ holes.

\subsection{Results Overview}
\label{introduction:results}

In our research, the analysis of synthetic domains served to find and
explore intuitions about how structure influences search
performance. The final results of the analysis are studies of a set of
prototypical behaviors. These prototypical behaviors are inherent also
to more realistic examples; in particular, we will outline some
examples from the competition benchmarks.

We prove upper and lower bounds on the size of the best-case DPLL
proof trees. We also investigate the best possible sets of branching
variables. Such variable sets were recently coined ``backdoors''
\cite{williams:etal:ijcai-03}. In our context, a backdoor is a subset
of the variables so that, for every value assignment to these
variables, unit propagation (UP) yields an empty clause.\footnote{In
general, a backdoor is defined relative to an arbitrary polynomial
time ``subsolver'' procedure. The subsolver can solve some class of
formulas that does not necessarily have a syntactic
characterization. Our definition here instantiates the subsolver with
the widely used unit propagation procedure.} That is, a smallest
possible backdoor encapsulates the best possible branching variables
for DPLL, a question of huge practical interest. Identifying backdoors
is also a technical device: we obtain our upper bounds as a side
effect of the proofs of backdoor properties. In all formula classes we
consider, we determine a backdoor subset of variables. We prove that
the backdoors are {\em minimal}: no variable can be removed without
losing the backdoor property. In small enough instances, we prove
empirically that the backdoors are in fact {\em optimal} - of minimal
size. We conjecture that the latter is true in general.

In the symmetrical case, for MAP and SPH there are exponential (in
$n$) lower bounds on the size of resolution refutations, and thus on
DPLL refutations \cite{bonet:etal:siam-00,beame:etal:jair-04}. For
SPH, the lower bound is just the known result
\cite{buss:pitassi:csl-97} for the standard Pigeon Hole problem. For
MAP, we construct a polynomial reduction of resolution proofs for MAP
to resolution proofs for a variant of the Pigeon Hole problem
\cite{razborov:jcss:04}. The reduction is via an intriguing temporal
version of the pigeon hole problem, where the holes correspond to the
time steps in the planning encoding, in a natural way. This
illustrates quite nicely the competition of tasks for time steps that
underlies also more realistic examples.

For SBW, it is an open question whether there exists an exponential
lower bound on DPLL proof size in the symmetrical case; we conjecture
that there is. In all the domains, the backdoor sets in the
symmetrical case are linear in the total number of variables:
$\Theta(n^2)$ for MAP and SPH, $\Theta(n^3)$ for SBW.

In the asymmetrical case, the DPLL proofs and backdoors become much
smaller. In SPH, the minimal backdoors have size $O(n)$. What
surprised us most is that in MAP and SBW the backdoors even become
logarithmic in $n$. Considering Figure~\ref{mapintro} (b), one would
suspect to obtain an $O(n)$ backdoor reasoning along the path to the
outlier goal node. However, it turns out that one can pick the
branching variables in a way exploiting the need to go back and forth
on that path. Going back and forth introduces a factor $2$, and so one
can double the number of time steps between each pair of
variables. The resulting backdoor has size $O(log n)$. This very
nicely complements recent work \cite{williams:etal:ijcai-03}, where
several benchmark planning tasks were identified that contain
exorbitantly small backdoors in the order of 10 out of 10000
variables. Our scalable examples with provably logarithmic backdoors
illustrate the reasons for this phenomenon. Notably, the DPLL proof
trees induced by our backdoors in the asymmetrical case degenerate to
lines. Thus we get an exponential gap in DPLL proof size for SPH, and
a doubly exponential gap for MAP.

To confirm that $\ASratio$ is an indicator of SAT solver performance
in more practical domains, we run large-scale experiments in 10
domains from the biennial International Planning Competitions since
2000.  The main criterion for domain selection is the availability of
an instance generator. These are necessary for our experiments, where
we generate and examine thousands of instances in each domain, in
order to obtain large enough samples with identical plan length and
$\ASratio$ (see below). We use the most successful SAT encoding for
Planning: the ``Graphplan-based'' encoding
\cite{kautz:selman:aaai-96,kautz:selman:ijcai-99}, which has been used
in all planning competitions since 1998. We examine the performance of
a state-of-the-art SAT solver, namely, \zchaff\
\cite{moskewicz:etal:dac-01}. We compare the distributions of search
tree size for pairs ${\mathcal P}^x$ and ${\mathcal P}^y$ of sets of
planning tasks. All the tasks share the same domain, the same instance
size parameters (taken from the original competition instances), and
the same optimal plan length. The only difference is that $\ASratio =
x$ for the tasks in ${\mathcal P}^x$, and $\ASratio = y$ for the tasks
in ${\mathcal P}^y$, where $x < y$. Very consistently, the mean search
tree size is significantly higher in ${\mathcal P}_m^x$ than in
${\mathcal P}_m^y$. The T test yields confidence level $95\%$
($99.9\%$, most of the time) in the vast majority of the cases in 8 of
the 10 domains. In one domain (Logistics), the T test fails for almost
all pairs ${\mathcal P}_m^x$ and ${\mathcal P}_m^y$; in another domain
(Satellite), $\ASratio$ does not show any variance so there is nothing
to measure.

Some remarks are in order. Note that $\ASratio$ characterizes a kind
of hidden structure. It cannot be computed efficiently even based on
the original planning task representation -- much less based on the
SAT representation. This nicely reflects practical situations, where
it is usually impossible to tell a priori how difficult a formula will
be to handle. In fact, one interesting feature of our MAP formulas is
that their syntax hardly changes between the two structural extreme
cases.  The content of a single clause makes all the difference
between exponential and logarithmic DPLL proofs.

Note further that $\ASratio$ is not a completely impractical
parameter. There exists a wealth of well-researched techniques
approximating plan length, e.g.\
\cite{blum:furst:ai-97,bonet:geffner:ai-01,hoffmann:nebel:jair-01,edelkamp:ecp-01,gerevini:etal:jair-03,helmert:icaps-04,haslum:etal:aaai-05}.
Such techniques can be used to approximate $\ASratio$. We leave this
as a topic for future work.

Note finally that, of course, $\ASratio$ can be fooled. (1) One can
easily modify the goal set in a way blinding $\ASratio$, for example
by replacing the goal set with a single goal that has the same
semantics. (2) One can also construct examples where a relevant
phenomenon is ``hidden'' behind an irrelevant phenomenon that controls
$\ASratio$, and DPLL tree size grows, rather than decreases, with
$\ASratio$. For (1), we outline in Section~\ref{conclusion} how this
could be circumvented.  As for (2), we explain the construction of
such an example in Section~\ref{strawman}. As a reply to both, it is
normal that heuristics can be fooled. What matters is that $\ASratio$
often is informative in the domains that people actually try to solve.

The paper is organized as follows. Section~\ref{related} discusses
related work. Section~\ref{prelim} provides notation and some details
on the SAT encodings we use. Section~\ref{real} contains our
experiments with $\ASratio$ in competition
benchmarks. Section~\ref{sd} describes our synthetic domains and our
analysis of DPLL proofs; Section~\ref{empirical} briefly demonstrates
that state-of-the-art SAT solvers -- \zchaff\
\cite{moskewicz:etal:dac-01} and \minisat\
\cite{een:soerensson:sat-03,een:biere:sat-05} -- indeed behave as
expected, in these domains. Section~\ref{strawman} describes our red
herring example, where $\ASratio$ is {\em not} an indicator of DPLL
proof size. Section~\ref{conclusion} concludes and discusses open
topics.

\ifLMCS{For readability, the paper includes only proof sketches. The
full proofs, and some other details, can be looked up in a TR
\cite{hoffmann:etal:tr-lmcs06}.}

\ifTR{Appendix~\ref{proofs} contains all proofs; the main text
contains proof sketches. Appendix~\ref{cutsets} provides details about
cutsets in our synthetic domains.}

%% file: related.tex
\section{Related Work}
\label{related}

A huge body of work on structure focusses on phase transition
phenomena, e.g.,
\cite{cheeseman:etal:ijcai-91,gent:walsh:ecai-94,Hogg96,rintanen:icaps-04}. There
are at least two major differences to our work. First, as far as we
are aware, all work on phase transitions has to do with transitions
between areas of under-constrained instances and over-constrained
instances, with the critically constrained area -- the phase
transition -- in the middle. In contrast, the Planning formulas we
consider are all just one step short of a solution. In that sense,
they are all ``critically constrained''. As one increases the plan
length bound, for a single planning task, from $0$ to $\infty$, one
naturally moves from an over-constrained into an under-constrained
area, where the bounds close to the first satisfiable iteration
constitute the most critically constrained region. Put differently,
phase transitions are to do with the balance of duties and resources;
in our formulas, per definition, the amount of resources is set to a
level that is just not enough to fulfill the duties.

A second difference to phase transitions is that these are mostly
concerned with random instance distributions. In such distributions,
the typical instance hardness is completely governed by the parameter
settings -- e.g., the numbers of variables and clauses in a standard
$k$-CNF generation scheme. This is very much not so in more practical
instance distributions. If the contents of the clauses are extracted
from some practical scenario, rather than from a random number
generator, then the numbers of variables and clauses alone are not a
good indicator of instance hardness: these numbers may not reflect the
semantics of the underlying application. A very good example for this
are our MAP formulas. These are syntactically almost identical at both
ends of the structure scale; their DPLL proofs exhibit a doubly
exponential difference.

Another important strand of work on structure is concerned with the
identification of tractable classes, e.g.,
\cite{bylander:ai-94,dechter03,chen:dalmau:cp-05}. Our work obviously
differs from this in that we do not try to identify general provable
connections between structure and hardness. We identify empirical
correlations, and we study particular cases.

Our analysis of synthetic examples is, in spirit, similar to the work
in proof complexity, e.g.,
\cite{cook:reckhow:jsl-79,haken:tcs-85,buss:pitassi:csl-97,iwama:miyazaki:ac-99}.
There, formula families such as pigeon hole problems have been the key
to a better understanding of resolution proofs. Generally speaking,
the main difference is that, in proof complexity, one investigates the
behavior of different proof calculi in the same example. In contrast,
we consider the single proof calculus DPLL, and modify the
examples. Major technical differences arise also due to the kinds of
formulas considered, and the central goal of the research. Proof
complexity considers any kind of synthetic formula provoking a certain
behavior, while we consider formulas from Planning. The goal in proof
complexity is mostly to obtain lower bounds, separating the power of
proof systems. However, to understand structure, and explain the good
performance of SAT solvers, interesting formula families with {\em
small} DPLL trees are more revealing.

There is some work on problem structure in the Planning
community. Some works \cite{barrett:weld:ai-94,veloso:blythe:aips-94}
investigate structure in the context of causal links in partial-order
planning. Howe and Dahlman \cite{howe:dahlman:jair-02} analyze planner
performance from a perspective of syntactic changes and computational
environments. Hoffmann \cite{hoffmann:jair-05} investigates
topological properties of certain wide-spread heuristic
functions. Obviously, these works are quite different from ours. A
more closely related piece of work is the aforementioned investigation
of backdoors recently done by Williams et al
\cite{williams:etal:ijcai-03}. In particular, this work showed
empirically that CNF encodings of many standard Planning benchmarks
contain exorbitantly small backdoors in the order of 10 out of 10000
variables. The existence of logarithmic backdoors in our synthetic
domains nicely reflects this.  In contrast to the previous results, we
also explain what these backdoor variables are -- what they correspond
to in the original planning task -- and how their interplay works.

A lot of work on structure can also be found in the Scheduling
community, e.g.,
\cite{watson:etal:ecp-01,watson:etal:informs-02,watson:etal:ai-03,streeter:smith:icaps-05}. To
a large extent, this work is to do with properties of the search space
surface, and its effect on the performance of local search. A closer
relative to our work is the notion of {\em critical paths} as used in
Scheduling, e.g,
\cite{kirkpatrick:clark:ibm-66,yen:etal:dac-89,sadowska:xiao:dac-01}. Based
on efficiently testable inferences, a critical path identifies the
(apparently) most critically constrained part of the scheduling
task. For example, a critical path may identify a long necessary
sequence of job executions implied by the task specification. This
closely corresponds to our notion of a high $\ASratio$, where a single
goal fact is almost as costly to solve as the entire planning
task. Indeed, such a goal fact tends to ease search by providing a
sort of critical path on which a lot of constraint propagation
happens. In that sense, our work is an application of critical paths
to Planning. Note that Planning and Scheduling are different. In
Planning, there is much more freedom of problem design. Our main
observation in here is that our notions of problem structure capture
interesting behavior across a {\em range} of domains.

There is also a large body of work on structure in the constraint
reasoning community, e.g.,
\cite{dechter:ai-90,cheeseman:etal:ijcai-91,gent:walsh:ecai-94,frank:etal:jair-97,rish:dechter:jar-00,SW01,Z02,dechter03,Nudelman04,Hulubei05,chen:dalmau:cp-05}.
As far as we are aware, all these works differ considerably from
ours. In particular, all works we are aware of define ``structure'' on
the level of the CNF formula/the CSP problem instance; in contrast, we
define structure on the level of the modelled application. Further,
empirical work on structure is mostly based on random problem
distributions, and theoretical analysis is mostly done in a proof
complexity sense, or in the context of identifying tractable classes.

One structural concept from the constraint reasoning community is
particularly closely related to the concept of a backdoor: {\em
cutsets} \cite{dechter:ai-90,rish:dechter:jar-00,dechter03}. Cutset
are defined relative to the {\em constraint graph}: the undirected
graph where nodes are variables and edges indicate common membership
in at least one clause. A cutset is a set of variables so that, once
these variables are removed from the constraint graph, that graph has
a bounded {\em induced width}; if the bound is $1$, then the graph is
cycle-free, i.e., can be viewed as a tree. Backdoors are a
generalization of cutsets in the sense that any cutset is a backdoor
relative to an appropriate subsolver (that exploits properties of the
constraint graph). The difference is that cutsets have an ``easy''
syntactic characterization: one can check in polytime if or if not a
given set of variables is a cutset. One can, thus, use strategies
looking for cutsets to design search algorithms.  Indeed, the cutset
notion was originally developed with that aim.  Backdoors, in
contrast, were proposed as a means to characterize phenomena relevant
for existing state-of-the-art solvers -- which all make use of
subsolvers whose capabilities (the solved classes of formulas) have no
easy-to-test syntactic characterization. In particular, the effect of
unit propagation depends heavily on what values are assigned to the
backdoor variables. We will see that, in the formula families
considered herein, there are no small cutsets.

%% file: prelim.tex
\section{Preliminaries}
\label{prelim}

We use the STRIPS formalism. States are described as sets of (the
currently true) propositional facts. A planning {\em task} is a tuple
of {\em initial state} (a set of facts), {\em goal} (also a set of
facts), and a set of {\em actions}. Actions $a$ are fact set triples:
the {\em precondition} $pre(a)$, the {\em add effect} $add(a)$, and
the {\em delete effect} $del(a)$.  The semantics are that an action is
applicable to a state (only) if $pre(a)$ is contained in the
state. When executing the action, the facts in $add(a)$ are included
into the state, and the facts in $del(a)$ are removed from it. The
intersection between $add(a)$ and $del(a)$ is assumed empty; executing
a non-applicable action results in an undefined state. A {\em plan}
for the task is a sequence of actions that, when executed iteratively,
maps the initial state into a state that contains the goal.

As a simple example, consider the task of finding a path from a start
node $n_0$ to a goal node $n_g$ in a directed graph $(N,E)$. The facts
have the form $at\mbox{-}n$ for $n \in N$. The initial state is
$\{at\mbox{-}n_0\}$, the goal is $\{at\mbox{-}n_g\}$, and the set of
actions is $\{move\mbox{-}x\mbox{-}y = (\{at\mbox{-}x\},
\{at\mbox{-}y\}, \{at\mbox{-}x\}) \mid (x,y) \in E\}$ -- precondition
$\{at\mbox{-}x\}$, add effect $\{at\mbox{-}y\}$, delete effect
$\{at\mbox{-}x\}$. Plans correspond to the paths between $n_0$ and
$n_g$.

CNF formulas are sets of clauses, where each clause is a set of
literals. For a CNF formula $\phi$ with variable set $V$, a variable
subset $B \subseteq V$, and a truth value assignment $a$ to $B$, by
$\phi_a$ we denote the CNF that results from inserting the values
specified by $a$, and removing satisfied clauses as well as
unsatisfied literals.  By $UP(\phi)$, we denote the result of iterated
application of unit propagation to $\phi$, where again satisfied
clauses and unsatisfied literals are removed.  For a CNF formula
$\phi$ with variable set $V$, a variable subset $B \subseteq V$, and a
value assignment $a$ to $B$, we say that $a$ is {\em UP-consistent} if
$UP(\phi_a)$ does not contain the empty clause.  $B$ is a {\em
backdoor} if it has no UP-consistent assignment. We sometimes use the
abbreviation {\em UP} also in informal text.

By a {\em resolution refutation}, also called {\em resolution proof},
of a (unsatisfiable) formula $\phi$, we mean a sequence $C_1, \dots,
C_m$ of clauses $C_i$ so that $C_m = \emptyset$, and each $C_i$ is
either an element of $\phi$, or derivable by the resolution rule from
two clauses $C_j$ and $C_k$, $j, k < i$. The {\em size} of the
refutation is $m$. A {\em DPLL refutation}, also called {\em DPLL
  proof} or {\em DPLL tree}, for $\phi$ is a tree of partial
assignments generated by the DPLL procedure, where the inner nodes are
UP-consistent, and the leaf nodes are not. The size of a DPLL
refutation is the total number of (inner and leaf) tree nodes.

Planning can be mapped into a sequence of SAT problems, by
incrementally increasing a plan length bound $b$: start with $b=0$;
generate a CNF $\phi(b)$ that is satisfiable iff there is a plan with
$b$ steps; if $\phi(b)$ is satisfiable, stop; else, increment $b$ and
iterate. This process was first implemented in the {\em Blackbox}
system
\cite{kautz:selman:ecai-92,kautz:selman:aaai-96,kautz:selman:ijcai-99}.
There are, of course, different ways to generate the formulas
$\phi(b)$, i.e., there are different encoding methods. In our
empirical experiments, we use the original {\em Graphplan-based}
encoding used in Blackbox. Variants of this encoding have been used by
Blackbox (more recently named {\em SATPLAN}) in all international
planning competitions since 1998. In our theoretical investigations,
we use a somewhat simplified version of the Graphplan-based
encoding.\footnote{One might wonder whether a different encoding would
fundamentally change the results herein. We don't see a reason why
that should be the case. Exploring this is a topic for future work.}

The Graphplan-based encoding is a straightforward translation of a
$b$-step {\em planning graph} \cite{blum:furst:ai-97} into a CNF. The
encoding has $b$ {\em time steps} $1 \leq t \leq b$. It features
variables for facts at time steps, and for actions at time steps. The
former encode commitments of the form ``fact $p$ is true/false at time
$t$'', the latter encode commitments of the form ``action $a$ is
executed/not executed at time $t$''. There are artificial {\em NOOP}
actions, i.e.\ for each fact $p$ there is an action $NOOP\mbox{-}p$
whose only precondition is $p$, and whose only effect is $p$. The
NOOPs are treated just like normal actions in the encoding. Setting a
NOOP variable to true means a commitment to ``keep fact $p$ true at
time $t$''.  Amongst others, there are clauses to ensure that all
action preconditions are satisfied, that the goals are true in the
last time step, and that no ``mutex'' actions are executed in the same
time step: actions can be executed in the same time step -- in
parallel -- if their effects and preconditions are not contradictory.
The set of fact and action variables at each time step, as well as
pairs of ``mutex'' facts and actions, are read off the planning graph
(which is the result of a propagation of binary constraints).

We do not describe the Graphplan-based encoding in detail since that
is not necessary to understand our experiments. For the simplified
encoding used in our theoretical investigations, some more details are
in order. The encoding uses variables only for the actions (including
NOOPs), i.e., $a(t)$ is $1$ iff action $a$ is to be executed at time
$t$, $1 \leq t \leq b$. A variable $a(t)$ is included in the CNF iff
$a$ is {\em present} at $t$. An action $a$ is present at $t=1$ iff
$a$'s precondition is true in the initial state; $a$ is present at
$t>1$ iff, for every $p \in pre(a)$, at least one action $a'$ is
present at $t-1$ with $p \in add(a')$. For each action $a$ present at
a time $t>1$ and for each $p \in pre(a)$, there is a {\em
  precondition} clause of the form $\{\neg a(t), a_1(t-1), \dots,
a_l(t-1)\}$, where $a_1, \dots, a_l$ are all actions present at $t-1$
with $p \in add(a_i)$.  For each goal fact $g \in G$, there is a {\em
  goal} clause $\{a_1(b), \dots, a_l(b)\}$, where $a_1, \dots, a_l$
are all actions present at $b$ that have $g \in add(a_i)$. Finally,
for each {\em incompatible} pair $a$ and $a'$ of actions present at a
time $t$, there is a {\em mutex} clause $\{\neg a(t), \neg a'(t)\}$.
Here, a pair $a$, $a'$ of actions is called incompatible iff either
both are not NOOPs, or $a$ is a NOOP for fact $p$ and $p \in del(a')$.

\vspace{-0.0cm} \widthfig{6.0}{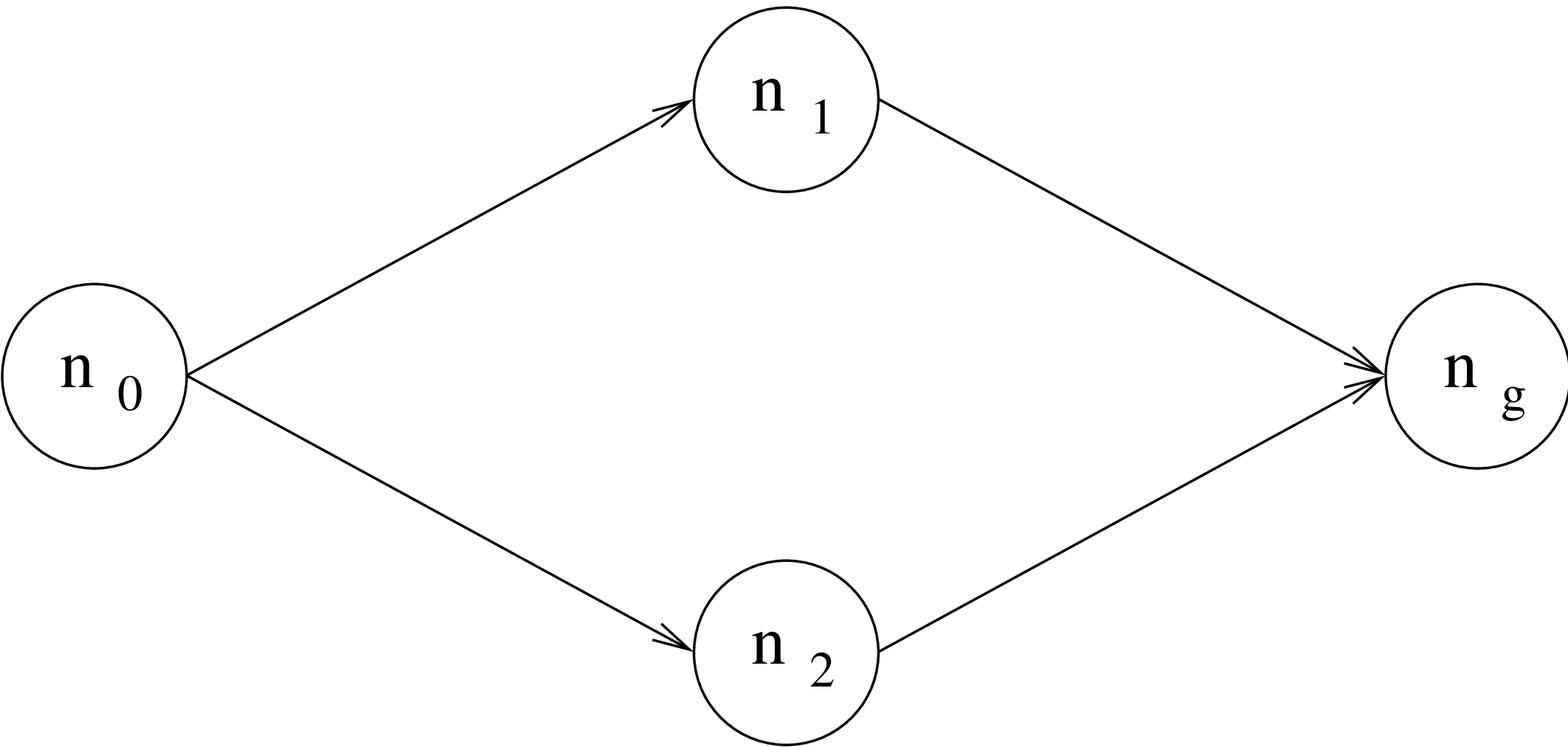}{An illustrative
example: the planning task is to move from $n_0$ to $n_g$ within $2$
steps.}{-0.0cm}

\begin{sloppypar}
  For example, reconsider the path-finding domain sketched above. Say
  $(N,E) = (\{n_0,n_1,n_2,n_g\},\{(n_0,n_1), (n_0,n_2), (n_1,n_g),
  (n_2,n_g)\})$. There are two paths from $n_0$ to $n_g$, one through
  $n_1$, the other through $n_2$. An illustration is in
  Figure~\ref{XFIG/graph.eps}. The simplified Graphplan-based encoding
  of this task, for bound $b=2$, is as follows. The variables for
  actions present at $t=1$ are $move\mbox{-}n_0\mbox{-}n_1(1)$,
  $move\mbox{-}n_0\mbox{-}n_2(1)$, $NOOP\mbox{-}at\mbox{-}n_0(1)$. The
  variables for actions present at $t=2$ are
  $move\mbox{-}n_0\mbox{-}n_1(2)$, $move\mbox{-}n_0\mbox{-}n_2(2)$,
  $NOOP\mbox{-}at\mbox{-}n_0(2)$, $move\mbox{-}n_1\mbox{-}n_g(2)$,
  $move\mbox{-}n_2\mbox{-}n_g(2)$, $NOOP\mbox{-}at\mbox{-}n_1(2)$,
  $NOOP\mbox{-}at\mbox{-}n_2(2)$. Note here that we can choose to do
  useless things such as staying at a node, via a NOOP action. We have
  to insert precondition clauses for all actions at $t=2$. For
  readability, we only show the ``relevant'' clauses, i.e., those that
  suffice, in our particular example here, to get the correct
  satisfying assignments. The relevant precondition clauses are those
  for the actions $move\mbox{-}n_1\mbox{-}n_g(2)$ and
  $move\mbox{-}n_2\mbox{-}n_g(2)$, which are $\{\neg
  move\mbox{-}n_1\mbox{-}n_g(2), move\mbox{-}n_0\mbox{-}n_1(1)\}$ and
  $\{\neg move\mbox{-}n_2\mbox{-}n_g(2),
  move\mbox{-}n_0\mbox{-}n_2(1)\}$, respectively; in order to move
  from $n_1$ to $n_g$ ($n_2$ to $n_g$) at time $2$, we must move from
  $n_0$ to $n_1$ ($n_0$ to $n_2$) at time $1$. We get similar clauses
  for the other (useless) actions, for example $\{\neg
  move\mbox{-}n_0\mbox{-}n_1(2), NOOP\mbox{-}at\mbox{-}n_0(1)\}$. We
  get the goal clause $\{move\mbox{-}n_1\mbox{-}n_g(2),
  move\mbox{-}n_2\mbox{-}n_g(2)\}$; to achieve our goal, we must move
  to $n_g$ from either $n_1$ or $n_2$, at time $2$.\footnote{Note
  that, if $b$ was $3$, then we would also have the option to simply
  {\em stay} at $n_g$, i.e., the goal clause would be
  $\{move\mbox{-}n_1\mbox{-}n_g(3), move\mbox{-}n_2\mbox{-}n_g(3),
  NOOP\mbox{-}at\mbox{-}n_g(3)\}$. At $t=2$, this NOOP is not
  present.} We finally get mutex clauses.  The relevant one is $\{\neg
  move\mbox{-}n_0\mbox{-}n_1(1), \neg
  move\mbox{-}n_0\mbox{-}n_2(1)\}$; we cannot move from $n_0$ to $n_1$
  and $n_2$ simultaneously. We also get the clause $\{\neg
  move\mbox{-}n_1\mbox{-}n_g(2), \neg
  move\mbox{-}n_2\mbox{-}n_g(2)\}$, and various mutex clauses between
  move actions and NOOPs. Now, what the relevant clauses say is: 1.\
  We must move to $n_g$ at time $2$, either from $n_1$ or from $n_2$
  (goal). 2.\ If we move from $n_1$ to $n_g$ at time $2$, then we must
  move from $n_0$ to $n_1$ at time $1$ (precondition). 3.\ If we move
  from $n_2$ to $n_g$ at time $2$, then we must move from $n_0$ to
  $n_2$ at time $1$ (precondition). 4.\ We cannot move from $n_0$ to
  $n_1$ and $n_2$ simultaneously at time $1$. Obviously, the
  satisfying assignments correspond exactly to the solution
  paths. Please keep in mind that, in contrast to our example here, in
  general {\em all} the clauses are necessary to obtain a correct
  encoding.
\end{sloppypar}

By making every pair of non-NOOPs incompatible in our simplified
Graphplan-based encoding, we allow at most one (non-NOOP, i.e.,
``real'') action to execute per time step.  This is a restriction in
domains where there exist actions that can be applied in parallel; in
our synthetic domains, no parallel actions are possible anyway, so the
mutex clauses have an effect only on the power of UP (unit
propagation). We also investigated backdoor size, in our synthetic
domains, with a weaker definition of incompatible action pairs,
allowing actions to be applied in parallel unless their effects and
preconditions directly contradict each other.  We omit the results for
the sake of readability.  In a nutshell, one obtains the same DPLL
lower bounds and backdoors in the symmetrical case, but larger
($O(n)$) backdoors and DPLL trees in the asymmetrical case. Note that,
thus, the restrictive definition of incompatible action pairs we use
in our simplified encoding gives us an exponential efficiency
advantage.  This is in line with the use of the Graphplan-based
encoding in our experiments: planning graphs discover the linear
nature of our synthetic domains, including all the mutex clauses
present in our simplified encoding.


%% file: real.tex
\section{Goal Asymmetry in Planning Benchmarks} 
\label{real}

In this section, we explore empirically how $\ASratio$ behaves in a
range of Planning benchmarks. We address two main questions:
\begin{enumerate}[(1)]
\item What is the distribution of $\ASratio$?
\item How does $\ASratio$ behave compared to search performance?
\end{enumerate}
Section~\ref{real:setup} gives details on the experiment
setup. Sections~\ref{real:distribution} and~\ref{real:performance}
address questions (1) and (2), respectively. Before we start, we
reiterate how $\ASratio$ is defined.

\begin{defi}\label{def:ASratio}
  Let $P$ be a planning task with goal $G$. For a conjunction $C$ of
  facts, let $\cost(C)$ be the length of a shortest plan achieving
  $C$.  The {\em asymmetry ratio} of $P$ is:
\[
\ASratio(P) := \frac{max_{g \in G} \cost(g)}{\cost(\bigwedge_{g \in G} g)}
\]
\end{defi}

\noindent
Note that $\cost(\bigwedge_{g \in G} g)$, in this definition, is the
optimal plan length; to simplify notation, we will henceforth denote
this with $m$. Note also that such a simple definition can not be
fool-proof. Imagine replacing $G$ with a single goal $g$, and an
additional action with precondition $G$ and add effect $\{g\}$; the
(new) goal is then no longer a set of ``sub-problems''. However, in
the benchmark domains that are used by researchers to evaluate their
algorithms, $G$ is almost always composed of several goal facts
resembling sub-problems.  A more stable approach to define $\ASratio$
is a topic for future work, outlined in Section~\ref{conclusion}; for
now, we will see that $\ASratio$ often works quite well.

\subsection{Experiment Setup} 
\label{real:setup}

Denote by $\phi(P,b)$, for a planning task $P$ and integer $b$, the
Graphplan-based CNF encoding of $b$ action steps. Our general
intuition is that $\phi(P,m-1)$ is easier to prove unsatisfiable for
tasks $P$ with higher $\ASratio$, than for tasks with lower
$\ASratio$, provided ``all other circumstances are equal''. By this,
we mean that the tasks all share the same {\em domain}, the same {\em
size parameters}, and the same {\em optimal plan length}. We will not
formalize the notions of ``domain'' and ``size parameters''; doing so
is cumbersome and not necessary to understand our experiments. An
informal explanation is as follows:
\begin{enumerate}[$\bullet$]
\item {\bf Domain.} A domain, or {\em STRIPS domain}, is a (typically
infinite) set of related planning tasks. More technically, a domain is
defined through a finite set of logical predicates (relations), a
finite set of {\em operators}, and an infinite set of {\em
instances}. Operators are functions from {\em object tuples} into
STRIPS actions;\footnote{In the planning community, constants are
commonly referred to as ``objects''; we adopt this terminology.} they
are described using predicates and variables. Applying an operator to
an object tuple just means to instantiate the variables, resulting in
a propositional STRIPS action description. Each instance defines a
finite set of objects, an initial state, and a goal condition. Both
initial state and goal condition are sets of instantiated predicates,
i.e., propositional facts.

An example is the MAP domain previewed in Section~\ref{introduction},
Figure~\ref{mapintro}. The predicates are ``edge(x,y)'', ``at(x)'',
and ``visited(x)'' relations; the single operator has the form
``move(x,y)''; the instances define the graph topology and the
initial/goal nodes. A more general example would be a transportation
domain with several vehicles and transportable objects, and additional
operators loading/unloading objects to/from vehicles.

\item {\bf Size parameters.} An instance generator for a domain is
usually parameterized in several ways. In particular, one needs to
specify how many objects of each type (e.g., vehicles, transportable
objects) there will be. For example, $n$ is the size parameter in the
MAP domain. If two instances of a transportation domain both contain
the same numbers of locations, vehicles, and transportable objects,
then they share the same size parameters.
\end{enumerate}
In our experiments, we always fix a domain $\domain$, and a setting
$\size$ of the size parameters of an instance generator for
$\domain$. We then generate thousands of (randomized) instances, and
divide those into classes $\ppp_m$ with identical optimal plan length
$m$. Within each $\ppp_m$, we then examine the distribution of
$\ASratio$, and its behavior compared to performance, in the formulas
$\{\phi(P,m-1) \mid P \in \ppp_m\}$. Note here that the size
parameters and the optimal plan length determine the size of the
formula. So the restriction we make by staying inside the classes
$\ppp_m$ basically comes down to fixing the domain, and fixing the
formula size.

We run experiments in a range of 10 STRIPS domains taken from the
biennial International Planning Competition (IPC) since 2000. The main
criterion for domain selection is the availability of an instance
generator:
\begin{enumerate}[$\bullet$]
\item {\bf IPC 2000}. From this IPC, we select the Blocksworld,
Logistics, and Miconic-ADL domains. Blocksworld is the classical
block-stacking domain using a robot arm to rearrange the positions of
blocks on a table; we use the instance generator by Slaney and
Thiebaux \cite{slaney:thiebaux:ai-01}. Logistics is a basic
transportation domain involving cities, places, trucks, airplanes, and
packages. Each city contains a certain number of places; trucks move
within cities, airplanes between them; packages must be
transported. Moves are instantaneous, i.e., a truck/airplane can move
in one action between any two locations. We implemented a
straightforward instance generator ourselves. Miconic-ADL is an
elevator transport domain coming from a real-world application
\cite{koehler:schuster:aips-00}. It involves all sorts of interesting
side constraints regarding, e.g., the prioritized transportation of
VIPs, and constraints on which people have access to which floors. The
2000 IPC instance generator, which we use, was written by one of the
authors. The domain makes use of first-order logic in action
preconditions; we used the ``adl2strips'' software
\cite{hoffmann:etal:jair-06} to translate this into STRIPS.
\item {\bf IPC 2002}. Here, instance generators are provided by the
organizers \cite{long:fox:jair-03}, so we select all the domains. They
are named Depots, Driverlog, Freecell, Rovers, Satellite, and
Zenotravel.  Depots is a mixture between Blocksworld and Logistics,
where blocks must be transported {\em and} arranged in stacks.
Driverlog is a version of Logistics where trucks need drivers; apart
from this, the main difference to the classical Logistics domain is
that drivers and trucks move on directed graph road maps, rather than
having instant access from any location to any other
location. Freecell encodes the well-known solitaire card game where
the task is to re-order a random arrangement of cards, following
certain stacking rules, using a number of ``free cells'' for
intermediate storage.  Rovers and Satellite are simplistic encodings
of NASA space-applications. In Rovers, rovers move along individual
road maps, and have to gather data about rock or soil samples, take
images, and transfer the data to a lander.  In Satellite, satellites
must take images of objects, which involves calibrating cameras,
turning the right direction, etc.  Zenotravel is a version of
Logistics where moving a vehicle consumes fuel that can be
re-plenished using a ``refuel'' operator.
\item {\bf IPC 2004}. From these domains, the only one for which a
random generator exists is called PSR, where one must reconfigure a
faulty power supply network. PSR is formulated in a complex language
involving first-order formulas and predicates whose value is derived
as an effect of the values of other predicates. (Namely, the flow of
power through the network is determined by the setting of the
switches.) One can translate this to STRIPS, but only through the use
of significant simplification methods \cite{hoffmann:etal:jair-06}. In
the thus simplified domain, every goal can be achieved in a single
step, making $\ASratio$ devoid of information. We emphasize that this
is not the case for the more natural original domain formulation.
\item {\bf IPC 2006}. From these domains, the only one that we can
use is called TPP; for all others, instance generators were not
available at the time of writing. TPP is short for Travelling Purchase
Problem. Given sets of products and markets, and a demand for each
product, one must select a subset of the markets so that routing cost
and purchasing cost are minimized. The STRIPS version of this problem
involves unit-cost products, and a road map graph for the markets.
\end{enumerate}
We finally run experiments in a domain of purely random instances
generated using Rintanen's \cite{rintanen:icaps-04} ``Model A''. We
consider this an interesting contrast to the IPC domains. We will
refer to this domain with the name ``Random''.

To choose the size parameters for our experiments, we simply rely on
the size parameters of the original IPC instances. This is justified
since the IPC instances are the main benchmark used in the
field. Precisely, we use the following method. For each domain, we
consider the original IPC test suite, containing a set of instances
scaling in size. For each instance, we generate a few random instances
with its size parameters, and test how fast we can compute
$\ASratio$. (That computation is done by a combination of calls to
Blackbox.) We select the largest instance for which each test run is
completed within a minute. For example, in Driverlog we select the
instance indexed 10 out of 20, and, accordingly, generate random
instances with 6 road junctions, 2 drivers, 6 transportable objects,
and 3 trucks. While the instances generated thus are not at the very
limit of the performance of SAT-based planners, they are reasonably
close to that limit. The runtime curves of Blackbox undergo a sharp
exponential increase in all the considered domains. There is not much
room between the last instance solved in a minute, and the first
instance not solved at all. For example, in Driverlog we use instance
number 10, while the performance limit of Blackbox is reached around
instance number 12.\footnote{In Logistics, we use IPC instance number
18 of 50, and the performance limit is around instance number 20. In
Miconic-ADL, we use IPC instance 7 out of 30, and the performance
limit is around instance 9. In Depots, these numbers are 8, 20, and
10; in Rovers, they are 8, 20, and 10; in Satellite, they are 7, 20,
and 9; in Zenotravel, they are 13, 20, and 15; in TPP, they are 15,
30, and 20. In Blocksworld, we use 11 blocks and the performance limit
is at around 13 blocks. In Freecell, Blackbox is very inefficient,
reaching its limit around IPC instance 2 (of 20), which is what we
use.} In the random domain, we use ``$40$ state variables'' which is
reasonable according to Rintanen's \cite{rintanen:icaps-04}
results. We also run a number of experiments where we modify the size
parameters in certain domains, with the aim to obtain a picture of
what happens as the parameters change. This will be detailed below.

\vspace{-0.0cm}
\begin{figure*}
  \begin{center}
    \begin{tabular}{cc}
\includegraphics[width=7.4cm]{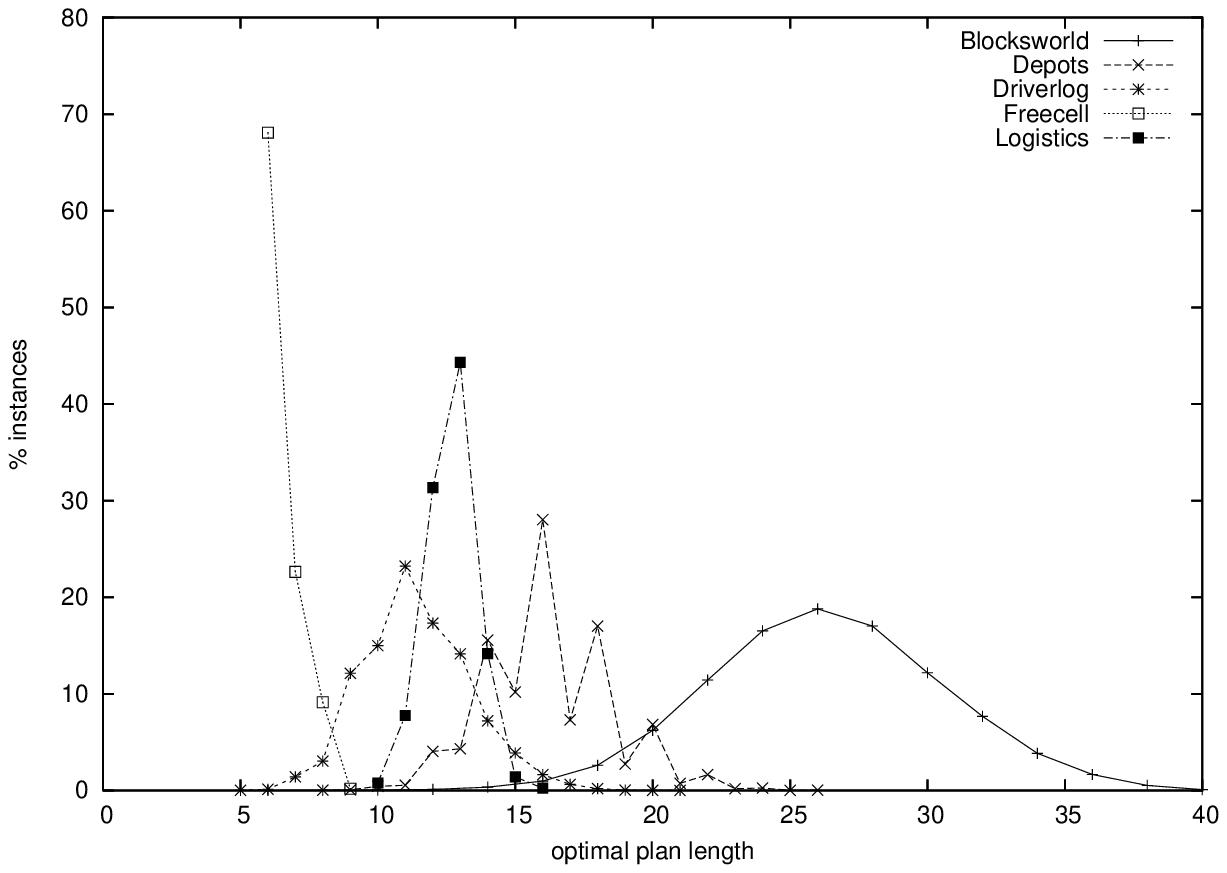} & \includegraphics[width=7.4cm]{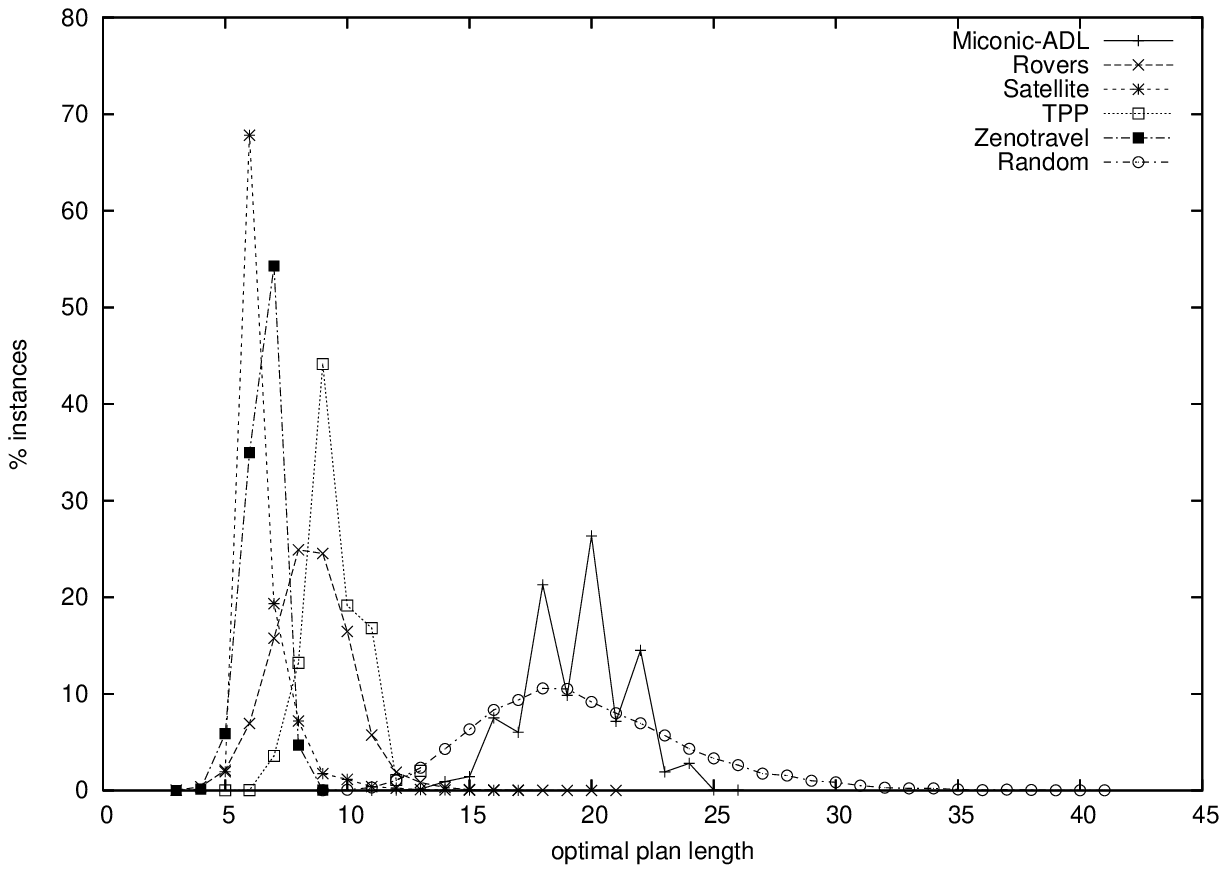}\\
 (a) IPC instances & (b) IPC instances and Random\\
    \end{tabular}
  \end{center}
  \vspace{-0.0cm}
  \caption{\label{gnu:length}Distribution of optimal plan
  length. Split across two graphs for readability.}
  \vspace{-0.5cm}
\end{figure*}

In most cases, the distribution of optimal plan length $m$ is rather
broad, for given domain $\domain$ and size parameters $\size$; see
Figure~\ref{gnu:length}.\footnote{We see that the plan length is
mostly normal distributed. The only notable exception is the Freecell
domain, Figure~\ref{gnu:length} (a), where the only plan lengths
occurring are 6, 7, 8, and 9, and $\ppp_6$ is by far the most populated
class.} Thus we need many instances in order to obtain reasonably
large classes $\ppp_m$. Further, we split each class $\ppp_m$ into
subsets, {\em bins}, with identical $\ASratio$. To make the bins
reasonably large, we need even more instances. In initial experiments,
we found that between 5000 and 20000 instances were usually sufficient
to identify the overall behavior of the domain and size parameter
setting ($\domain$ and $\size$). To be conservative, we decided to fix
the number of instances to 50000, per $\domain$ and $\size$.  To avoid
noise, we skip bins with less than 100 elements; the remaining bins
each contain around a few 100 up to a few 1000 instances.

\subsection{AsymRatio Distribution} 
\label{real:distribution}

\begin{figure*}
  \begin{center}
    \begin{tabular}{cc}
\includegraphics[width=7.4cm]{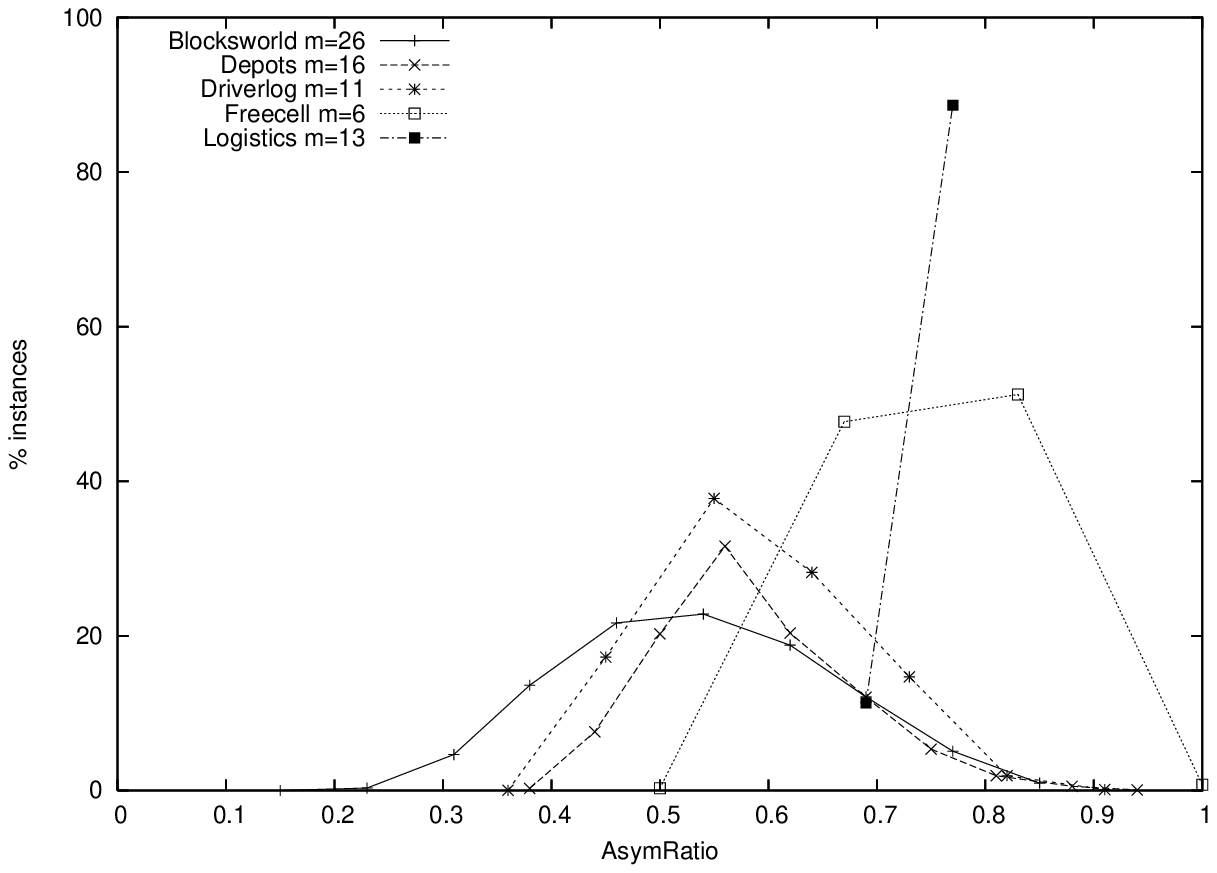} & \includegraphics[width=7.4cm]{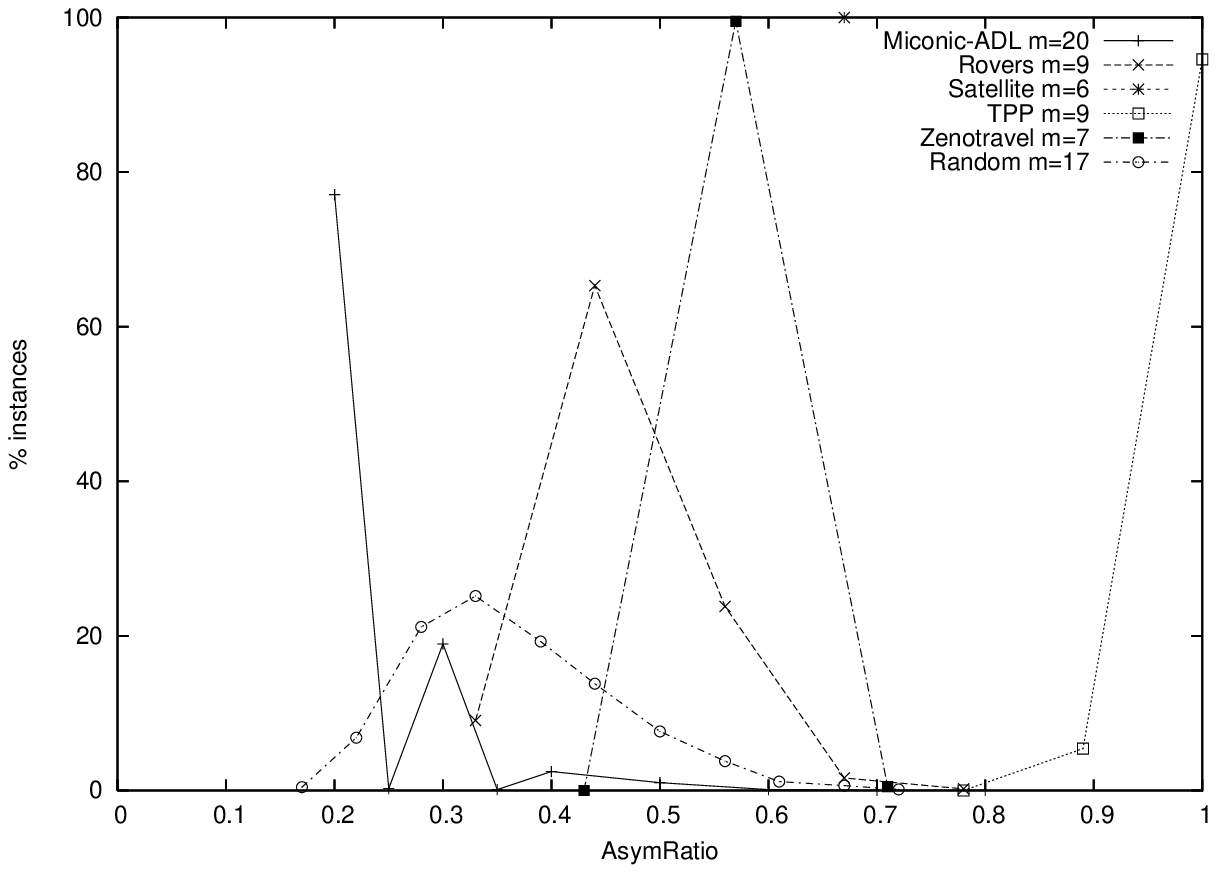}\\\vspace{0.3cm}
 (a) IPC instances & (b) IPC instances and Random\\
\includegraphics[width=7.4cm]{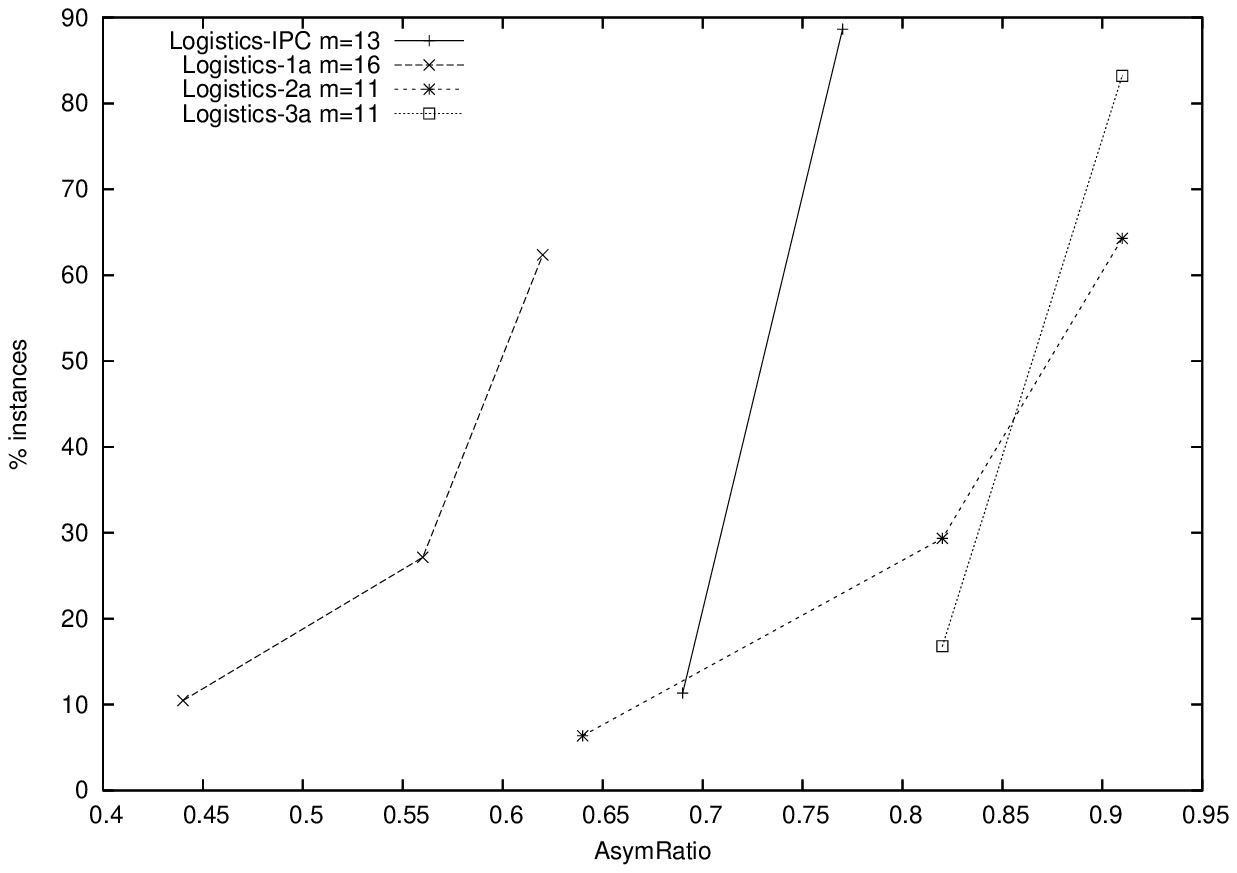} & \includegraphics[width=7.4cm]{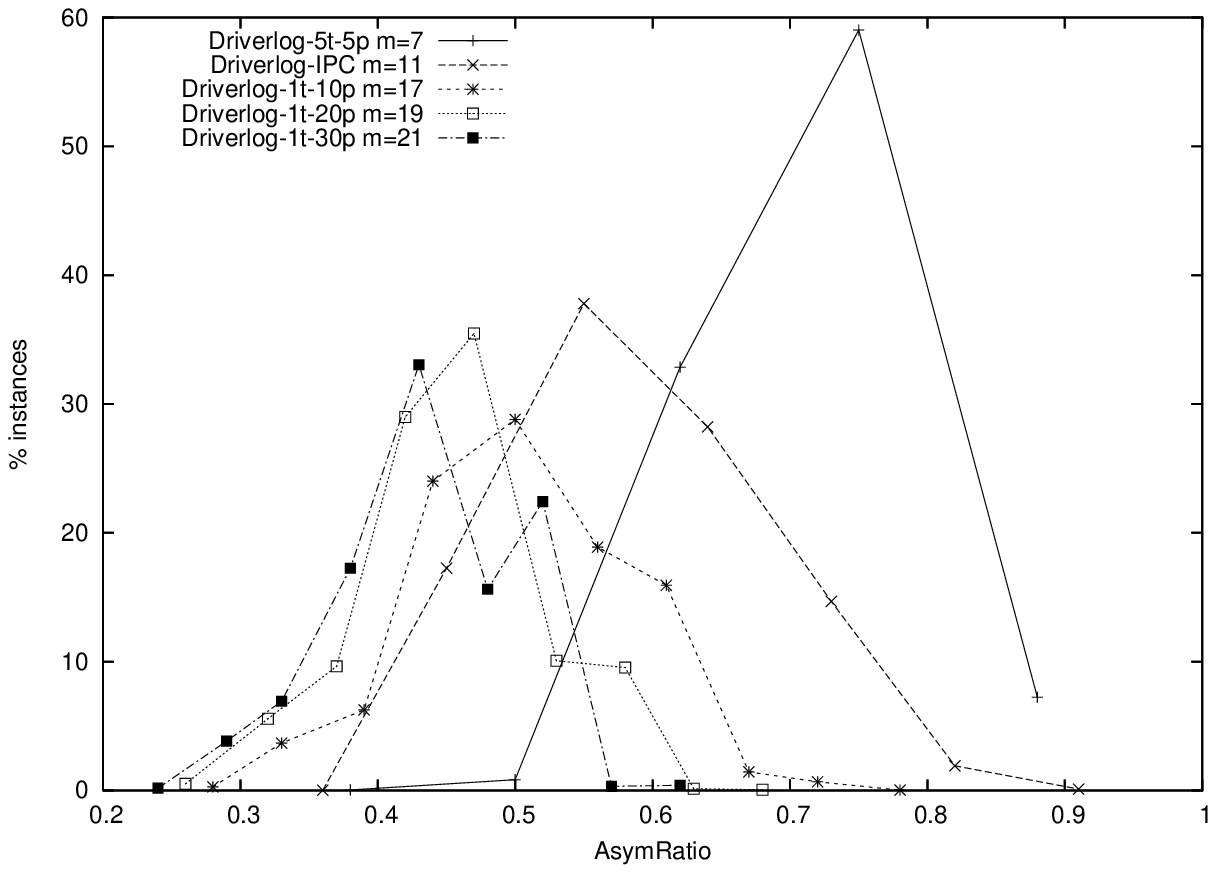}\\\vspace{0.3cm}
 (c) Logistics & (d) Driverlog \\
\end{tabular}
    \begin{tabular}{c}
\includegraphics[width=7.4cm]{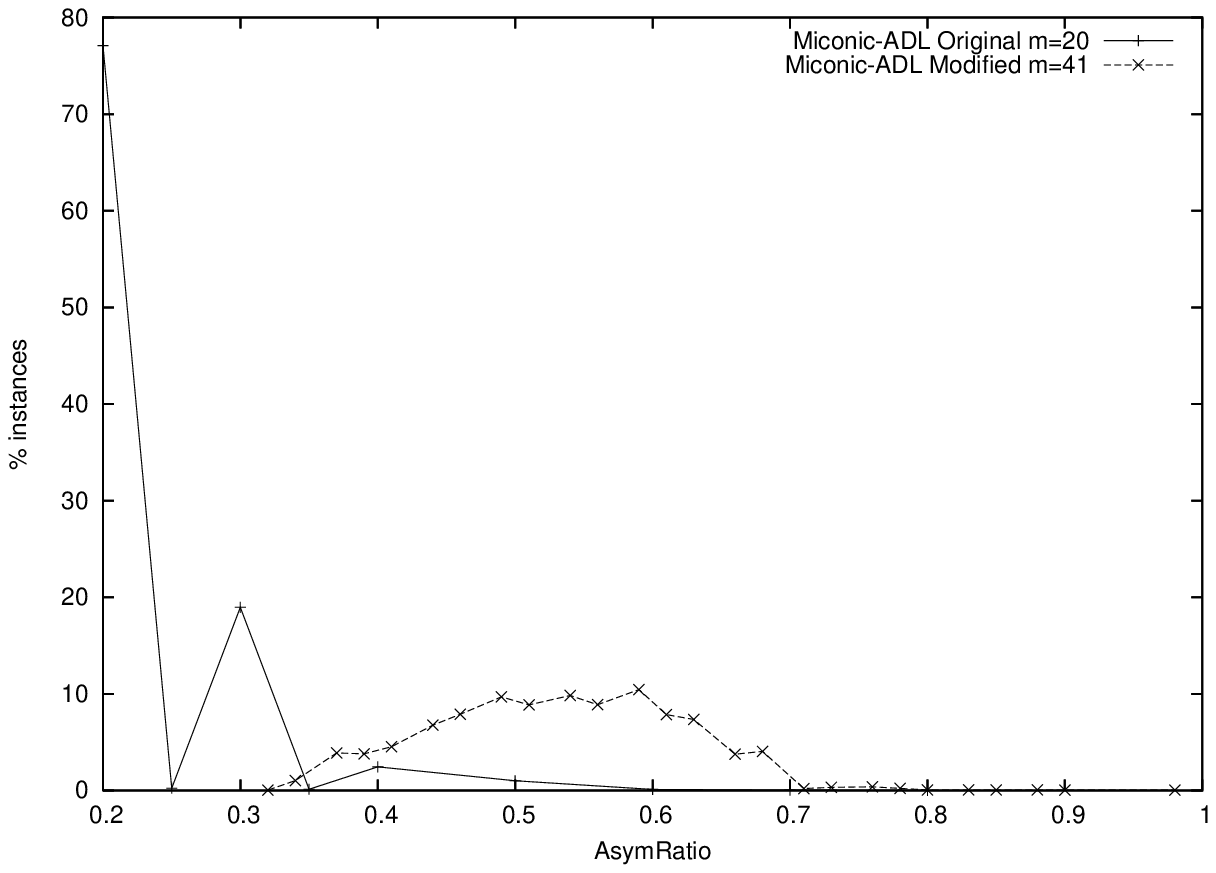}\\\vspace{0.3cm}
 (e) Miconic-ADL \\
    \end{tabular}
  \end{center}
  \vspace{-0.0cm}
  \caption{\label{gnu:asratio}Distribution of $\ASratio$, for the
  respective most populated class $\ppp_m$ as indicated. (a) and (b),
  for IPC settings of $\size$. (c), for various settings of $\size$ in
  Logistics. (d), for various settings of $\size$ in Driverlog. (e),
  for two variants of Miconic-ADL. Explanations see text.}
  \vspace{-0.5cm}
\end{figure*}

What interests us most in examining the distribution of $\ASratio$ is
how ``spread out'' the distribution is, i.e., how many different
values we obtain within the $\ppp_m$ classes, and how far they are
apart. The more values we obtain, the more cases can we distinguish
based on $\ASratio$; the farther the values are apart, the more
clearly will those cases be distinct. Figure~\ref{gnu:asratio} shows
some of the data; for all the settings of $\domain$ and $\size$ that
we explored, it shows the $\ASratio$ distribution in the most
populated class $\ppp_m$. 


\vspace{-0.0cm}

In Figure~\ref{gnu:asratio}, the $x$ axes show $\ASratio$, and the $y$
axes show percentage of instances within $\ppp_m$. Let us first
consider Figure~\ref{gnu:asratio} (a) and (b), which show the plots
for the IPC settings $\size$ of the size parameters. In
Figure~\ref{gnu:asratio} (a), we see vaguely normal distributions for
Blocksworld, Depots, and Driverlog. Freecell is unusual in its large
weight at high $\ASratio$ values, and also in having relatively few
different $\ASratio$ values. This can be attributed to the unusually
small plan length in Freecell.\footnote{Remember that our optimal
planner scales only to instance number 2 out of 20 in the IPC 2002
test suite.} For Logistics, we note that there are only 2 different
values of $\ASratio$. This can be attributed to the trivial road maps
in this domain, where one can instantaneously move between any two
locations.  The instance shown in Figure~\ref{gnu:asratio} has 6
cities and 2 airplanes, and thus there is not much variance in the
cost of achieving a single goal (transporting a single package). This
will be explored further below.

In Figure~\ref{gnu:asratio} (b), at first glance one sees that there
are quite a few odd-looking distributions. The distributions for
Rovers and Random are fairly spread out. Miconic-ADL has $77\%$ of its
weight at $\ASratio=0.20$, $18\%$ of its weight at $\ASratio=0.30$,
and $2.4\%$ at $\ASratio=0.4$; a lot of other $\ASratio$ values have
non-zero weights of less than $1\%$. This distribution can be
attributed to the structure of the domain, where the elevator can
instantaneously move between any two floors; most of the time, each
goal (serving a passenger) thus takes $4$ steps to achieve,
corresponding to $\ASratio=0.25$ in our picture. Due to the side
constraints, however, sometimes serving a passenger involves more
effort (e.g., some passengers need to be attended in the lift) --
hence the instances with higher $\ASratio$ values. We will consider
below a Miconic-ADL variant with non-instantaneous lift moves.

\begin{sloppypar}
Zenotravel has an extreme $\ASratio$ distribution, with $99\%$ of the
weight at $\ASratio=0.57$. Again, this can be attributed to the lack
of a road map graph, i.e., to instantaneous vehicle moves. In TPP,
$94\%$ of the time there is a single goal that takes as many steps to
achieve as the entire goal set. It is unclear to us what the reason
for that is (the goal sets are large, containing 10 facts in all
cases). The most extreme $\ASratio$ distribution is exhibited by
Satellite: in all classes $\ppp_m$ in our experiments, all instances
have the same $\ASratio$ value ($\ASratio=0.67$ in
Figure~\ref{gnu:asratio} (b)). Partly, once again this is because of
instantaneous ``moves'' -- here, changing the direction a satellite
observes. Partly it seems to be because of the way the 2002 IPC
Satellite instance generator works, most of the time including at
least one goal with the same maximal cost.
\end{sloppypar}

To sum the above up, $\ASratio$ does show a considerable spread of
values across all our domains except Logistics, Satellite, TPP, and
Zenotravel. For Logistics, Satellite, and Zenotravel, this can be
attributed to the lack of road map graphs in these domains. This is a
shortcoming of the domains, rather than a shortcoming of $\ASratio$;
in reality one does not move instantaneously (though sometimes one
would wish to \dots). We ran additional experiments in Logistics to
explore this further. We generated instances with 8 instead of 6
cities, and with just a single airplane instead of 2 airplanes; we had
to decrease the number of packages from 18 to 8 to make this
experiment feasible. We then repeated the experiment, but with the
number of airplanes set to 2 and 3,
respectively. Figure~\ref{gnu:asratio} (c) shows the results,
including the IPC $\size$ setting for comparison. ``Logistics-$x$a'
denotes our respective new experiment with $x$ airplanes. Considering
these plots, we see clearly that the $\ASratio$ distribution shifts to
the right, and becomes more dense, as we increase the number of
airplanes; eventually (with more and more airplanes) all weight will
be in a single spot.


In Driverlog, we run experiments to explore what happens as we change
the relative numbers of vehicles and transportable
objects. Figure~\ref{gnu:asratio} (d) shows the results. The IPC
instance has 6 road junctions, 2 drivers, 6 transportable objects, and
3 trucks. For ``Driverlog-1t-10p'' in Figure~\ref{gnu:asratio} (d), we
set this to 6 road junctions, 1 driver, 10 transportable objects, and
1 truck. For ``Driverlog-1t-20p'' and ``Driverlog-1t-30p'', we
increased the number of transportable objects to 20 and 30,
respectively. ``Driverlog-5t-5p'', on the other hand, has 6 road
junctions, 5 drivers, 5 transportable objects, and 5 trucks. The
intuition is that the position of the $\ASratio$ distribution depends
on the ratio between the number of transportable objects and the
number of means for transport. The higher that ratio is, the lower do
we expect $\ASratio$ to be -- if there are many objects for a single
vehicle then it is unlikely that a single object will take all the
time. Also, for very low and very high values of the ratio we expect
the distribution of $\ASratio$ to be dense -- if every object takes
almost all the time/no object takes a significant part of the time,
then there should not be much variance. Figure~\ref{gnu:asratio} (d)
clearly shows this tendency for ``Driverlog-5t'' (ratio = $5/5$),
``Driverlog-IPC'' (ratio = $6/3$), ``Driverlog-10p'' (ratio = $10/1$),
and ``Driverlog-20p'' (ratio = $20/1$). Interestingly, the step from
``Driverlog-20p'' to ``Driverlog-30p'' does not make as much of a
difference. Still, if we keep increasing the number of transportable
objects then eventually the distribution will trivialize. Note here
that 30 objects in a road map with only 6 nodes already constitute a
rather dense transportation problem. Further, transport problems with
30 objects really push the limits of the capabilities of current SAT
solvers.

Let us finally consider Figure~\ref{gnu:asratio} (e). ``Miconic-ADL
Modified'' denotes a version of Miconic-ADL where the elevator is
constrained to take one move action for every move between adjacent
floors, rather than moving between any two floors instantaneously. As
one would expect, this changes the distribution of $\ASratio$
considerably. We get a distribution spreading out from $\ASratio=0.34$
to $\ASratio=0.75$, and further -- with very low percentages -- to
$\ASratio=0.97$.

\subsection{AsymRatio and Search Performance}
\label{real:performance}

Herein, we examine whether $\ASratio$ is indeed an indicator for
search performance. Let us first state our hypothesis more
precisely. Fixing a domain $\domain$, a size parameter setting
$\size$, and a number $\delta \in [0;1]$, our hypothesis is:

\begin{hypothesis}[$\domain, \size, \delta$]\label{hypo:ASratio}
  Let $m$ be an integer, and let $x, y \in [0;1]$, where $y - x >
  \delta$. Let $\ppp_m$ be the set of instances in $\domain$ with size
  $\size$ and optimal plan length $m$. Let $\ppp_m^x = \{P \in \ppp_m
  \mid \ASratio(P) = x\}$ and $\ppp_m^y = \{P \in \ppp_m \mid
  \ASratio(P) = y\}$.

  If both $\ppp_m^x$ and $\ppp_m^y$ are non-empty, then the mean DPLL
  search tree size is significantly higher for $\{\phi(P,m-1) \mid P
  \in \ppp_m^x\}$ than for $\{\phi(P,m-1) \mid P \in \ppp_m^y\}$.
\end{hypothesis}

The hypothesis is parameterized by $\domain$, $\size$, and by an {\em
$\ASratio$ difference threshold $\delta$}. Regarding $\domain$, we do
not mean to claim that the stated correlation will be true for every
domain; we construct a counter example in Section~\ref{strawman}, and
we will see that Hypothesis~\ref{hypo:ASratio} is not well supported
by two of the domains in our experiments. Regarding $\size$, we have
already seen that the size parameter setting influences the
distribution of $\ASratio$; we will see that this is also true for the
behavior of $\ASratio$ compared to performance. Regarding the
$\ASratio$ difference threshold $\delta$, this serves to trade-off the
strength of the claim vs its applicability. In other words, one can
make Hypothesis~\ref{hypo:ASratio} weaker by increasing $\delta$, at
the cost of discriminating between less classes of instances. We will
explore different settings of $\delta$ below.

For every planning task $P$ generated in our experiments, with optimal
plan length $m$, we run \zchaff \cite{moskewicz:etal:dac-01} on the
formula $\phi(P,m-1)$. We measure the search tree size (number of
backtracks) as our indication of how hard it is to prove a formula
unsatisfiable. We compare the distributions of search tree size
between pairs of $\ASratio$ bins within a class $\ppp_m$. Some
quantitative results will be discussed below. First, we show
qualitative results summarizing how often the distributions differ
significantly. For every $\domain$ and $\size$, we distinguish between
three different values of $\delta$; these will be explained below. For
every class $\ppp_m$, and for every pair $\ppp_m^x$, $\ppp_m^y$ of
$\ASratio$ bins with $y - x > \delta$, we run a T-test to determine
whether or not the mean search tree sizes differ
significantly. Namely, we run the Student's T-test for unequal sample
sizes, and check whether the means differ with a confidence of $95\%$
and/or a confidence of $99.9\%$. Table~\ref{tab:ttest} summarizes the
results.

\begin{table}
\begin{center}
{\small
\begin{tabular}{|l||r|r||r|r|r|r||r|r|r|r|}\hline
           &   \multicolumn{2}{|c||}{$\delta=0$}  &   \multicolumn{4}{|c||}{$\delta_{95}(\domain,\size)$} & \multicolumn{4}{|c|}{$\delta_{100}(\domain,\size)$} \\
Experiment      & $95$ & $99.9$ & $\delta$ & \# & $95$ & $99.9$ & $\delta$ & \# & $95$ & $99.9$ \\\hline\hline
Blocksworld     &  94 &  87 & 0.07 &  84 &  95 &  90 & 0.15 &  55 & 100 &  97\\\hline
Depots          &  84 &  78 & 0.08 &  70 &  95 &  90 & 0.16 &  39 & 100 &  98\\\hline
Driverlog-IPC   &  97 &  94 & 0.00 & 100 &  97 &  94 & 0.06 &  92 & 100 &  98\\\hline
Freecell        & 100 & 100 & 0.00 & 100 & 100 & 100 & 0.00 & 100 & 100 & 100\\\hline
Logistics-IPC   &  16 &  16 & 0.28 &   0 &     &     & 0.28 &   0 &     &    \\\hline
Miconic-ADL Org &  85 &  70 & 0.15 &  29 & 100 &  87 & 0.15 &  29 & 100 &  87\\\hline
Rovers          &  96 &  96 & 0.00 & 100 &  96 &  96 & 0.06 &  96 & 100 & 100\\\hline
TPP             & 100 & 100 & 0.00 & 100 & 100 & 100 & 0.00 & 100 & 100 & 100\\\hline
Zenotravel      & 100 &  50 & 0.00 & 100 & 100 &  50 & 0.00 & 100 & 100 &  50\\\hline\hline
Miconic-ADL Mod &  86 &  74 & 0.04 &  77 &  97 &  91 & 0.09 &  49 & 100 & 100\\\hline\hline
Random          &  40 &  27 & 0.24 &   5 & 100 &  71 & 0.24 &   5 & 100 &  71\\\hline\hline
Logistics-1a    &  91 &  83 & 0.08 &  62 & 100 & 100 & 0.08 &  62 & 100 & 100\\\hline
Logistics-2a    &  77 &  77 & 0.12 &  55 & 100 & 100 & 0.12 &  55 & 100 & 100\\\hline
Logistics-3a    &  66 &  66 & 0.12 &  33 & 100 & 100 & 0.12 &  33 & 100 & 100\\\hline\hline
Driverlog-5t-5p &  90 &  90 & 0.15 &  50 & 100 & 100 & 0.15 &  50 & 100 & 100\\\hline
Driverlog-1t-10p&  93 &  86 & 0.06 &  72 &  96 &  94 & 0.11 &  53 & 100 &  98\\\hline
Driverlog-1t-20p&  88 &  84 & 0.10 &  49 &  96 &  96 & 0.13 &  33 & 100 & 100\\\hline
Driverlog-1t-30p&  80 &  75 & 0.11 &  37 &  95 &  90 & 0.13 &  28 & 100 &  94\\\hline
\end{tabular}}
\end{center}
\vspace{-0.0cm}
\caption{\label{tab:ttest}Summary of T-test results comparing the
distributions of \zchaff's search tree size for every pair $\ppp_m^x$,
$\ppp_m^y$ of $\ASratio$ bins within every class $\ppp_m$, for every
$\domain$ and $\size$, and different settings of $\delta$; explanation
see text. The ``$\delta$'' columns give the value of $\delta$; the
``\#'' columns give the percentage of pairs with $y - x > \delta$; the
``95'' and ``99.9'' columns give the percentage of pairs whose
distribution means differ significantly as hypothesized, at the
respective level of confidence.}
\vspace{-0.5cm}
\end{table}

The three different values of $\delta$ distinguished for every
$\domain$ and $\size$ are $\delta=0$, and two values called
$\delta_{95}(\domain,\size)$ and
$\delta_{100}(\domain,\size)$. $\delta=0$ serves to show the situation
for all pairs; $\delta_{95}(\domain,\size)$ and
$\delta_{100}(\domain,\size)$ show how far one has to increase
$\delta$ in order to obtain $95\%$ and $100\%$ accuracy of
Hypothesis~\ref{hypo:ASratio}, respectively. Precisely,
$\delta_{95}(\domain,\size)$ is the smallest number $\delta \in \{0,
0.01, 0.02, \dots, 1.0\}$ so that the T-test succeeds for at least
$95\%$ of the pairs with $y - x > \delta$, with confidence level
$95\%$. Similarly, $\delta_{100}(\domain,\size)$ is the smallest
number $\delta \in \{0, 0.01, 0.02, \dots, 1.0\}$ so that the T-test
succeeds for all the pairs with $y - x > \delta$, with confidence
level $95\%$. Note that, for some $\domain$ and $\size$,
$\delta_{95}(\domain,\size) = \delta_{100}(\domain,\size)$, or $0 =
\delta_{95}(\domain,\size)$, or even $0 = \delta_{95}(\domain,\size) =
\delta_{100}(\domain,\size)$.

The upper most part of Table~\ref{tab:ttest} is for the IPC domains
and the respective settings of $\size$. Satellite is left out because
not a single class $\ppp_m$ in this domain contained more than one
$\ASratio$ bin. Then there are separate parts for the modified version
of Miconic-ADL, for the Random domain, and for our exploration of
different $\size$ settings in Logistics and Driverlog.

One quick way to look at the data is to just examine the leftmost
columns, where $\delta = 0$. We see that there is good support for
Hypothesis~\ref{hypo:ASratio}, with the T-test giving a $95\%$
confidence level in the vast majority of cases (pairs of $\ASratio$
bins). Note that most of the T-tests succeed with a confidence level
of $99.9\%$. Logistics-IPC and Random, and to some extent
Logistics-3a, are the only experiments (rows of Table~\ref{tab:ttest})
that behave very differently. In Freecell, TPP, and Zenotravel, there
are only few $\ASratio$ pairs, c.f.\ Figure~\ref{gnu:asratio};
precisely, we got 9 pairs in Freecell, 4 pairs in TPP, and only 2
pairs in Zenotravel. For all of these pairs, the T-test succeeds, with
$99.9\%$ confidence in all cases except one of the pairs in
Zenotravel.

Another way to look at the data is to consider the values of
$\delta_{95}(\domain,\size)$ and $\delta_{100}(\domain,\size)$, for
each experiment. The smaller these values are, the more support do we
have for Hypothesis~\ref{hypo:ASratio}. In particular, at
$\delta_{100}(\domain,\size)$, every pair of $\ASratio$ bins
encountered within 50000 random instances behaves as
hypothesized. Except in Logistics-IPC and Random, the maximum
$\delta_{100}(\domain,\size)$ we get in any of our experiments is the
$\delta=0.16$ needed for Depots. For some of the other experiments,
$\delta_{100}(\domain,\size)$ is considerably lower; the mean over the
IPC experiments is $9.55$, the mean over all experiments is
$11.27$. The mean value of $\delta_{95}(\domain,\size)$ over the IPC
experiments is $6.44$ ($3.75$ without Logistics-IPC), the mean over
all experiments is $8.88$.

When considering $\delta_{95}(\domain,\size)$ and
$\delta_{100}(\domain,\size)$, we must also consider the relative
numbers of $\ASratio$ pairs that actually remain given these $\delta$
values.  This information is provided in the ``\#'' columns in
Table~\ref{tab:ttest}.  We can nicely observe how increasing $\delta$
trades off the accuracy of Hypothesis~\ref{hypo:ASratio} vs its
applicability. In particular, we see that no pairs remain in
Logistics-IPC, and that hardly any pairs remain in Random; for these
settings of $\domain$ and $\size$, the only way to make
Hypothesis~\ref{hypo:ASratio} ``accurate'' is by raising $\delta$ so
high that it excludes almost all pairs. So here $\ASratio$ is
useless. The mean percentage of pairs remaining at
$\delta_{95}(\domain,\size)$ is $75.88$ for the IPC experiments
($85.37$ without Logistics-IPC), and $60.16$ for all experiments.  The
mean percentage of pairs remaining at $\delta_{100}(\domain,\size)$ is
$67.88$ for the IPC experiments ($76.37$ without Logistics-IPC), and
$54.38$ for all experiments.

As expected, in comparison to the original version of Miconic-ADL,
$\ASratio$ is more reliable in our modified version, where lift moves
are not instantaneous between any pair of floors. It is unclear to us
what the reason for the unusally bad results in Logistics and Random
is. A general speculation is that, in these domains, even if there is
a goal that takes relatively many steps to achieve, this does not
imply tight constraints on the solution. More concretely, regarding
Logistics, consider a transportation domain, and a goal that takes
many steps to achieve because it involves travelling a long way on a
road map. Then the number of options to achieve the goal corresponds
to the number of optimal paths on the map. Unless the map is densely
connected, this will constrain the search considerably. In Logistics,
however, the map is actually fully connected. So, (A), there is not
much variance in how many steps a goal needs; and, (B), it is not a
tight constraint to force one of the airplanes to move from one
location to another at a certain time step, particularly if there are
many airplanes. Given (A), it is surprising that in Logistics-1a
$\ASratio$ is as good an indicator of performance as in most other
experiments. Given (B), it is understandable why this phenomenon gets
weaker for Logistics-2a and Logistics-3a. Logistics-IPC is even
tougher than Logistics-3a, presumably due to the larger ratio between
number of packages and road map size.

In comparison to Logistics, the results for Driverlog-5t-5p show the
effect of a non-trivial road map. Even with 5 trucks for just 5
transportable objects, $\ASratio$ is a good performance
indicator. Increasing the number of transportable objects over
Driverlog-1t-10p, Driverlog-1t-20p, and Driverlog-1t-30p, $\ASratio$
becomes less reliable. Note, however, that stepping from
Driverlog-1t-10p to Driverlog-1t-20p affects the behavior more than
stepping from Driverlog-1t-20p to Driverlog-1t-30p. In the former
step, $\delta_{95}(\domain,\size)$ and $\delta_{100}(\domain,\size)$
increase by $0.04$ and $0.02$, respectively; in the latter step,
$\delta_{95}(\domain,\size)$ increases by $0.01$ and
$\delta_{100}(\domain,\size)$ remains the same. Also, we reiterate
that 30 objects on a map with 6 nodes -- Driverlog-1t-30p -- already
constitute a very dense transportation problem, and that optimal
planners do not scale beyond 30 objects anyway.\footnote{In fact, we
were surprised that Driverlog instances with 30 objects can be
solved. If one increases the number of trucks (state space branching
factor) or the number of map nodes (branching factor and plan length)
only slightly, the instances become extremely challenging; even with
20 packages and 10 map nodes, \zchaff\ often takes hours.}

To sum our observations up, on the negative side there are domains
like Random where Hypothesis~\ref{hypo:ASratio} is mostly wrong, and
there are domains like Logistics where it holds only for very
restricted settings of $\size$; in other domains, $\ASratio$ is
probably devoid in extreme $\size$ settings. On the positive side,
Hypothesis~\ref{hypo:ASratio} holds in almost all cases -- all pairs
of $\ASratio$ bins -- we encountered in the IPC domains (other than
Logistics) and IPC parameter size settings. With increasing $\delta$,
the bad cases quickly disappear; in our experiments, all bad cases are
filtered out at $\delta = 0.28$, and all bad cases except those in
Logistics and Random are filtered out at $\delta = 0.16$.

It would be interesting to explore how the behavior of $\ASratio$
changes as a function of all the parameters of the domains, i.e., to
extend our above observations regarding Logistics and Driverlog, and
to perform similar studies for all the other domains. Given the number
of parameters the domains have, such an investigation is beyond the
scope of this paper, and we leave it as a topic for future work.

\vspace{-0.0cm}
\begin{figure*}
  \begin{center}
    \begin{tabular}{cc}
      \includegraphics[width=7.4cm]{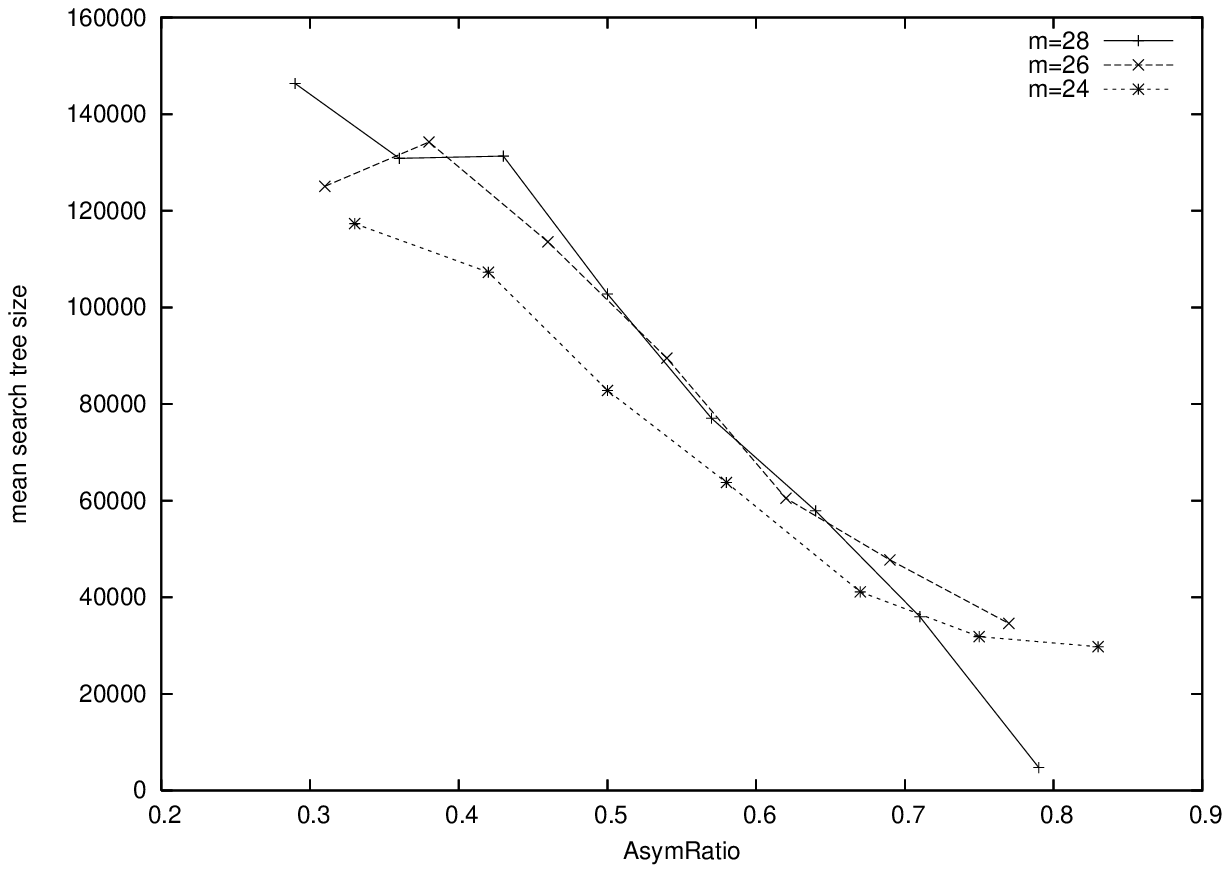} & \includegraphics[width=7.4cm]{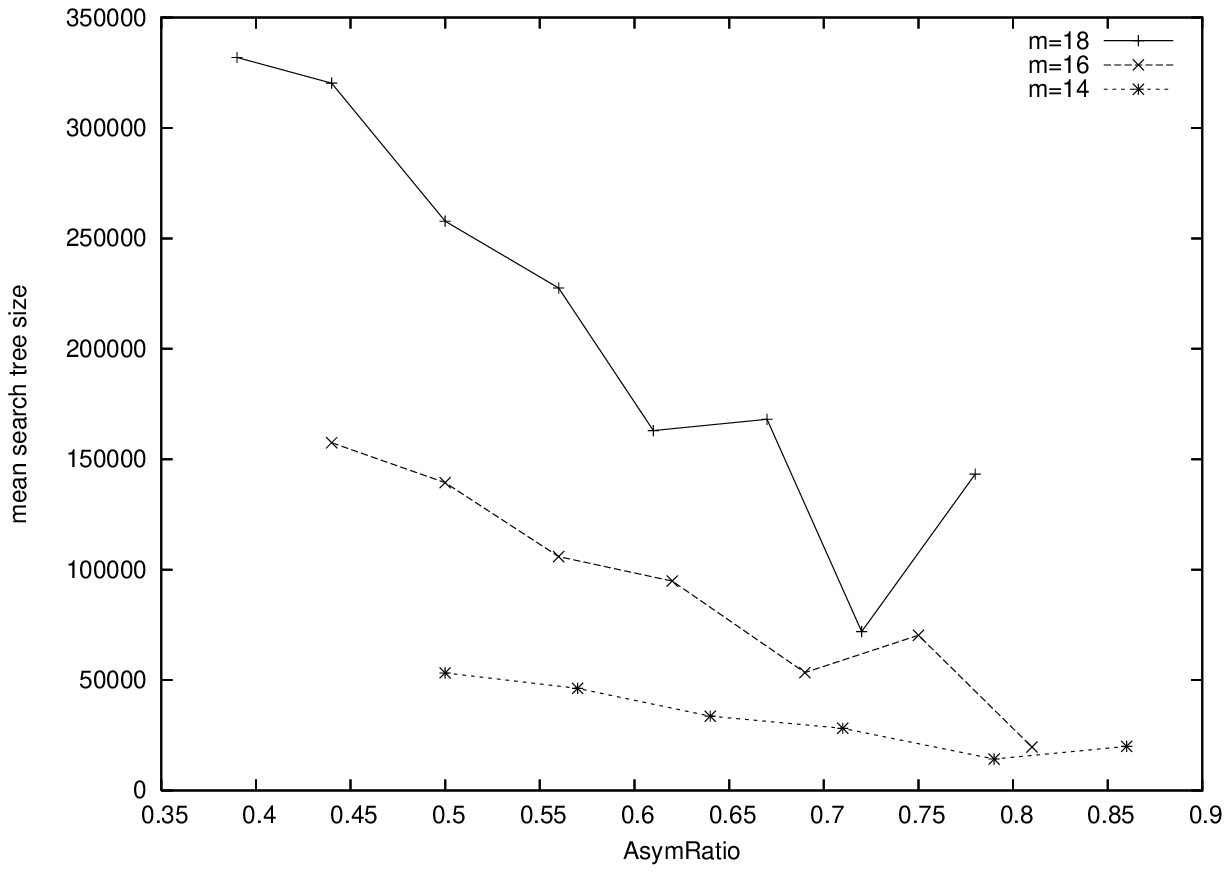}\\\vspace{0.3cm}
      (a) Blocksworld & (b) Depots\\
     \includegraphics[width=7.4cm]{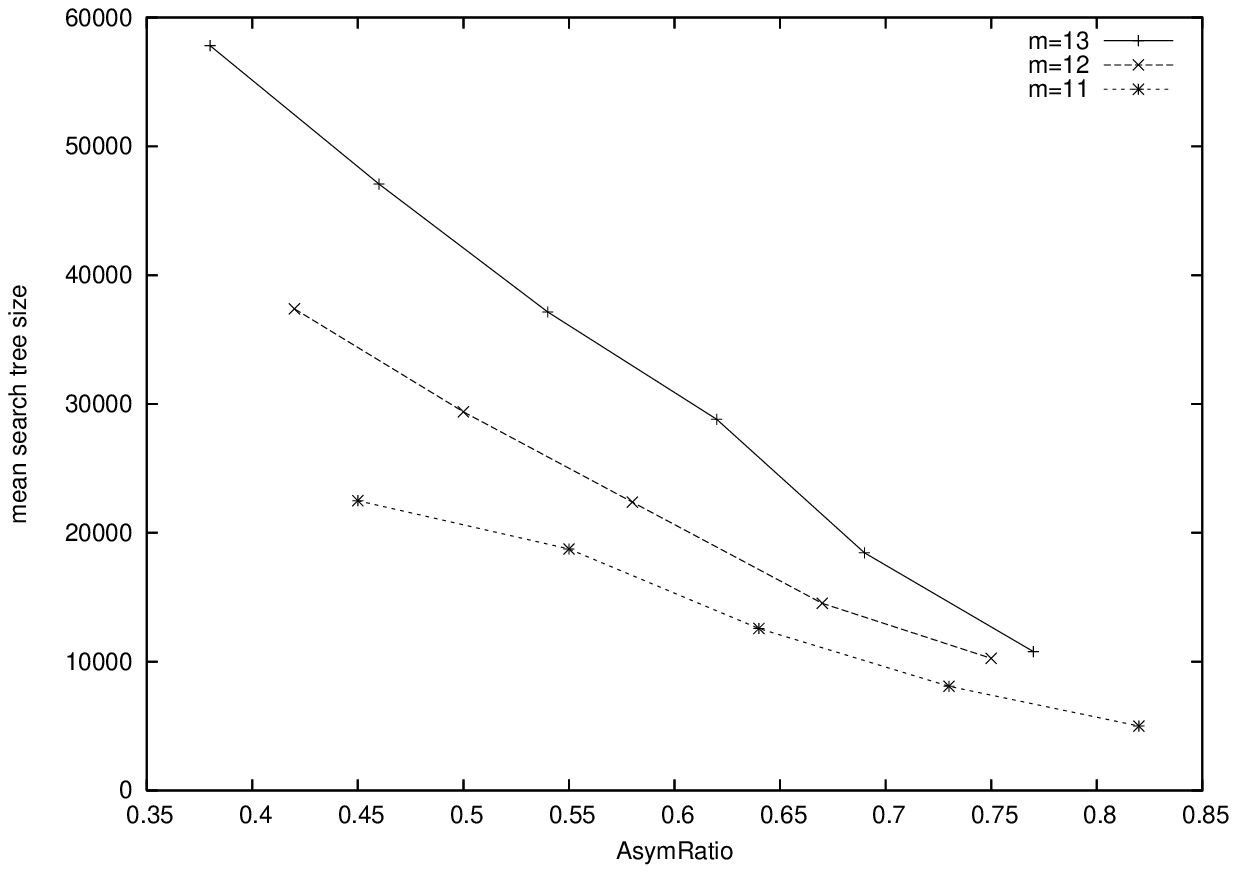} & \includegraphics[width=7.4cm]{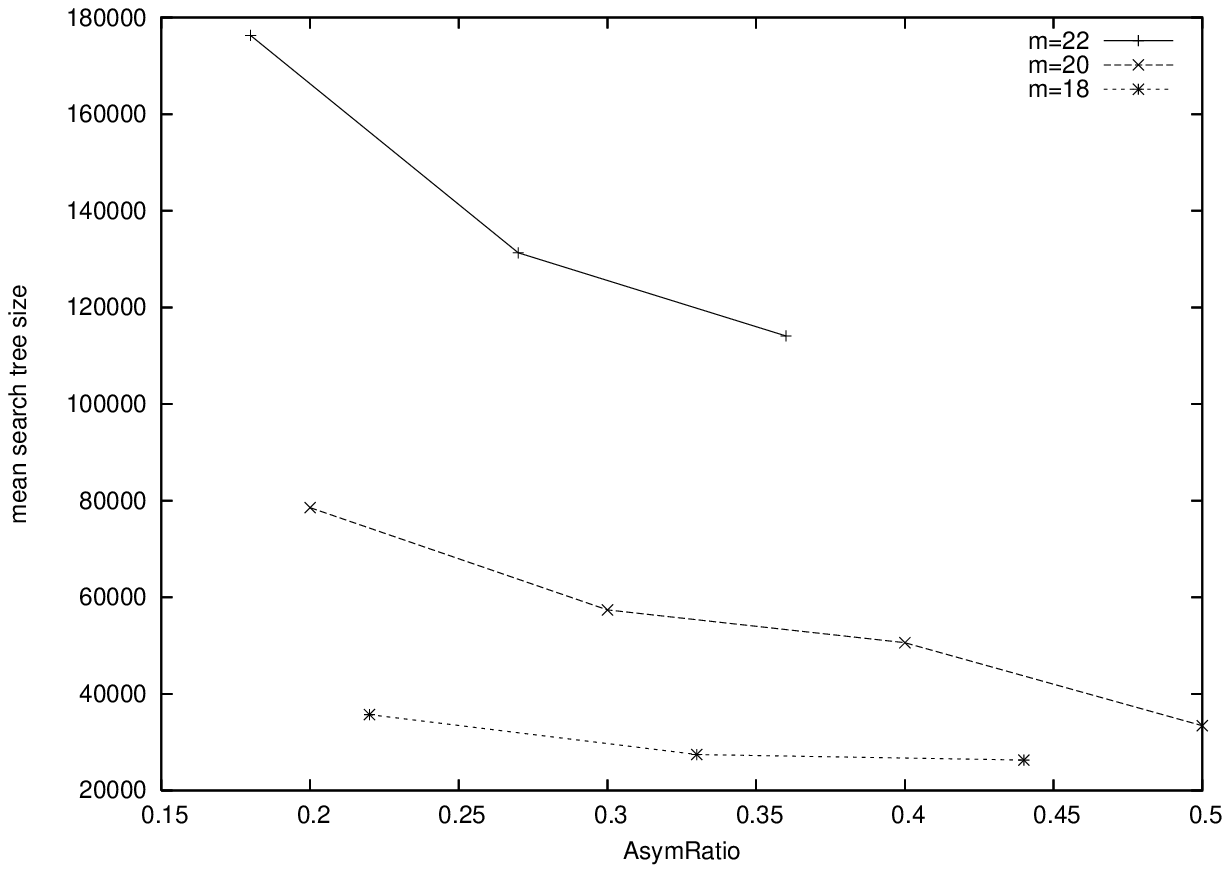}\\\vspace{0.3cm}
      (c) Driverlog & (d) Miconic-ADL\\
    \includegraphics[width=7.4cm]{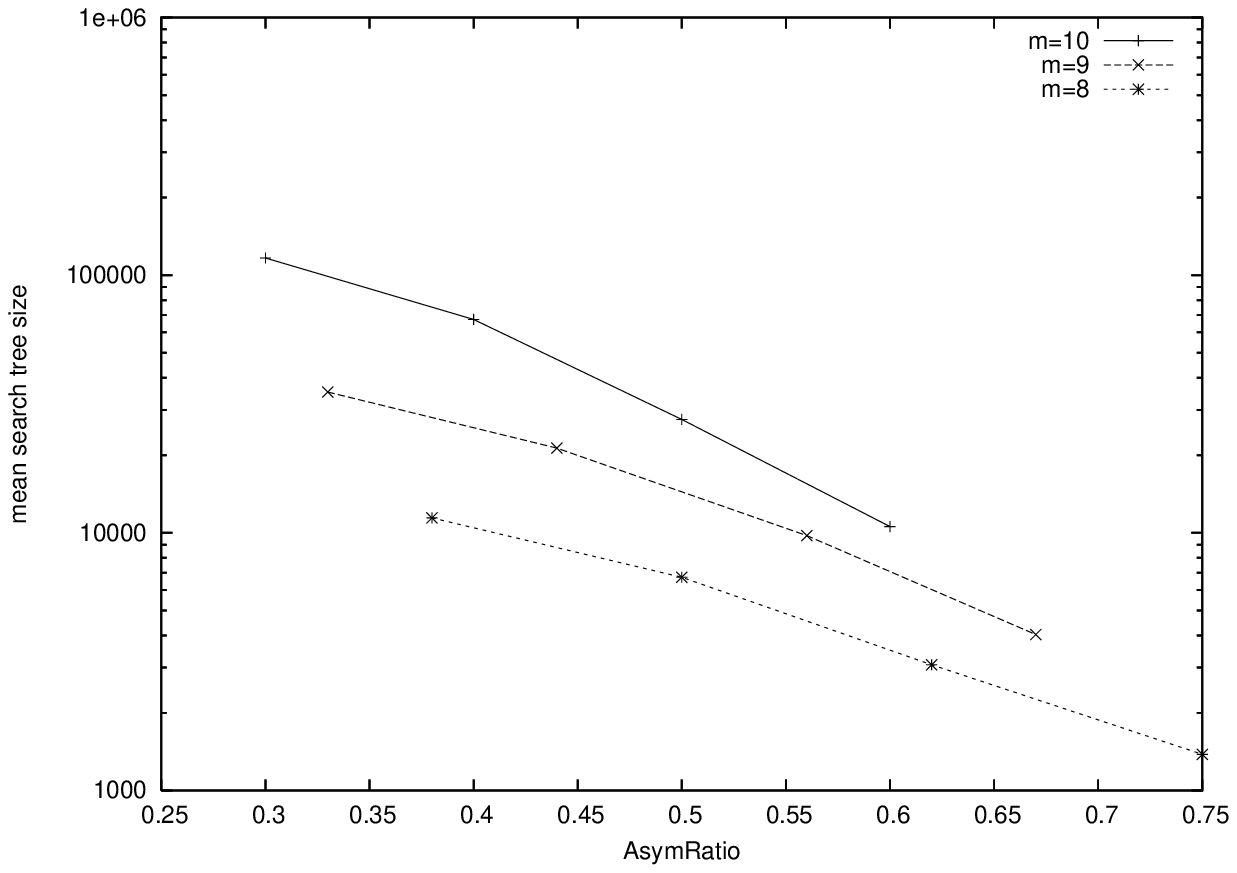} & \includegraphics[width=7.4cm]{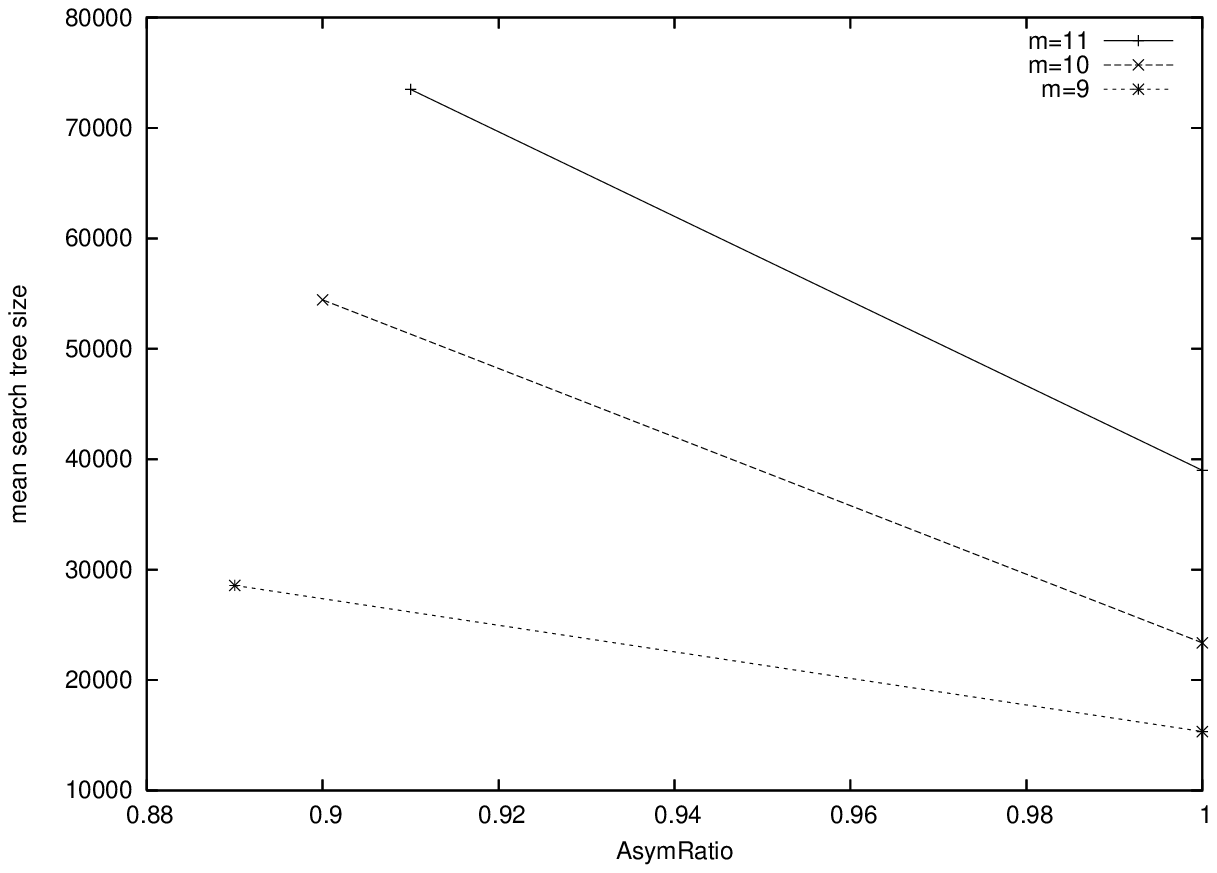}\\\vspace{0.3cm}
      (e) Rovers & (f) TPP\\
    \end{tabular}
  \end{center}
  \vspace{-0.0cm}
  \caption{\label{gnu:snodes}Mean search tree size of \zchaff, plotted
    against $\ASratio$ (log-plotted in (e)). Shown for the 3 most
    populated $\ppp_m$ classes of the respective IPC settings of
    $\size$.}
\end{figure*}

\vspace{-0.0cm}
\begin{figure*}
  \begin{center}
    \begin{tabular}{cc}
      \includegraphics[width=7.4cm]{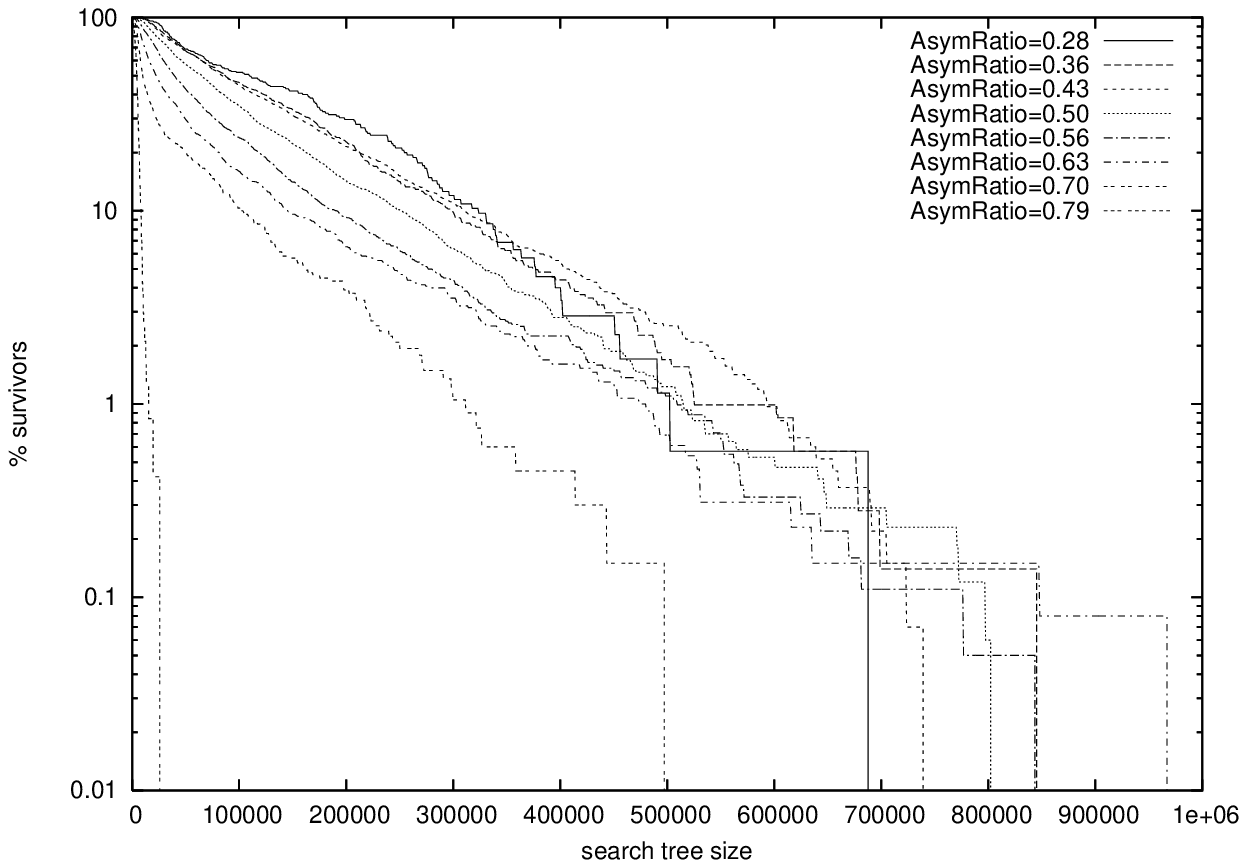} & \includegraphics[width=7.4cm]{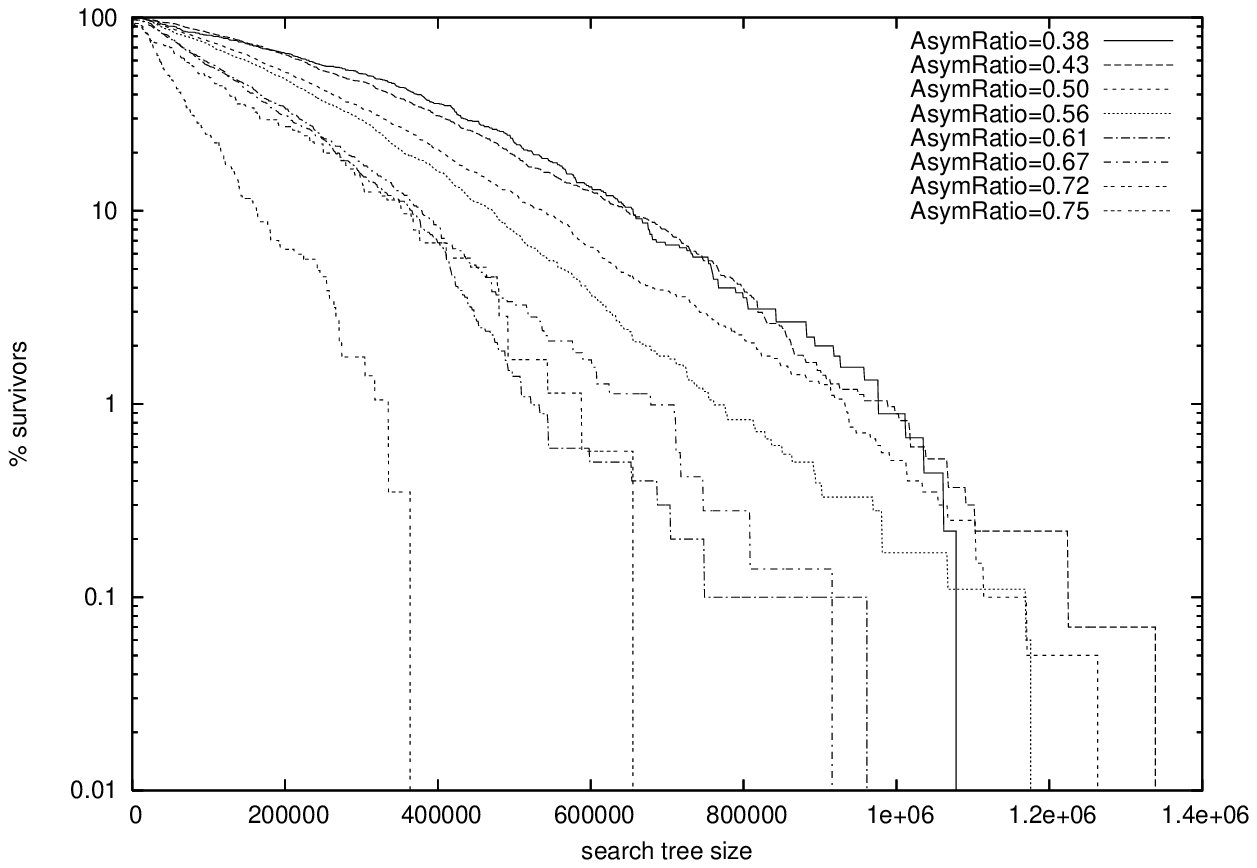}\\\vspace{0.3cm}
      (a) Blocksworld & (b) Depots\\
     \includegraphics[width=7.4cm]{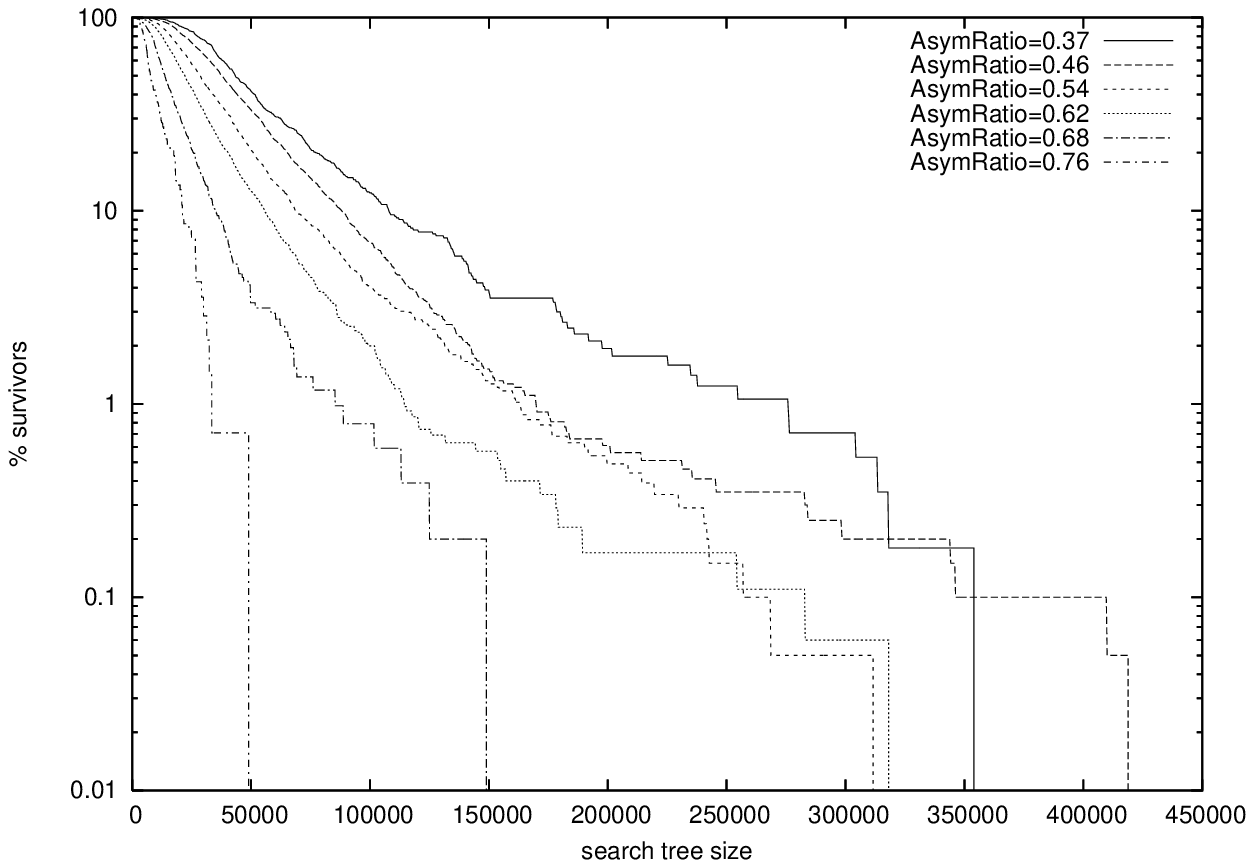} & \includegraphics[width=7.4cm]{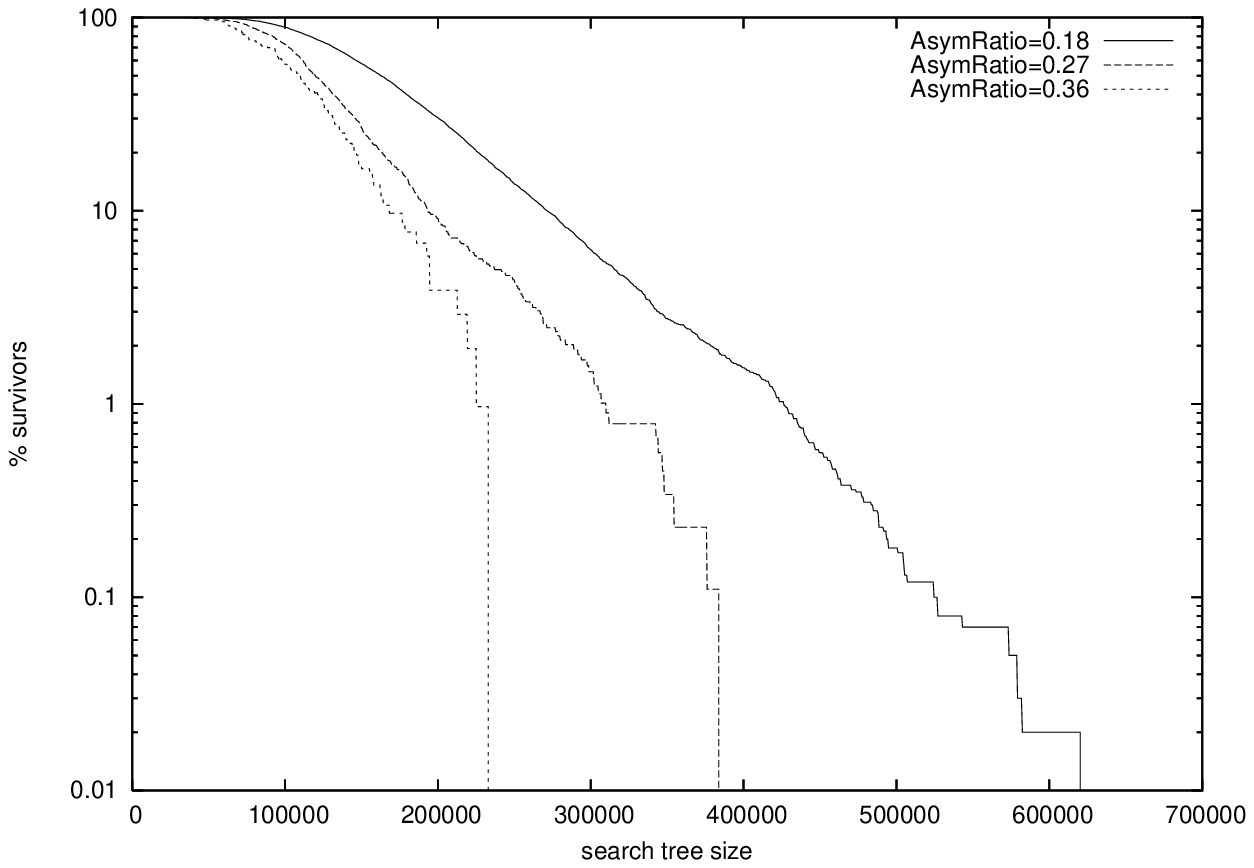}\\\vspace{0.3cm}
      (c) Driverlog & (d) Miconic-ADL\\
    \includegraphics[width=7.4cm]{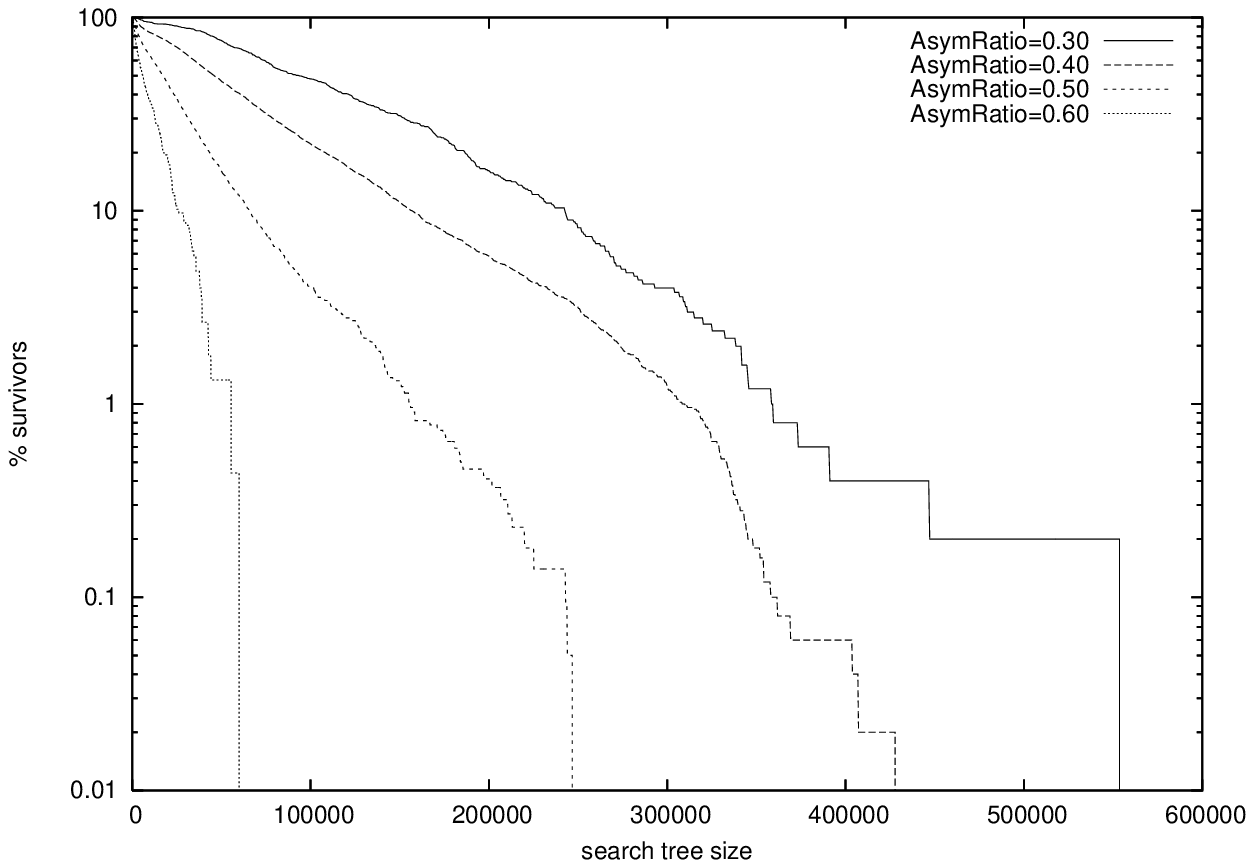} & \includegraphics[width=7.4cm]{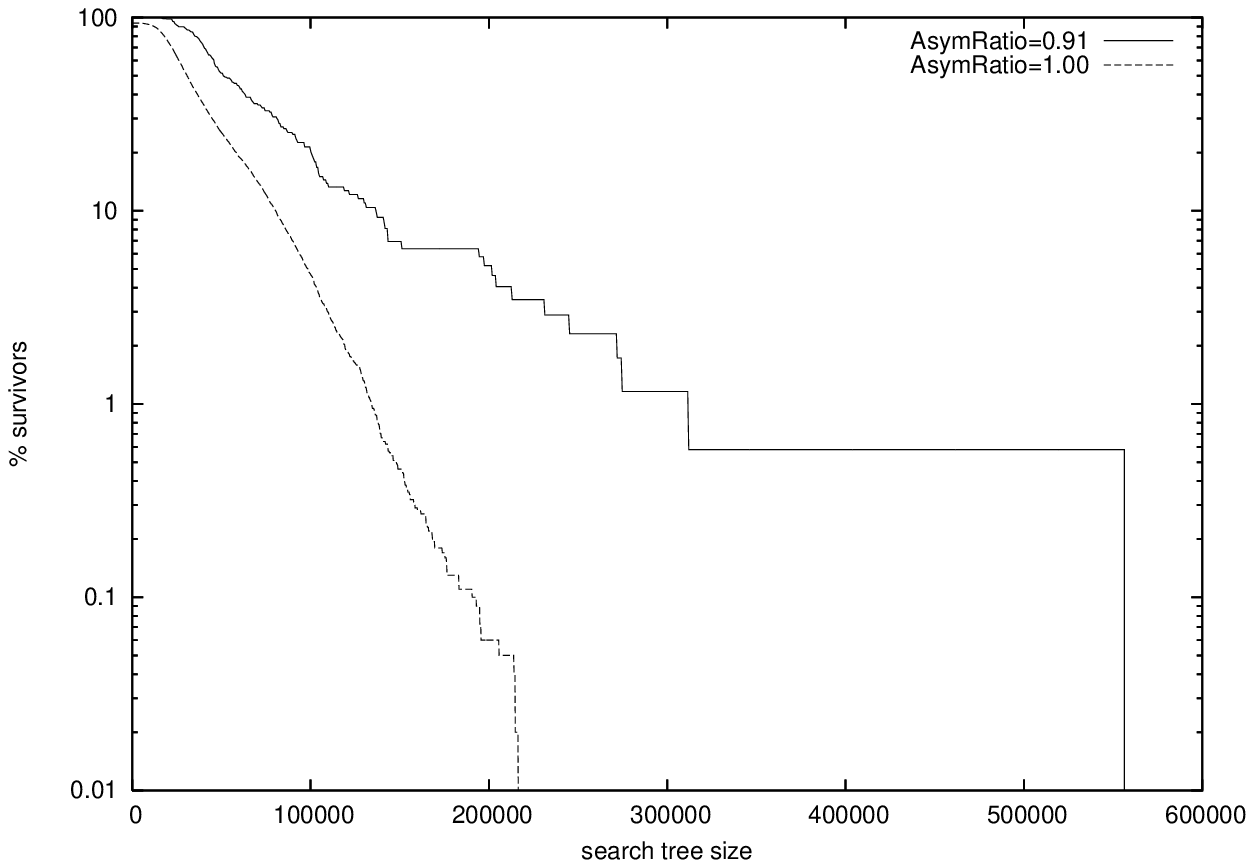}\\\vspace{0.3cm}
      (e) Rovers & (f) TPP\\
    \end{tabular}
  \end{center}
  \vspace{-0.0cm}
  \caption{\label{gnu:survivors}Search tree size distributions for
  different settings of $\ASratio$, in terms of the survivor functions
  over search tree size; log-scaled in $y$. Shown for the maximum $m$
  class $\ppp_m$ for each of the domains from Figure~\ref{gnu:snodes},
  with the IPC settings of $\size$.}
\end{figure*}

Figures~\ref{gnu:snodes} and~\ref{gnu:survivors} provide some
quantitative results. In Figure~\ref{gnu:snodes}, we take the mean
value of each $\ASratio$ bin, and plot that over $\ASratio$. We do so
for a selection of domains, with IPC size parameter settings, and for
a selection of classes $\ppp_m$. We select Blocksworld, Depots,
Driverlog, Miconic-ADL, Rovers, and TPP; we show data for the 3 most
populated $\ppp_m$ classes. The decrease of mean search tree size over
$\ASratio$ is very consistent, with some minor perturbations primarily
in Blocksworld and Depots. The most remarkable behavior is obtained in
Rovers, where the mean search tree size decrases exponentially over
$\ASratio$; note the logarithmic scale of the $y$ axis in
Figure~\ref{gnu:snodes} (e).\footnote{We can only speculate what the
reason for this extraordinarily strong behavior in Rovers is. High
$\ASratio$ values arise mainly due to long paths to be travelled on
the road map. Perhaps there is less need to go back and forth in
Rovers than in domains like Driverlog; one can transmit data from many
locations. This might have an effect on how tightly the search gets
constrained.} From the relative positions of the different curves, one
can also nicely see the influence of optimal plan length/formula size
-- the longer the optimal plan, the larger the search tree.

Figure~\ref{gnu:survivors} provides some concrete examples of the
search tree size distributions. They are plotted in terms of their
survivor functions.  In these plots, search tree size increases on the
$x$ axis, and the $y$ axis shows the percentage of instances that
require more than $x$ search nodes. Figure~\ref{gnu:survivors} shows
data for the same domains and size settings as
Figure~\ref{gnu:snodes}. For each domain, the maximum $m$ of the 3
most populated $\ppp_m$ classes is selected -- in other words, we
select the maximum $m$ classes from Figure~\ref{gnu:snodes}.  Each
graph contains a separate survivor function for every distinct value
of $\ASratio$, within the respective domain and ${\mathcal P}_m$. The
graphs thus show how the survivor function changes with
$\ASratio$. The $y$ axes are log-scaled to improve readability;
without the log-scale, the outliers, i.e., the few instances with a
very large search tree size, cannot be seen.

The behavior we expect to see, according to
Hypothesis~\ref{hypo:ASratio}, is that the survivor functions shift to
the left -- to lower search tree sizes -- as $\ASratio$
increases. Indeed, with only few exceptions, this is what
happens. Consider the upper halfs of the graphs, containing $99\%$ of
the instances in each domain. Almost without exception, the survivor
functions are parallel curves shifting to the left over increasing
$\ASratio$. In the lower halfs of the graphs, the picture is a little
more varied; the curves sometimes cross in Blocksworld, Depots, and
Driverlog. In Driverlog, for example, with $\ASratio = 0.37$, the
maximum costly instance takes 353953 nodes; with $\ASratio = 0.46$,
$0.1\%$ of the instances require more search nodes than that. So it
seems that, sometimes, $\ASratio$ is not as good an indicator on
outliers. Still, the bulk of the distributions behaves according to
Hypothesis~\ref{hypo:ASratio}.

\vspace{-0.0cm}
\begin{figure*}
  \begin{center}
    \begin{tabular}{cc}
\includegraphics[width=7.4cm]{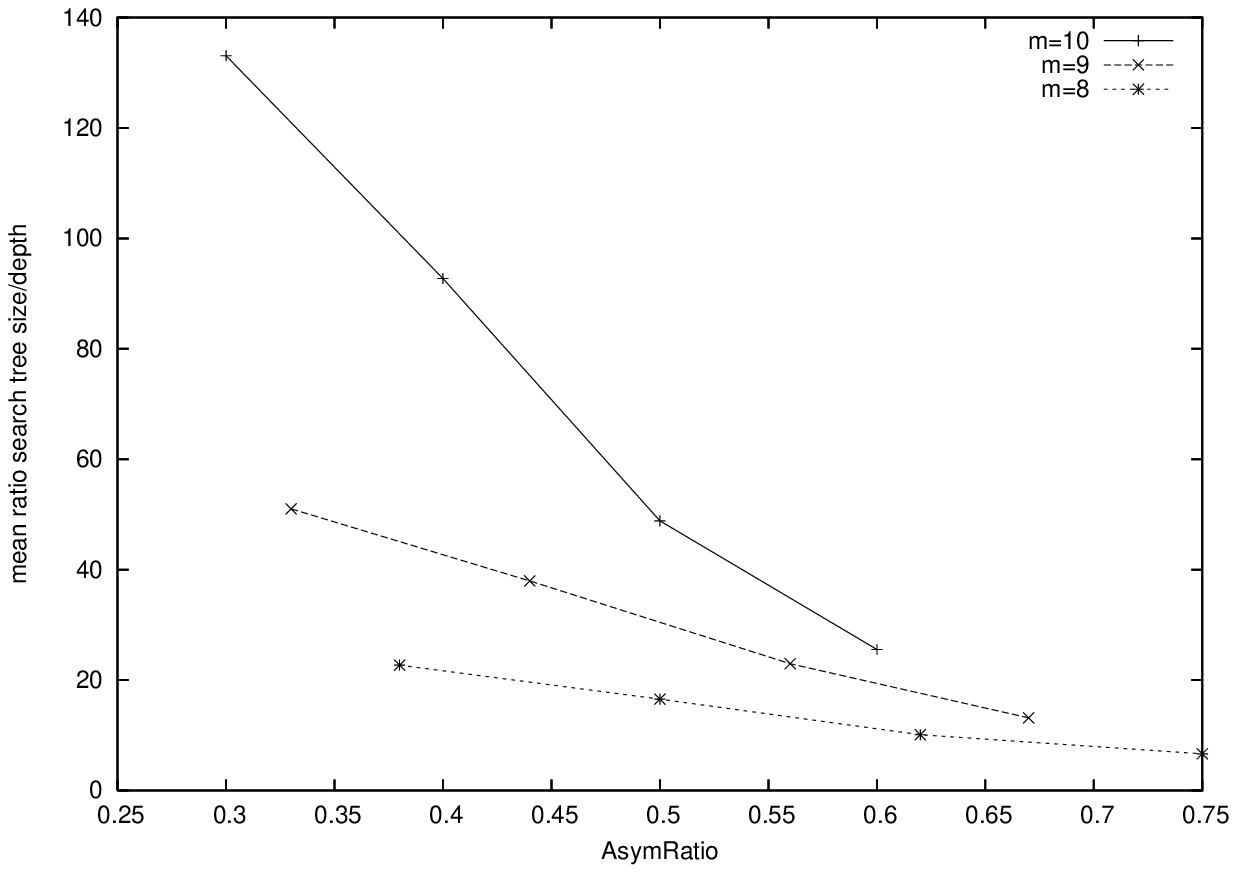} & \includegraphics[width=7.4cm]{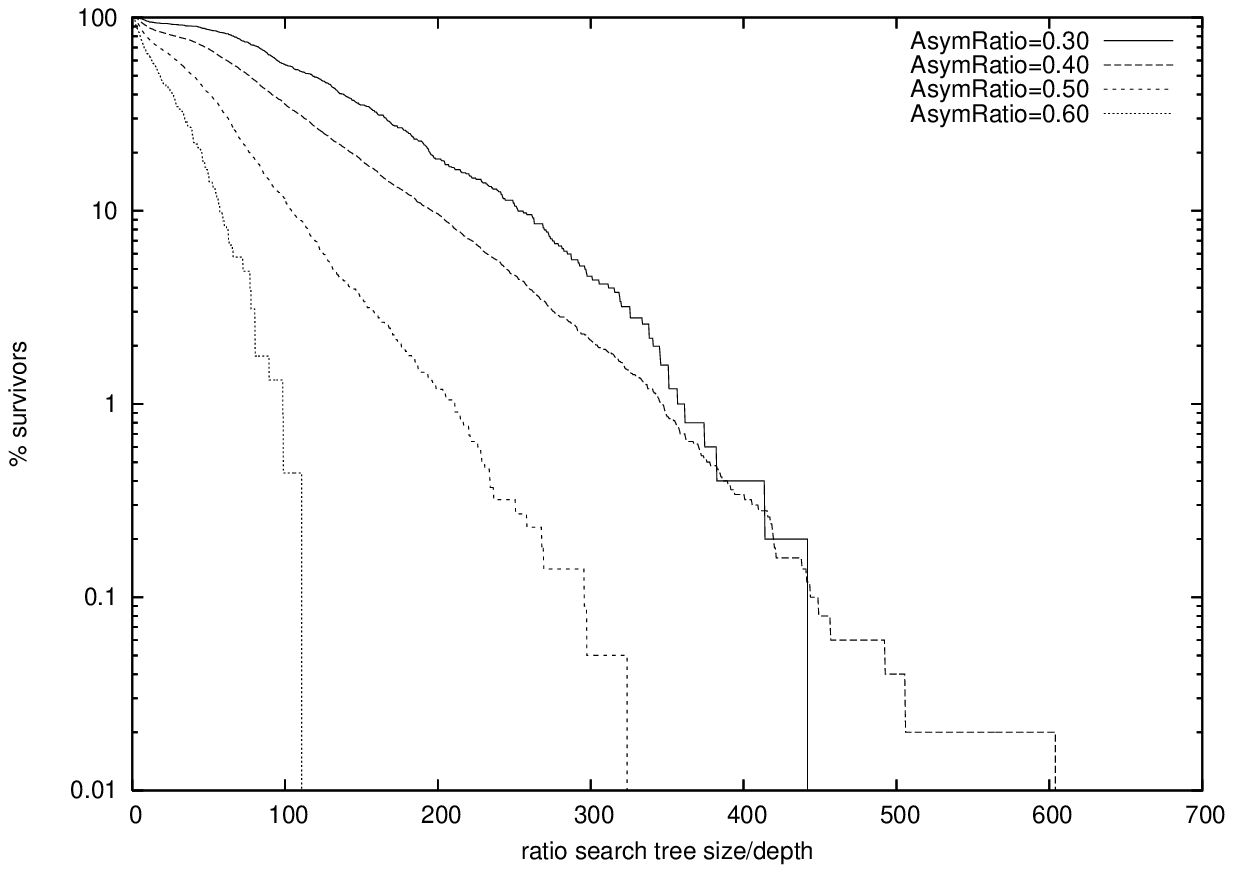}\\
    \end{tabular}
  \end{center}
  \vspace{-0.0cm}
  \caption{\label{gnu:ratio}Size/depth ratio of \zchaff's search trees
  in Rovers. Left hand side: mean values plotted against
  $\ASratio$. Right hand side: survivor functions for $m=10$.}
  \vspace{-0.5cm}
\end{figure*}

Beside the search tree size of \zchaff, we also measured the (maximum)
search tree depth, the size of the identified backdoors (the sets of
variables branched upon), and the ratio between size and depth of the
search tree. The latter gives an indication of how ``broad'' or
``thin'' the shape of the search tree is. Denoting depth with $d$, if
the tree is full binary then the ratio is $(2^{d+1}-1)/d$; if the tree
is degenerated to a line then the ratio is $(2d+1)/d$. Plotting these
parameters over $\ASratio$, in most domains we obtained behavior very
similar to what is shown in Figures~\ref{gnu:snodes}
and~\ref{gnu:survivors}. As an example, Figure~\ref{gnu:ratio} shows
the size/depth ratio data for the Rovers domain. We find the results
regarding size/depth ratio particularly interesting. They nicely
reflect the intuition that, as problem structure increases, UP can
prune many branches early on and so makes the search tree grow
thinner.

%% file: sd.tex
\section{Analyzing Goal Asymmetry in Synthetic Domains}
\label{sd}

In this section, we perform a number of case studies. We analyze
synthetic domains constructed explicitly to provoke interesting
behavior regarding $\ASratio$. The aim of the analysis is to obtain a
better understanding of how this sort of problem structure affects the
behavior of SAT solvers; in fact, the definition of $\ASratio$ was
motivated in the first place by observations we made in synthetic
examples.

The analytical results we obtain in our case studies are, of course,
specific to the studied domains. We do, however, identify a set of
prototypical patterns of structure that also appear in the planning
competition examples; we will point this out in the text.

We analyze three classes of synthetic domains/CNF formulas, called
MAP, SBW, and SPH. MAP is a simple transportation kind of domain, SBW
is a block stacking domain. SPH is a structured version of the pigeon
hole problem. Each of the domains/CNF classes is parameterized by size
$n$ and structure $k$. In the planning domains, we use the simplified
Graphplan-based encoding described in Section~\ref{prelim}, and
consider CNFs that are one step short of a solution. We denote the
CNFs with $MAP_n^k$, $SBW_n^k$, and $SPH_n^k$, respectively.

We choose the MAP and SBW domains because they are related to
Logistics and Blocksworld, two of the most classical Planning
benchmarks. We chose SPH for its close relation to the formulas
considered in proof complexity. The reader will notice that the
synthetic domains are {\em very} simple.  The reasons for this are
threefold. First, we wanted to capture the intended intuitive problem
structure in as clean a form as possible, without ``noise''. Second,
even though the Planning tasks are quite simple, the resulting CNF
formulas are complicated -- e.g., much more complicated than the
pigeon hole formulas often considered in proof complexity. Third, we
identify provably minimal backdoors.  To do so, one has to take
account of every tiny detail of the effects of unit propagation. The
respective proofs are already quite involved for our simple domains --
for MAP, e.g., they occupy 9 pages, featuring myriads of interleaved
case distinctions.  To analyze more complex domains, one probably has
to sacrifice precision. \ifTR{For the sake of readability, the proofs
are moved to Appendix~\ref{proofs}, and only briefly sketched here.}

\ifLMCS{For the sake of readability, herein we discuss only MAP in
detail, and we replace the proofs with proof sketches. The details for
SBW and SPH, and the proofs, are in the TR
\cite{hoffmann:etal:tr-lmcs06}.}

\input{MAP}

\ifTR{

\input{SBW}

\input{SPH}

}

\ifLMCS{

\subsection{SBW}
\label{sd:sbw}

This is a block-stacking domain, with stacking restrictions on what
blocks can be stacked onto what other blocks.  The blocks are
initially all located side-by-side on a table $t_1$.  The goal is to
bring all blocks onto another table $t_2$, that has only space for a
single block; so the $n$ blocks must be arranged in a single stack on
top of $t_2$. The parameter $k$, $0 \leq k \leq n$, defines the amount
of restrictions. There are $k$ ``bad'' blocks $b_1, \dots, b_k$ and
$n-k$ ``good'' blocks $g_1, \dots, g_{n-k}$. Each $b_i$, $i>1$, can
only be stacked onto $b_{i-1}$; $b_1$ can be stacked onto $t_2$ and
any $g_i$.  The $g_i$ can be stacked any $g_j$, and onto $t_2$.

Independently of $k$, the optimal plan length is $n$: move actions
stack one block onto another block or a table. $\ASratio$ is $1/n$ if
$k=0$, and $k/n$ otherwise. In the symmetrical case, $k=0$, we
identify backdoors of size $\Theta(n^3)$ -- linear in the total number
of variables.  In the asymmetrical case, $k=n-2$, there are $O(log n)$
DPLL refutations and backdoors. It is an open question whether there
is an exponential lower bound in the symmetrical case.

\subsection{SPH}
\label{sd:sph}

For this domain, we modified the pigeon hole problem. In our $SPH_n^k$
formulas, as usual the task is to assign $n+1$ pigeons to $n$
holes. The new feature is that there is one ``bad'' pigeon that
requires $k$ holes, and $k-1$ ``good'' pigeons that can share a hole
with the bad pigeon. The remaining $n-k+1$ pigeons are normal, i.e.,
need exactly one hole each. The range of $k$ is between $1$ and $n-1$.
Independently of $k$, $n+1$ holes are needed overall.

We identify minimal backdoors for all combinations of $k$ and $n$;
their size is $(n-k)*(n-1)$. For $k = n-1$ we identify an $O(n)$ DPLL
refutation. With results by Buss and Pitassi
\cite{buss:pitassi:csl-97}, this implies an exponential DPLL
complexity gap to $k = 1$.

}

%% file: MAP.tex
\subsection{MAP} 
\label{sd:map}

\ifLMCS{\subsubsection{Domain Definition}}

In this domain, one moves on the undirected graph shown in
Figure~\ref{maproadmap} (a) and (b).  The available actions take the
form $move\mbox{-}x\mbox{-}y$, where $x$ is connected to $y$ with an
edge in the graph. The precondition is $\{at\mbox{-}x\}$, the add
effect is $\{at\mbox{-}y, visited\mbox{-}y\}$, and the delete effect
is $\{at\mbox{-}x\}$.

The number of nodes in the graph is $3n-3$. Initially one is located
at $L^0$.  The goal is to visit a number of locations.  Which
locations must be visited depends on the value of $k
\in \{1, 3, \dots, 2n-3\}$. If $k=1$ then the goal is to visit each of
$\{L_1^1, \dots, L_n^1\}$. For each increase of $k$ by $2$, the goal
on the $L_1$-branch goes up by two steps, and one of the other goals
is skipped. For $k=2n-3$ the goal is $\{L_1^{2n-3}, L_2^1\}$. (For $k
= 2n-1$, $MAP_n^k$ contains an empty clause: no supporting action for
the goal is present at the last time step.) We refer to $k=1$ as the
{\em symmetrical case}, and to $k=2n-3$ as the {\em asymmetrical
case}, see Figure~\ref{maproadmap} (a) and Figure~\ref{maproadmap}
(b), respectively.\footnote{In
Figure~\ref{maproadmap} (a), the graph nodes $L_2^1,
\dots, L_n^1$ can be permuted. Running SymChaff \cite{sabharwal:aaai-05}, a
version of \zchaff\ extended to exploit symmetries, we could solve the
MAP formulas relatively easily. This is an artefact of the simplified
structure of the MAP domain. In some experiments we ran on low
$\ASratio$ examples from Driverlog and Rovers, exploiting symmetries
did not make a discernible difference. Intuitively, like in MAP, all
goals are cheap to achieve; unlike in MAP, they are not completely
symmetric -- this is a side-effect of MAP's overly abstract
nature. Exploring this in more depth is a topic for future work.}

\vspace{-0.0cm}
\begin{figure}
\begin{center} 
\begin{tabular}{cc}\vspace{0.3cm}
    \includegraphics[width=5.8cm]{XFIG/map-base.eps} \hspace{0.3cm} &
    \hspace{0.3cm}
    \includegraphics[width=5.8cm]{XFIG/map-top.eps}\\\vspace{0.6cm}
    (a) goals symmetrical case & (b) goals asymmetrical case\\\vspace{0.3cm}
    \includegraphics[width=5.8cm]{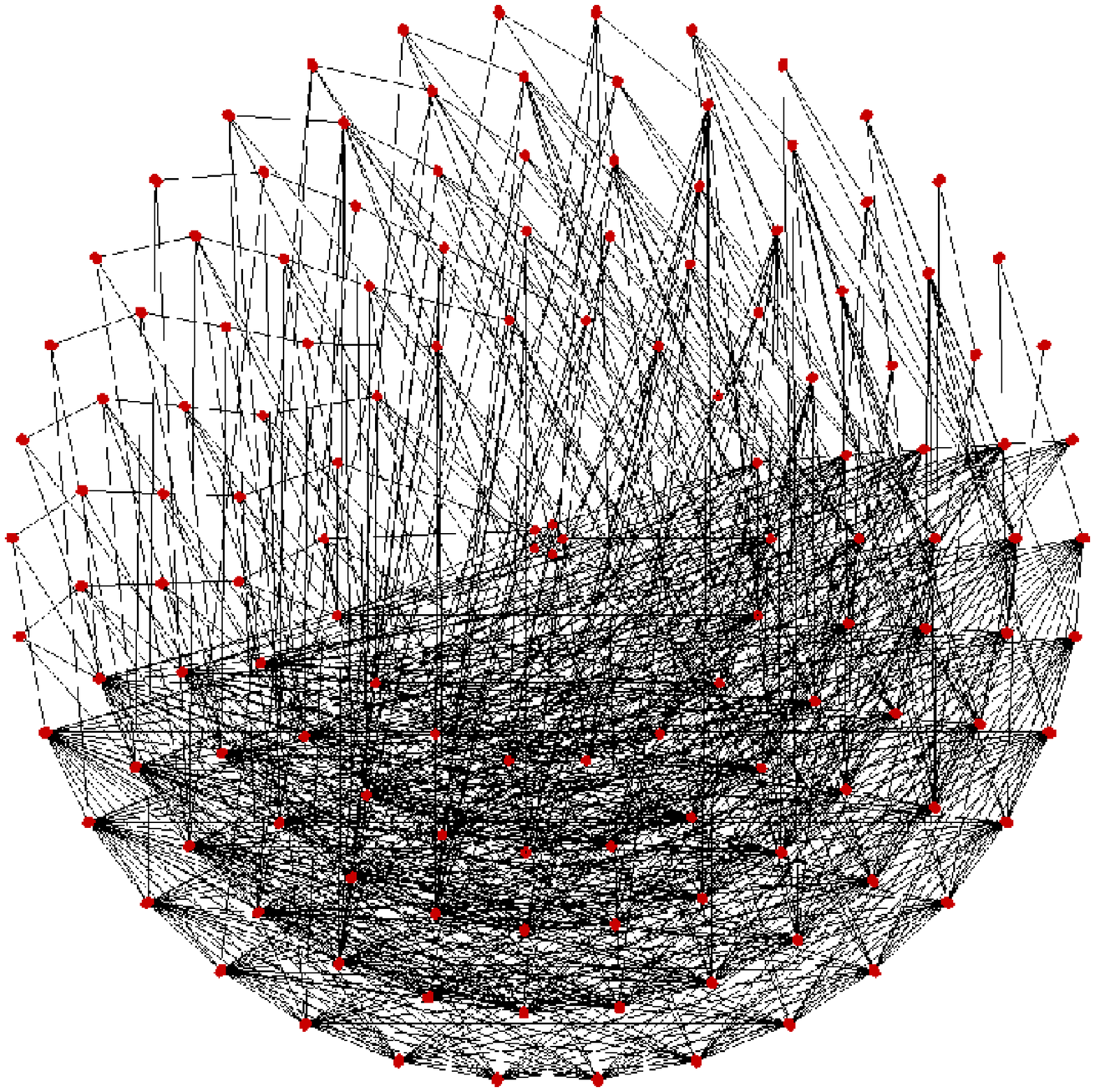}
    \hspace{0.3cm} & \hspace{0.3cm}
    \includegraphics[width=5.8cm]{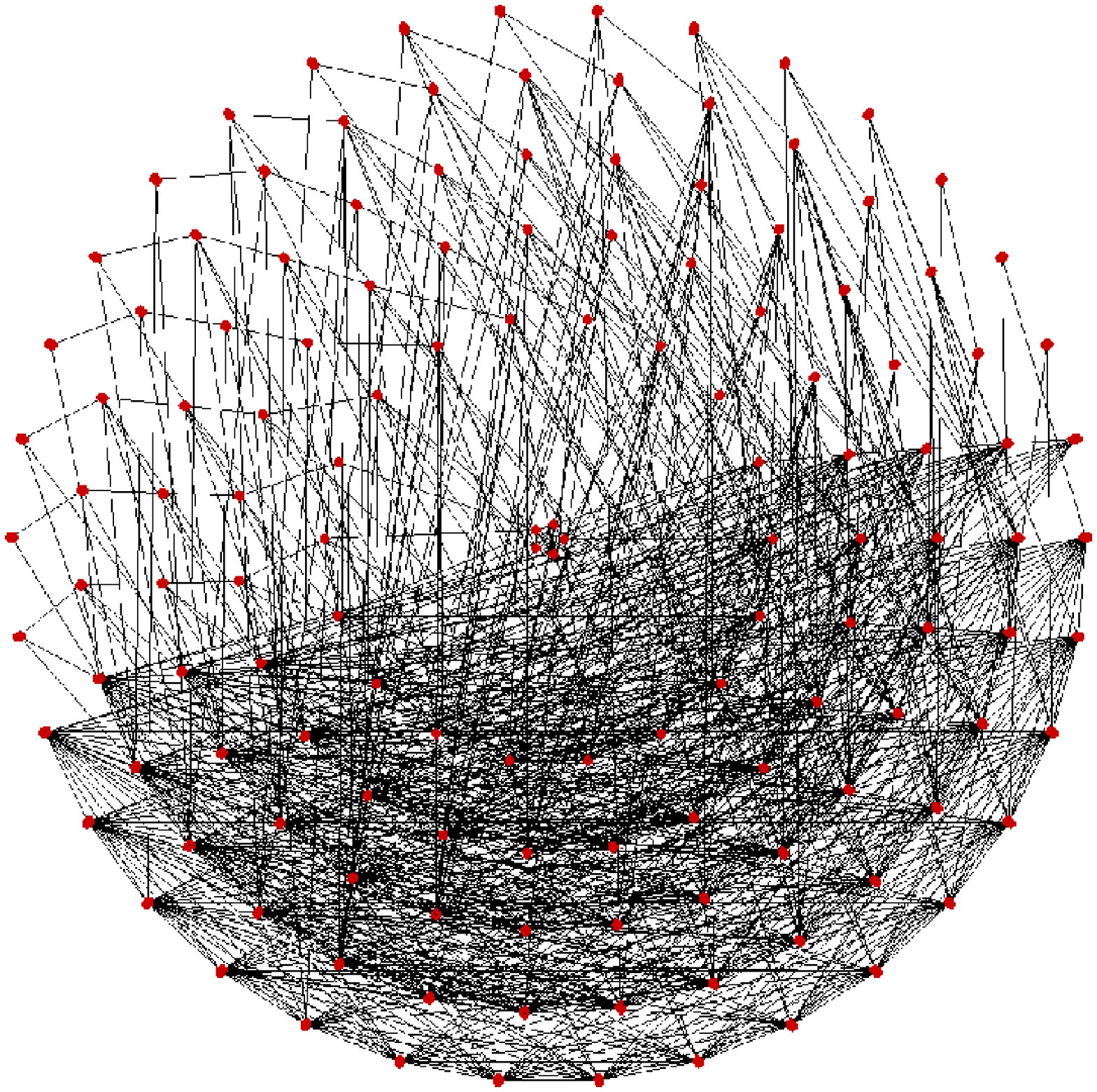}\\ 
    (c) constraint graph symmetrical case & (d) constraint graph asymmetrical case\\
\end{tabular} 
\end{center} 
\vspace{-0.0cm}
\caption{\label{maproadmap} Goals and constraint graphs in MAP. In
    (a) and (b), goal locations are indicated in bold face. In (c) and
    (d), $n=4$; the variables at growing time steps lie on circles
    with growing radius, edges indicate common membership in at least
    one clause.}\vspace{-0.0cm}
\end{figure}

The length of a shortest plan is $2n-1$ independently of $k$: one
first visits all goal locations on the right branches (i.e., the
branches except the $L_1$-branch), going forth and back from $L^0$;
then one descends into the $L_1$-branch. Our CNFs encode $2n-2$
steps. $\ASratio$ is $k/(2n-1)$ because achieving the goal on the
$L_1$-branch takes $k$ steps. In the symmetrical case, $\ASratio =
1/(2n-1)$ which converges to $0$; in the asymmetrical case, $\ASratio
= (2n-3)/(2n-1)$ which converges to $1$.

Figure~\ref{maproadmap} (c) and (d) illustrate that the setting of $k$
has only very little impact on the size and shape of the constraint
graph. As mentioned in Section~\ref{related}, the constraint graph is
the undirected graph where the nodes are the variables, and the edges
indicate common membership in at least one clause. Constraint graphs
form the basis for many notions of structure that have been
investigated in the area of constraint reasoning, e.g.,
\cite{dechter:ai-90,rish:dechter:jar-00,dechter03}.

Clearly, the constraint graphs do not capture the difference between
the symmetrical and asymmetrical cases in MAP. When stepping from
Figure~\ref{maproadmap} (c) to Figure~\ref{maproadmap} (d), one new
edge within the outmost circle is added, and three edges within the
outmost circle disappear (one of these is visible on the left side of
the pictures, just below the middle). More generally, between formulas
$MAP_n^k$ and $MAP_n^{k'}$, $k'>k$, there is no difference except that
$k'-k$ goal clauses are skipped, and that the content of the goal
clause for the $L_1$-branch changes. Thus, the problem structure here
cannot be detected based on simple examinations of CNF syntax. In
particular, it is easy to see that there are no small cutsets in the
MAP formulas, irrespectively of the setting of $k$. The reason are
large cliques of variables present in the constraint graphs for all
these formulas. \ifTR{Details on this are in Appendix~\ref{cutsets}.}
\ifLMCS{Details on this are in the TR
\cite{hoffmann:etal:tr-lmcs06}.}

\ifLMCS{\subsubsection{Symmetrical Case}}

The structure in our formulas does not affect the formula syntax much,
but it does affect the size of DPLL refutations, and backdoors. First,
we proved that, in the symmetrical case, the DPLL trees are large.

\begin{thm}[MAP symmetrical case, Resolution LB]\label{the:map-base-lb}
  Every resolution refutation of $MAP_n^1$ must have size exponential
  in $n$.
\end{thm}

\begin{cor}[MAP symmetrical case, DPLL LB]\label{the:map-symm-dpll-lb}
  Every DPLL refutation of $MAP_n^1$ must have size exponential in $n$.
\end{cor}

The proof of Theorem~\ref{the:map-base-lb} proceeds by a ``reduction''
of $MAP_n^1$ to a variant of the pigeon hole problem. A reduction here
is a function that transforms a resolution refutation of $MAP_n^1$
into a resolution refutation of the pigeon hole. Given a reduction
function from formula class A into formula class B, a lower bound on
the size of resolution refutations of B is also valid for A, modulo
the maximal size increase induced by the reduction.

We define a reduction function from $MAP_n^1$ into the {\em onto
functional pigeon hole problem}, $ofPHP_n$. This is the standard
pigeon hole -- where $n$ pigeons must be assigned to $n-1$ holes --
plus ``onto'' clauses saying that at least one pigeon is assigned to
each hole, and ``functional'' clauses saying that every pigeon is
assigned to at most one hole. Razborov \cite{razborov:jcss:04} proved
that every resolution refutation of $ofPHP_n$ must have size
$exp(\Omega(n/(log (n+1))^2))$.

Our reduction proceeds by first setting many variables in $MAP_n^1$ to
$0$ or $1$, and identifying other variables (renaming $x$ and $y$ to a
new variable $z$).\footnote{For example, we set all $NOOP\mbox{-}at$
variables to $0$. Such a variable will never be set to $1$ in an
optimal plan, since that would mean a commitment to not move at all in
a time step. Similar (more complicated) intuitions are behind all the
operations performed.} By some rather technical (but essentially
simple) arguments, we prove that such operations do not increase the
size of a resolution refutation.  The reduced formula is a
``temporal'' version of the onto pigeon hole problem; we call it
$oTPHP_n$. We will discuss this in detail below. We prove that, from a
resolution refutation of $oTPHP_n$, one can construct a resolution
refutation of $ofPHP_n$ by replacing each resolution step with at most
$n^2+n$ new resolution steps. This proves
Theorem~\ref{the:map-base-lb}.  Corollary~\ref{the:map-symm-dpll-lb}
follows immediately since DPLL corresponds to a restricted form of
resolution. The same is true for DPLL with clause learning
\cite{beame:etal:jair-04}, as done in the \zchaff\ solver we use in
our experiments.

The temporal pigeon hole problem is similar to the standard pigeon
hole problem except that now the ``holes'' are time steps. This nicely
reflects what typically goes on in Planning encodings, where the
available time steps are the main resource. So it is worth having a
closer look at this formula. We skip the ``onto'' clauses since these
are identical in both the standard and the temporal version. We denote
the standard pigeon hole with $PHP_n$, and the temporal version with
$TPHP_n$. Both make use of variables $x\_y$, meaning that pigeon $x$
goes into hole $y$. Both have clauses of the form $\{\neg x\_y, \neg
x'\_y\}$, saying that no pair of pigeons can go into the same
hole. $PHP_n$ has clauses of the form $\{x\_1, \dots, x\_(n-1)\}$,
saying that each pigeon needs to go into at least one hole. In
$TPHP_n$, these clauses are replaced with the following constructions
of clauses:
\begin{enumerate}
\item $\{\neg px\_2, x\_1\}$
\item $\{\neg px\_3, x\_2, px\_2\} \dots \{\neg px\_(n-1), x\_(n-2), px\_(n-2)\}$
\item $\{x\_(n-1), px\_(n-1)\}$
\end{enumerate}
Here, the $px\_y$ are new variables whose intended meaning is that
$px\_y$ is set to $1$ iff $x$ is assigned to some hole $y' < y$ --
some earlier time step than $y$. The $px\_y$ variables correspond to
the NOOP actions in Graphplan-based encodings. Clause (1) says that if
we decide to put $x$ into a time step earlier than $2$, we must put it
into step $1$. This is an action precondition clause. Clauses (2) say
that if we decide to put $x$ into a step earlier than some $y$, then
we must put $x$ into either $y-1$ or earlier than $y-1$. These also
are action precondition clauses. Clause (3) is a goal clause; it says
that we must have put $x$ somewhere by step $n-1$ at the
latest.\footnote{Note that $PHP_n$ can be obtained from $TPHP_n$
within few resolution steps resolving on the $px\_y$ variables. So it
is easy to turn a refutation of $PHP_n$ into a refutation of
$TPHP_n$. The inverse direction, which we need for our proof, is less
trivial; one replaces each variable $px\_y$ with its meaning $\{x\_1,
\dots, x\_(y-1)\}$, and then reasons about how to repair the
resolution steps.}

We believe that the temporal pigeon hole problem is quite typical for
planning situations; at least, it is clearly contained in some
Planning benchmarks more involved than MAP. A simple example is the
Gripper domain, where the task is to move $n$ balls from one room into
another, two at a time. With similar reduction steps as we use for
MAP, this problem can be transformed into the $TPHP$. The same is true
for various transportation domains with trivial road maps, e.g., the
so-called Ferry and Miconic-STRIPS domains. Likewise, Logistics can be
transformed into $TPHP$ if there is only one road map (i.e., only one
city, or only airports); in the general case, one obtains a kind of
sequenced $TPHP$ formula, where each pigeon must be assigned a
sequence of holes (corresponding to truck, airplane, truck). In a
similar fashion, assigning samples to time steps in the Rovers domain
is a temporal pigeon hole problem. Formulated very generally,
situations similar to the $TPHP$ arise when there is not enough time
to perform a number of duties. If each duty takes only few steps, then
there are many possible distributions of the duties over the time
steps. In its most extreme version, this situation is the $TPHP$ as
defined above.

\widthfig{11}{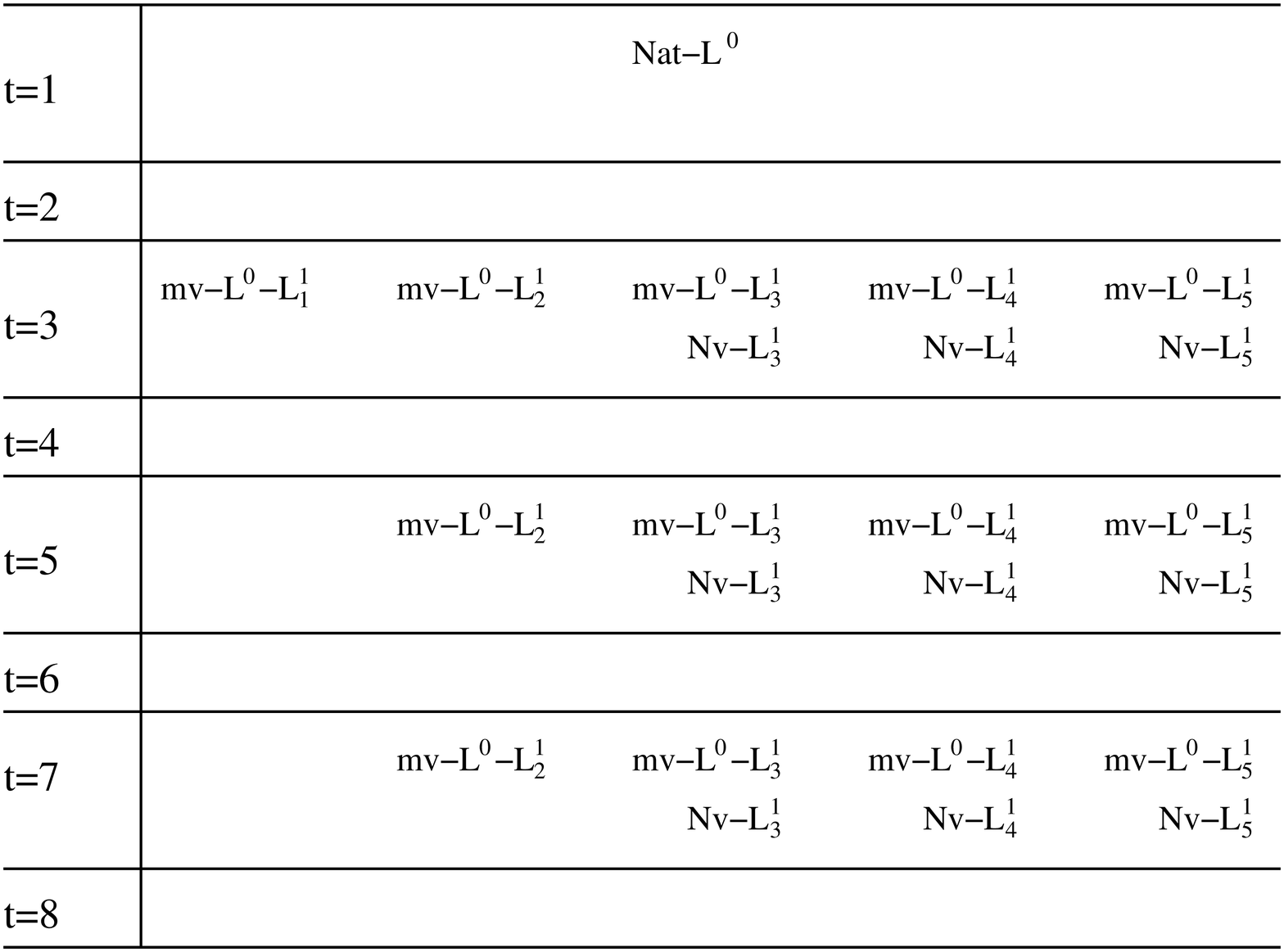}{The $MAP_n^1B$ variables for $n
  = 5$. The vertical axis corresponds to time steps $t$, the
  horizontal axis corresponds to branches on the map. ``NOOP-at'' is
  abbreviated as ``Nat'', ``NOOP-visited'' is abbreviated as ``Nv'',
  ``move'' is abbreviated as ``mv''.}{-0.0cm}

The proof of Theorem~\ref{the:map-base-lb} by reduction to the pigeon
hole is intriguing, but not particularly concrete about what is
actually going on inside a DPLL procedure run on $MAP_n^1$. To shed
more light on this, we now investigate the best choices of branching
variables for such a procedure. The backdoor we identify in the
symmetrical case, called $MAP_n^1B$, is shown in
Figure~\ref{XFIG/map-base-newBD.eps} for $n=5$. $MAP_n^1B$ contains,
at every time step with an odd index, $move\mbox{-}L^0\mbox{-}L_i^1$
and $NOOP\mbox{-}visited\mbox{-}L_i^1$ variables for all branches $i$
on the MAP, with some exceptions. A few additional variables are
needed, including $NOOP\mbox{-}at\mbox{-}L^0$ at the first time step
of the encoding. \ifTR{Full details are in Appendix~\ref{proofs}.}
\ifLMCS{Full details are in the TR
\cite{hoffmann:etal:tr-lmcs06}.} The size of $MAP_n^1B$ is $\Theta(n^2)$: 
$\Theta(n)$ time steps with $\Theta(n)$ variables each. Remember that
the total number of variables is also $\Theta(n^2)$, so the backdoor
is a linear-size variable subset.

\begin{thm}[MAP symmetrical case, BD]\label{mapbottom}
  $MAP_n^1B$ is a backdoor for $MAP_n^1$.
\end{thm}

For the proof, first note that, in the encoding, {\em any pair of move
  actions is incompatible}. So if one move action is set to $1$ at a
  time step, then all other move actions at that step are forced out
  by UP over the mutex clauses -- the time step is ``occupied'' (this
  is relevant also in the asymmetrical case below).  Now, to see the
  basic proof argument, assume for the moment that $MAP_n^1B$ contains
  all $move\mbox{-}L^0\mbox{-}L^1_i$ and
  $NOOP\mbox{-}visited\mbox{-}L_i^1$ variables, at each odd time step.
  Assigning values to all these variables results, by UP, in a sort of
  goal regression. In the last time step of the encoding, $t=2n-2$,
  the goal clauses form $n$ constraints requiring to either visit a
  location $L^1_i$, or to have visited it earlier already (i.e., to
  achieve it via a NOOP). Examining the interactions between $move$s
  and $NOOP$s at $t=2n-3$, one sees that, if all these are set, then
  at least $n-1$ goal constraints will be transported down to
  $t=2n-4$.  Iterating the argument over the $n-2$ odd time steps, one
  gets two goal constraints at $t=2$: two nodes $L^1_i$ must be
  visited within the first two time steps. It is easy to see, then,
  that branching over $NOOP\mbox{-}at\mbox{-}L^0$ at time $1$ yields
  an empty clause in either case. What makes identifying a
  non-redundant (minimal) backdoor difficult is that UP is slightly
  more powerful than just performing the outlined ``regression''.
  $MAP_n^1B$ contains hardly any variables for branch $i=1$. So at
  $t=2$ one gets only a single goal constraint, achieving which isn't
  a problem.  We perform an intricate case distinction about the
  precise pattern of time steps that are occupied after the
  regression, taking account of, e.g., such subtleties as the
  possibility to achieve $visited\mbox{-}L_1^1$ by moving in from
  $L_1^2$ above.  In the end, one can show that UP enforces
  commitments to accommodate also the two
  $move\mbox{-}L^0\mbox{-}L^1_i$ actions that weren't accommodated in
  the regression. For this, there is not enough room left.

We conjecture that the backdoor identified in Theorem~\ref{mapbottom}
is also a minimum size (i.e., an optimal) backdoor; for $n \leq 4$ we
verified this empirically, by enumerating all smaller variable sets.
(Enumerating variable sets in small enough examples was also our
method to find the backdoors in the first place.) We proved that the
backdoor is minimal.

\begin{thm}[MAP symmetrical case, BD
  minimality]\label{mapbottom:upcons}
  Let $B'$ be a subset of $MAP_n^1B$ obtained by removing one
  variable.  Then the number of UP-consistent assignments to the
  variables in $B'$ is always greater than $0$, and at least $(n-3)!$
  for $n \geq 3$.
\end{thm}

To prove this theorem, one figures out how wrong things can go when a
variable is missing in the proof of Theorem~\ref{mapbottom}.

\widthfig{11}{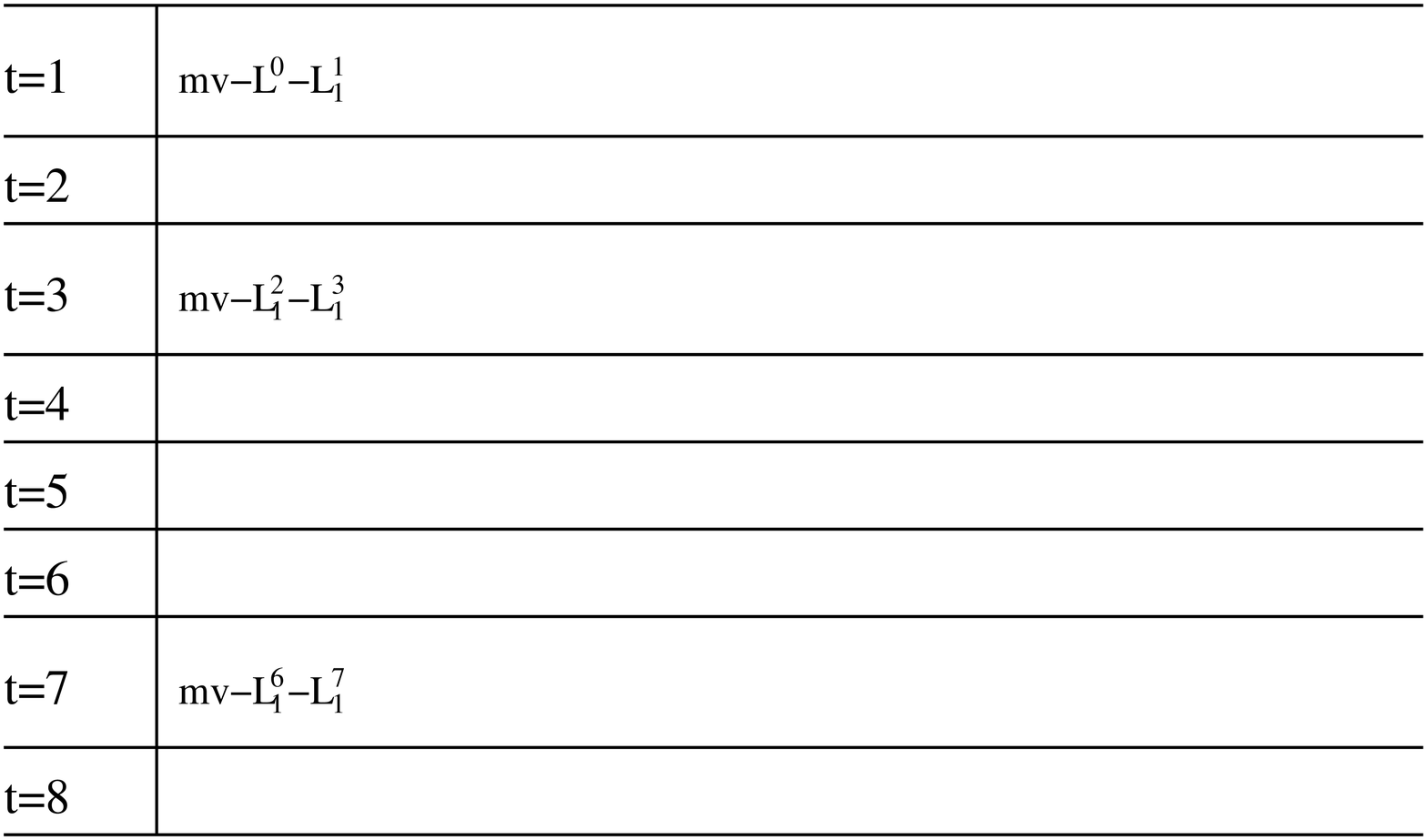}{The $MAP_n^{2n-3}B$ variables for
  $n = 5$. The vertical axis corresponds to time steps $t$, the
  horizontal axis corresponds to branches on the map (of which only
  one, the $L_1$-branch, is relevant for the backdoor). The variables
  stay the same for $n=6, 7, 8$. Compare to
  Figure~\ref{XFIG/map-base-newBD.eps}.}{-0.0cm}

\ifLMCS{\subsubsection{Asymmetrical Case}}

When starting to investigate the backdoors in the asymmetrical case,
our expectation was to obtain $\Theta(n)$ backdoors involving only the
map branch to the outlier goal node. We were surprised to find that
one can actually do much better. The backdoor we identify, called
$MAP_n^{2n-3}$, is shown in Figure~\ref{XFIG/map-base-newBD.eps} for
$n=5$. The general form is:
\begin{itemize}
\item $move\mbox{-}L^{0}\mbox{-}L_1^{1}$ at step $1$
\item $move\mbox{-}L_1^{2}\mbox{-}L_1^{3}$ at step $3$
\item $move\mbox{-}L_1^{6}\mbox{-}L_1^{7}$ at step $7$
\item $move\mbox{-}L_1^{14}\mbox{-}L_1^{15}$ at step $15$
\item \dots
\end{itemize}
That is, starting with $t=2$ one has
$move\mbox{-}L_1^{t-2}\mbox{-}L_1^{t-1}(t-1)$ variables where the
value of $t$ is doubled between each two variables. The size of
$MAP_n^{2n-3}B$ is $\lceil log_2 n \rceil$.

\begin{thm}[MAP asymmetrical case, BD]\label{maptop}
  $MAP_n^{2n-3}B$ is a backdoor for $MAP_n^{2n-3}$.
\end{thm}

We again conjecture that this is also a minimum size backdoor. For $n
\leq 8$ we verified this empirically. Note that, while the size of the
backdoor is $O(log n)$, $n$ itself is asymptotic to the square root of
the number of variables in the formula. We can show that the backdoor
is minimal.  Precisely, we have:

\begin{thm}[MAP asymmetrical case, BD
  minimality]\label{maptop:upcons}
  Let $B'$ be a subset of $MAP_n^{2n-3}B$ obtained by removing one
  variable. Then there is exactly one UP-consistent assignment to the
  variables in $B'$.
\end{thm}

The proof of Theorem~\ref{maptop} is explained with an example
below. Proving Theorem~\ref{maptop:upcons} is a matter of figuring out
what can go wrong in the proof to Theorem~\ref{maptop}, after removing
one variable.

We consider it particularly interesting that the $MAP_n^{2n-3}$
formulas have {\em logarithmic} backdoors. This shows, on the one
hand, that these formulas are (potentially) easy for Davis Putnam
procedures, having polynomial-size refutations.\footnote{We will see
below that the DPLL refutations in fact degenerate to lines, having
the same size as the backdoors themselves.} On the other hand, the
formulas are non-trivial, in two important respects. First, they do
have non-constant backdoors and are not just solved by unit
propagation. Second, finding the logarithmic backdoors involves, at
least, a non-trivial branching heuristic -- the worst-case DPLL
refutations of $MAP_n^{2n-3}$ are still exponential in $n$. For
example, UP will not cause any propagation if the choice of branching
variables is $\{NOOP\mbox{-}visited\mbox{-}L_i^1(t) \mid 2 \leq i \leq
n\}$ for one time step $3 \leq t
\leq 2n-3$.

More generally, our identification of $O(log n)$ backdoors rather than
$O(n)$ backdoors is interesting since it may serve to explain recent
findings in more practical examples. Williams et al
\cite{williams:etal:ijcai-03} have empirically found backdoors in the 
order of 10 out of 10000 variables in CNF encodings of many standard
Planning benchmarks (as well as Verification benchmarks). In this
context, it is instructive to have a closer look at how the
logarithmic backdoors in the MAP formulas arise. We do so in detail
below. A high-level intuition is that one can pick the branching
variables in a way exploiting the need to go back and forth on the
path to the outlier goal node. Going back and forth introduces a
factor $2$, and so one can double the number of time steps between
each pair of variables. Similar phenomena arise in other domains, in
the presence of ``outlier goal nodes'', i.e., goals that necessitate a
long sequence of steps. Good examples for this are transportation of a
package to a far-away destination, or the taking of a far-away sample
in Rovers. Note that this corresponds to a distorted version of the
$TPHP$, where one pigeon must be assigned to an entire sequence of
time steps, and hence the number of different distributions of pigeons
over time steps is small.

We now examine an example formula, $MAP_n^{2n-3}$ for $n=8$, in
detail. The proof of Theorem~\ref{maptop} uses the following two
properties of UP, in $MAP_n^{2n-3}$:
\begin{sloppypar}
\begin{enumerate}
\item[(1)] If one sets a variable
  $move\mbox{-}L_1^{i-1}\mbox{-}L_1^{i}(i)$ to $1$, then at all time
  steps $j < i$ a move variable is set to $1$ by
  UP.
\item[(2)] If one sets a variable
  $move\mbox{-}L_1^{i-1}\mbox{-}L_1^{i}(i)$ to $0$, then at all time
  steps $j > i$ a move variable is set to $1$ by
  UP.
\end{enumerate}
\end{sloppypar}
Both properties are caused by the ``tightness'' of branch 1, i.e., by
UP over the precondition clauses of the actions moving along that
branch.  Other than what one may think at first sight, the two
properties by themselves are {\em not} enough to determine the
log-sized backdoor.  The properties just form the foundation of a
subtle interplay between the different settings of the backdoor
variables, exploiting exponentially growing UP implication chains on
branch $1$. For $n=8$, the backdoor is $\{
move\mbox{-}L^0\mbox{-}L_1^{1}(1),$
$move\mbox{-}L_1^{2}\mbox{-}L_1^{3}(3),$
$move\mbox{-}L_1^{6}\mbox{-}L_1^{7}(7) \}$.
Figure~\ref{XFIG/map-top-n8bd.eps} contains an illustration.

\vspace{-0.0cm} \widthfig{8.0}{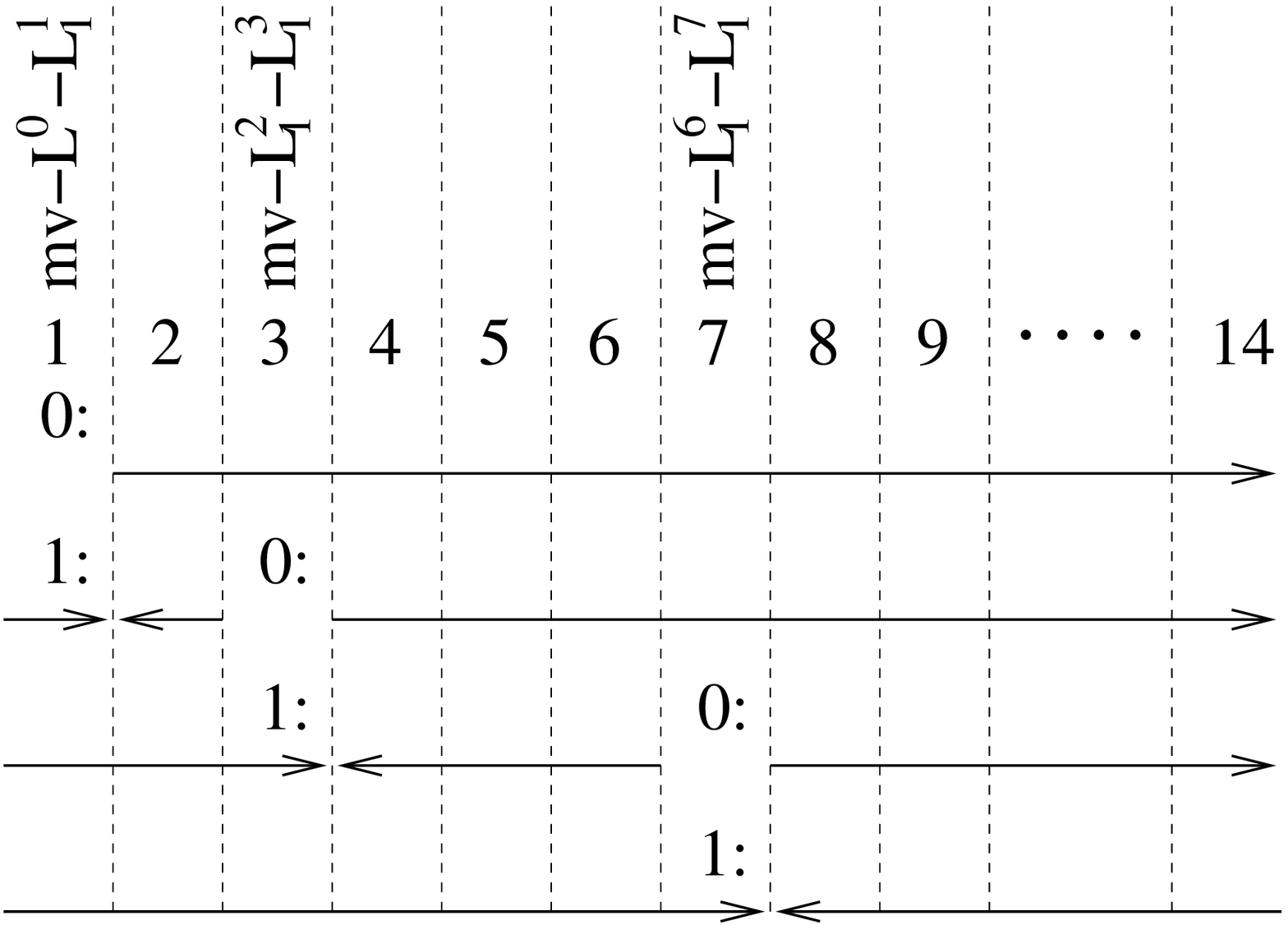}{The workings
  of the optimal backdoor for $MAP_8^{13}$. Arrows indicate moves on
  the $L_1$-branch forced to $1$ by UP. Direction $\rightarrow$ means
  towards $L_1^{13}$, $\leftarrow$ means towards $L^0$. When only a
  single open step is left, $move\mbox{-}L^0\mbox{-}L_2^{1}$ is forced
  to $1$ at that step by UP, yielding an empty clause.}{-0.0cm}

Consider the first (lowest) variable in the backdoor,
$move\mbox{-}L^0\mbox{-}L_1^{1}(1)$. If one sets this to $0$, then
property (2) applies: only $13$ of the $14$ available steps are left
to move towards the goal location $L_1^{13}$; UP recognizes this, and
forces moves towards $L_1^{13}$ at all steps $2 \leq t \leq 14$.
Since $t=1$ is the only remaining time step not occupied by a move
action, UP over the $L_2^1$ goal clause sets
$move\mbox{-}L^0\mbox{-}L_2^{1}(1)$ to $1$, yielding a
contradiction to the precondition clause of the move set to $1$ at
time $2$. So $move\mbox{-}L^0\mbox{-}L_1^{1}(1)$ must be set to
$1$.

Consider the second variable in the backdoor,
$move\mbox{-}L_1^{2}\mbox{-}L_1^{3}(3)$. Say one sets this to $0$. By
property (2) this forces moves at all steps $4 \leq t \leq 14$. So the
goal for $L_2^{1}$ must be achieved by an action at step $3$. But we
have committed to $move\mbox{-}L^0\mbox{-}L_1^{1}$ at step $1$.
This forces us to move back to $L^0$ at step $2$ and to move to
$L_2^{1}$ at step $3$. But then the move forced in earlier at $4$
becomes impossible. It follows that we must assign
$move\mbox{-}L_1^{2}\mbox{-}L_1^{3}(3)$ to $1$. With property (1),
this implies that, by UP, all time steps below $3$ get occupied with
move actions. (Precisely, in our case here,
$move\mbox{-}L_1^{1}\mbox{-}L_1^{2}(2)$ is also set to $1$.)

Consider the third variable in the backdoor,
$move\mbox{-}L_1^{6}\mbox{-}L_1^{7}(7)$. If we set this to $0$, then
by property (2) moves are forced in by UP at the time steps $8 \leq t
\leq 14$. So, to achieve the $L_2^{1}$ goal at step $7$, we have to
take {\bf three steps to move back from $L_1^{3}$ to $L^0$}: steps
$4$, $5$, and $6$. A move to $L_2^{1}$ is forced in at step $7$, in
contradiction to the move at $8$ forced in earlier. Finally, if we
assign $move\mbox{-}L_1^{6}\mbox{-}L_1^{7}(7)$ to $1$, then by
property (1) moves are forced in by UP at all steps below $7$. We need
{\bf seven steps to move back from $L_1^{7}$ to $L^0$}, and an
eighth step to get to $L_2^{1}$. But we have only the $7$ steps $8,
\dots, 14$ available, so the goal for $L_2^{1}$ is unachievable.

The key to the logarithmic backdoor size is that, to achieve the
$L_2^{1}$ goal, we have to move back from $L_1^{t}$ locations we
committed to earlier (as indicated in bold face above for $t=3$ and
$t=7$). We committed to move to $L_1^{t}$, and the UP propagations
force us to move back, thereby occupying $2*t$ steps in the encoding.
This yields the possibility to double the value of $t$ between
variables.

The DPLL tree for $MAP_n^{2n-3}$ degenerates to a line:

\begin{cor}[MAP asymmetrical case, DPLL UB]\label{map:dpll:top}
  For $MAP_n^{2n-3}$, there is a DPLL refutation of size $2*\lceil
  log_2 n \rceil+1$.
\end{cor}

\begin{sloppypar}
\begin{proof}
  Follows directly from the proof to Theorem~\ref{maptop}: If one
  processes the $MAP_n^{2n-3}B$ variables from $t=1$ upwards, then,
  for every variable, assigning the value $0$ yields a contradiction
  by UP. There are $\lceil log_2 n \rceil$ non-failed search nodes.
  Adding the failed search nodes, this shows the claim.
\end{proof}
\end{sloppypar}

This is an interesting result since it reflects the intuition that, in
practical examples, constraint propagation can cut out many search
branches early on, yielding a nearly degenerated search tree. This
phenomenon is also visible in our empirical data regarding search tree
size/depth ratio, c.f. Section~\ref{real}, Figure~\ref{gnu:ratio}.

Note that we have now shown a doubly exponential gap between the sizes
of the best-case DPLL refutations in the symmetrical case and the
asymmetrical case. It would be interesting to determine what the
optimal backdoors are in general, i.e., in $MAP_n^k$, particularly at
what point the backdoors become logarithmic. Such an investigation
turns out to be extremely difficult. For interesting combinations of
$n$ and $k$, it is practically impossible to find the optimal
backdoors empirically, and so get a start into the theoretical
investigation. We developed an enumeration program that exploits
symmetries in the planning task to cut down on the number of variable
sets to be enumerated. Even with that, the enumeration didn't scale up
far enough. We leave this topic for future work.

%% file: SBW.tex
\subsection{SBW} 
\label{sd:sbw}

\begin{sloppypar}
  We also constructed and examined a structured version of the
  Blocksworld planning domain, with ``stacking restrictions'' on what
  blocks can be stacked onto what other blocks. The parameter $n$ is
  the number of blocks. They are initially all located side-by-side on
  a table $t_1$.  The goal is to bring all blocks onto another table
  $t_2$, that has only space for a single block. That is, the $n$
  blocks must be arranged in a single large stack on top of $t_2$.
  The parameter $k$ defines the amount of {\em stacking restrictions}.
  There are $0 \leq k \leq n$ ``bad'' blocks $b_1, \dots, b_k$ and
  $n-k$ ``good'' blocks $g_1, \dots, g_{n-k}$. For $1 < i \leq k$,
  $b_i$ can only be stacked onto $b_{i-1}$; $b_1$ can be stacked onto
  $t_2$ and any $g_i$. The $g_i$ can be stacked onto each other, and
  onto $t_2$. See an illustration in Figure~\ref{XFIG/sbw-illus.eps}.
\end{sloppypar}

\widthfig{8}{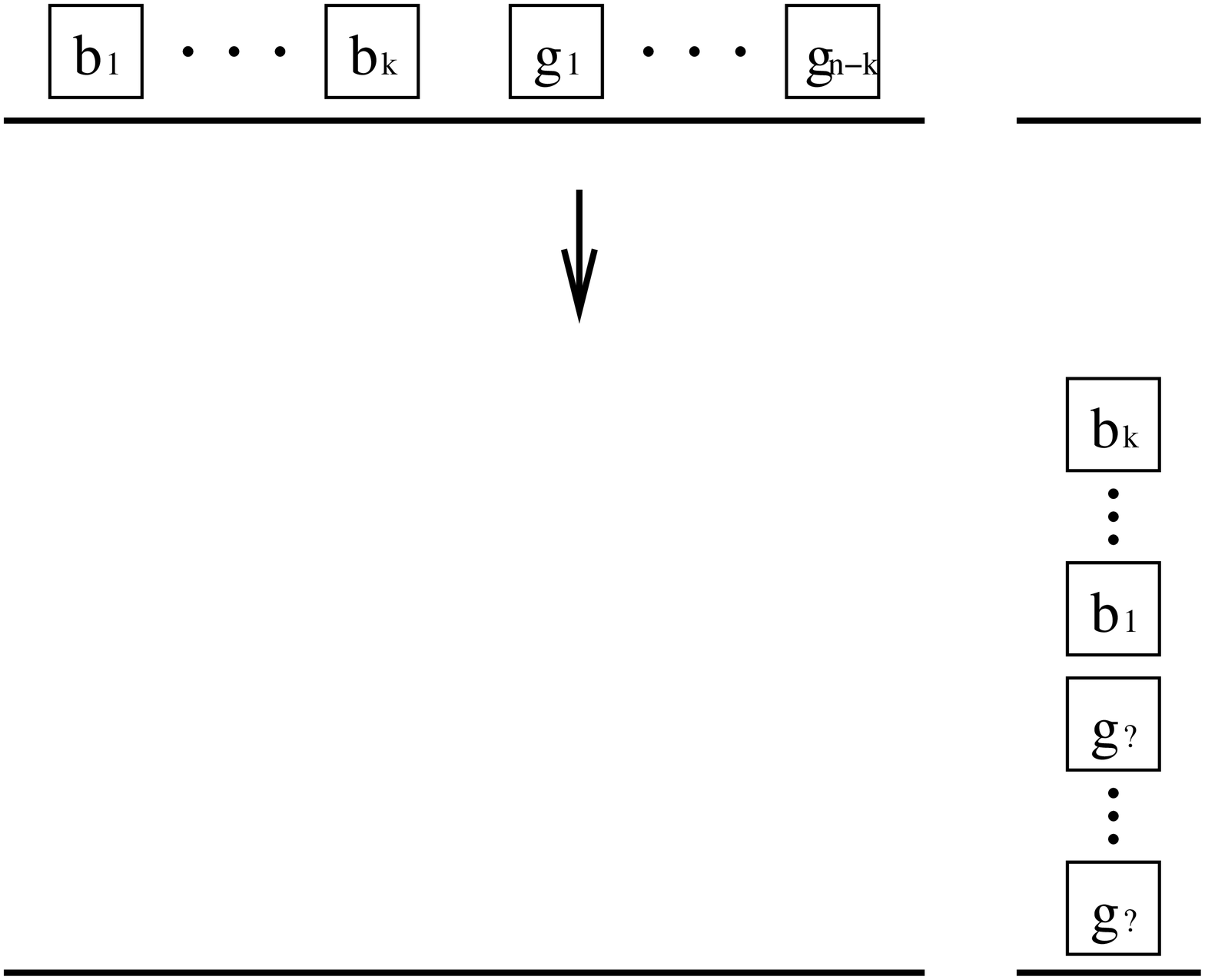}{An (incomplete) illustration of the SBW
  domain.}{-0.0cm}

The operators are the usual Blocksworld operators of the domain
version that does not make use of an explicit robot arm, i.e. the
predicates dealt with are ``on'', ``on-table'', and ``clear'' (as well
as some supplementary stuff to implement the stacking restrictions and
the goal). More precisely, there are two kinds of moving actions:
those that move a block $x$ from the table to a block $y$ that is
above $t_2$ (a la Move-x-from-t-to-t2, short ``movetot2-x-y''), and
that move a block $x$ from a block $y$ above $t_2$, or from $t_2$
itself, to $t_1$ (a la Move-x-from-t2-to-t, short ``movefromt2-x-y'').
The former action adds a fact stating that now $x$ is ``above'' $t_2$,
the latter action deletes that fact. Initially only $t_2$ is ``above''
itself.  The goal condition is the conjunction of ``above'' for all
blocks. As in MAP, a very basic feature in our proofs is that every
pair of move actions is incompatible, and so setting a move at time
$t$ to $1$ excludes, by UP, all other move actions at $t$. We use the
convention that $g_0$ denotes $t_2$, i.e. $t_2$ is the ``lower most
good block'' -- this is useful to simplify the notation.

The optimal plan length in SBW is $n$, independently of $k$: one has
to first move the good blocks above $t_2$ in any order, and then the
bad blocks in increasing order (of index). $\ASratio$ is $\frac{1}{n}$
if $k=0$, and $\frac{k}{n}$ otherwise -- in order to get $b_k$ above
$t_2$, all of $b_1, \dots, b_{k-1}$ must be moved there first. As
before, we consider the unsatisfiable CNFs that are just a single step
short of a solution: we have $n-1$ time steps $1, \dots, n-1$.

With $n=1$, optimal plan length is $1$ so our CNF encoding would have
$0$ steps and be empty. It is easy to see that, with $n=2$, even with
$k=0$ $SBW_{n,k}$ is solved by UP -- there is just a single time step,
into which two moving actions must be fit. It is also easy to see that
the CNF contains an empty clause -- no goal achievers are present --
for $SBW_{n,n}$, since only $n-1$ steps are available to stack the $n$
blocks $b_i$ in sequence. Also, $SBW_{n,n-1}$ is solved by UP because
there is only just enough time to stack the sequence of bad blocks;
each time step is assigned such a stacking action, and there is no
time to stack the single good block. Altogether, in what follows we
thus assume that $n \geq 3$, and $0 \leq k \leq n-2$. Since, with high
$k$, there are less options of stacking one block onto another, the
number of variables in the CNF differs considerably for different
values of $k$.  Precisely, that number is $3n^3 - 5n^2 - 1$ for $k=0$,
and $4n^2 + 13n - 44$ for $k=n-2$. At both ends of the $k$ scale, for
any constant $b$, the $b$-cutset size is lower bounded by a linear
function in the respective number of variables; details on this are in
Appendix~\ref{cutsets}.

We were not able to prove an exponential lower bound on DPLL
refutation size for the symmetrical case (bottom end of the $k$ scale)
of SBW. We conjecture that there is such a bound, and pose this
question as an open problem. Here, we identify SBW backdoors.  Similar
as in the MAP family of formulas, in the symmetrical case the smallest
backdoors we could identify have a size linear in the total number of
variables.  Precisely, the backdoor we identified, called $SBW_n^0B$,
is defined as follows: {\small
\medskip
\begin{tabbing}
  $SBW_n^0B$ := \=  $\{movetot2\mbox{-}g_i\mbox{-}g_j(t) \mid 1 \leq i \leq n-2, 
  0 \leq j \leq n, j \neq i, 2 \leq t \leq n-1\} \; \; \setminus$\\ 
  \> $\{movetot2\mbox{-}g_i\mbox{-}g_j(i+1) \mid max(2,n-4) \leq i \leq n-2, 0 \leq
  j \leq n-2\}$
\end{tabbing}}
\medskip
\noindent
In words, to construct $SBW_n^0B$ we first include, for all blocks
$g_i$, $i \in \{1, \dots, n-2\}$, all possibilities of getting $g_i$
above $t_2$, at all time steps except the first one. Thereafter, for
all blocks $g_i$, $i \in \{max(2,n-4), \dots, n-2\}$, that is, for the
last $3$ blocks in the set $\{2, \dots, n-2\}$ (or less if $|\{2,
\dots, n-2\}| < 3$), we subtract all possible moves except those to 
$g_{n-1}$ and $g_n$. We refer to these move variables as {\em
cut-fields}. Figure~\ref{XFIG/sbw-base-newBD.eps} illustrates the
backdoor for $n=7$.
The size of $SBW_n^0B$ is $\Theta(n^3)$: at $\Theta(n)$ time steps, we
talk about $\Theta(n)$ blocks, each possibly being stacked onto
$\Theta(n)$ other blocks.\footnote{Precisely, the size is $n^3 - 5n^2
  + 9n - 6$ for $n \leq 6$, and $n^3 - 4n^2 + n + 6$ for $n > 6$.}
Since the total number of variables in the formula is also
$\Theta(n^3)$, we have a linear size variable subset.

\widthfig{13}{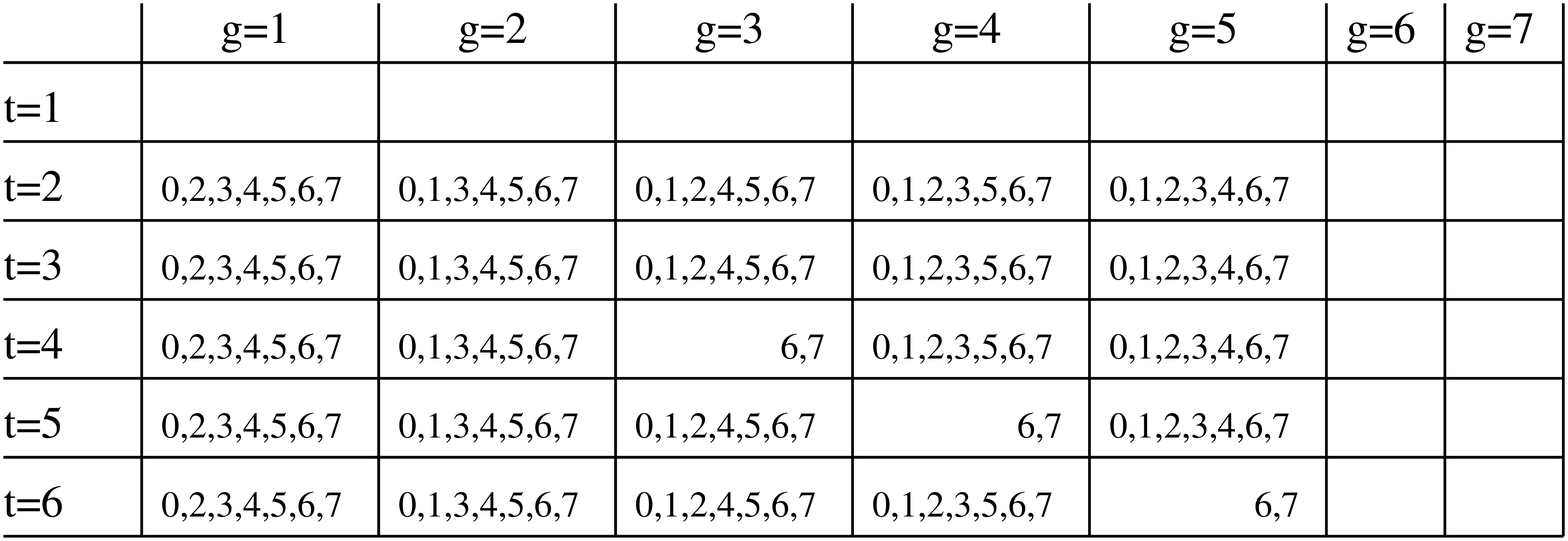}{An illustration of $SBW_n^0B$
  for $n=7$. The vertical axis corresponds to time steps $t$, the
  horizontal axis corresponds to blocks $g$. For block $g=1$, i.e.\
  $g_1$, the notations $0,2,3,4,5,6,7$ are shorthand for
  $movetot2\mbox{-}g_1\mbox{-}t_2$, $movetot2\mbox{-}g_1\mbox{-}g_2$,
  $movetot2\mbox{-}g_1\mbox{-}g_3$, $movetot2\mbox{-}g_1\mbox{-}g_4$,
  $movetot2\mbox{-}g_1\mbox{-}g_5$, $movetot2\mbox{-}g_1\mbox{-}g_6$,
  $movetot2\mbox{-}g_1\mbox{-}g_7$. Likewise for the other
  blocks.}{-0.5cm}

\begin{thm}[SBW symmetrical case, BD]\label{sbwbottom}
  $SBW_n^0B$ is a backdoor for $SBW_n^0$.
\end{thm}

We remark that it is very easy to prove that $SBW_n^0B$ is a backdoor
if one does not remove any variables from the upper set in the above
definition, i.e., if one considers the set
$\{movetot2\mbox{-}g_i\mbox{-}g_j(t) \mid 1 \leq i \leq n-2, 0 \leq j
\leq n, j \neq i, 2 \leq t \leq n-1\}$.  The difficult bit is to prove
that the cut-fields can be removed without losing the backdoor
property. We proved empirically that $SBW_n^0B$ is optimal for $n=3$,
and that it is minimal for $n \leq 6$. It is {\em not} minimal for
$n=7$ (and, presumably, for larger values of $n$). There are
apparently certain subtleties that appear only in the larger
instances, and due to that a few more variables can be skipped from
$SBW_n^0B$. For $n=7$, $3$ out of $160$ variables can be removed from
$SBW_n^0B$ until the backdoor set is minimal. Since the proof to
Theorem~\ref{sbwbottom} is already rather complicated, and the number
of additional variables that can be removed appears to be very small,
we did not investigate this in detail. The important message is that
the backdoor in the symmetrical case is in the order of $n^3$, talking
about a linear number of blocks being put onto a linear number of
other blocks at a linear number of time steps, and that a few
variables can be spared due to some more subtle phenomena (e.g., see
below).

To prove Theorem~\ref{sbwbottom}, one has to reason about how the goal
constraints for the single blocks are transported backwards over the
time steps in the CNF, starting from the last (goal) time step. In
that sense, the style of the proof is reminiscent of the proof for the
MAP symmetrical case, Theorem~\ref{mapbottom}. However, in SBW with
$k=0$, we have $n$ possibilities to achieve each goal (rather than the
single possibility we have in MAP). We can stack any block $g$ onto
any other block $g'$, or onto $t_2$, to achieve the goal for $g$.
Still, when not removing the cut-fields, the proof is relatively
simple and comes down to a reasoning that says: ``After UP, at least
$n-2$ time steps will be occupied with moves towards $t_2$; the
remaining two blocks are forced into the single remaining time step,
which causes a contradiction.'' If we do remove the cut-fields,
however, then one has to take account of quite a number of subtleties
caused by the structure of the domain, and how UP recognizes the
restrictions imposed by them. Just to give one example, under certain
circumstances, if a variable sequence of the form
$movetot2\mbox{-}g_1\mbox{-}g_0(1),$
$movetot2\mbox{-}g_2\mbox{-}g_1(2), \dots,$
$movetot2\mbox{-}g_{k}\mbox{-}g_{k-1}(k)$ (or any other permutation of
a $k$-subset) are set to $1$, then at the time steps above $k+1$ only
moves onto $g_k$, or blocks not in the committed sequence, are
possible -- that is, all other moves are forced to $0$ by UP. The
overall proof structure is to perform case distinctions over the
possible situations after assigning the variables regarding those
blocks that have no cut-field.

In the asymmetrical case (top end of the $k$ scale, $k=n-2$), as
before a logarithmic number of decision variables suffices. The
backdoor we identified, called $SBW_n^{n-2}B$, is the following:
{\small
\medskip
\begin{tabbing}
$SBW_n^{n-2}B$ := \=  $\{movetot2\mbox{-}g_1\mbox{-}t_2(1),
movetot2\mbox{-}g_2\mbox{-}t_2(1)\} \; \; \cup$\\
\> $\{ movetot2\mbox{-}b_{3*2^{i-1}-1}\mbox{-}b_{3*2^{i-1}-2}(3*2^{i-1}-1)
\mid 1 \leq i \leq \lceil log_2(n/3) \rceil\}$
\end{tabbing}}
\medskip
\noindent
Table~\ref{tab:map-bd-top} illustrates this for up to $n=48$.

\begin{table}[htb]
\begin{center}
{\small
\begin{tabular}{|l|l|c|}\hline
$n$      &   $i$     &  (additional) backdoor variable(s) \\\hline\hline
$3 = 3*2^{i}$      &   $0$     & \hspace{1.0cm}  $movetot2\mbox{-}g_1\mbox{-}t_2(1)$  \hspace{1.0cm} \\
        &        &  $movetot2\mbox{-}g_2\mbox{-}t_2(1)$  \\\hline
$4$      &   $1$     &  $movetot2\mbox{-}b_2\mbox{-}b_1(2)$ \\
$5$      &           &                           \\
$6 = 3*2^{i}$      &           &                           \\\hline
$7$      &   $2$     &  $movetot2\mbox{-}b_5,b_4(5)$ \\
 .       &           &                           \\
 .       &           &                           \\
$12 = 3*2^{i}$     &           &                           \\\hline
$13$      &   $3$     &  $movetot2\mbox{-}b_{11}\mbox{-}b_{10}(11)$ \\
 .       &           &                           \\
 .       &           &                           \\
$24 = 3*2^{i}$     &           &                           \\\hline
$25$      &   $4$     &  $movetot2\mbox{-}b_{23}\mbox{-}b_{22}(23)$ \\
 .       &           &                           \\
 .       &           &                           \\
$48 = 3*2^{i}$     &           &                           \\\hline
\end{tabular}}
\vspace{0.3cm}
\caption{\label{tab:map-bd-top} An illustration of $SBW_n^{n-2}B$, 
for $n \leq 48$. Index $i$ indicates the number of variables added on
  top of $movetot2\mbox{-}g_1\mbox{-}t_2(1)$ and
  $movetot2\mbox{-}g_1\mbox{-}t_2(1)$. This is basically the index of
  the exponentially growing ``equivalence classes'' of subsequent
  formulas with the same backdoor set.}
\end{center}
\vspace{-0.5cm}
\end{table}

\begin{thm}[SBW asymmetrical case, BD]\label{sbwtop}
  $SBW_n^{n-2}B$ is a backdoor for $SBW_n^{n-2}$.
\end{thm}

The size of $SBWn^{n-2}B$ is $2 + \lceil log_2(n/3) \rceil$.  We
verified this as a lower bound for up to $n=8$, and we conjecture that
it is a lower bound in general.
We proved that $SBW_n^{n-2}B$, is minimal.

\begin{thm}[SBW asymmetrical case, BD minimality]\label{sbwtop:upcons}
  Let $n>2$. Let $B'$ be a subset of $SBW_n^{n-2}B$ obtained by
  removing one variable.  Then there is exactly one UP-consistent
  assignment to the variables in $B'$.
\end{thm}

The reason for the existence of the logarithmic size backdoors is,
similarly to before, that UP can exploit long implication chains
between variables at time steps whose distance doubles between each
pair of variables.  The variables (except those regarding the two good
blocks) encode commitments regarding moves of bad blocks at
appropriate time steps. The interplay of the variables, and our proof
of their backdoor property, i.e., the proof of Theorem~\ref{sbwtop},
is quite reminiscent of the MAP asymmetrical case (i.e.
Theorem~\ref{maptop}).  By induction over $i$, each variable
$movetot2\mbox{-}b_{3*2^{i-1}-1}\mbox{-}b_{3*2^{i-1}-2}(3*2^{i-1}-1)$
must be set to $1$ or else UP enforces moves ``towards the goal'' at
$1, \dots, 3*2^{i-2}-1$, moves ``away from the goal'' at $3*2^{i-2},
\dots, 3*2^{i-1}-2$, and moves ``towards the goal'' at $3*2^{i},
\dots, n-1$. Setting the topmost (at the highest time step) variable
in the backdoor to $1$ yields a contradiction because UP enforces
moves ``away from the goal'' at all higher time steps. In difference
to the MAP asymmetrical case, the time-step distance between the
backdoor variables contains a factor $3$. This is due to {\em one
  step} needed for scheduling the two good blocks (similar to MAP),
{\em plus two steps} needed to move (remove) $b_2$ and $b_1$ to (from)
$t_2$ (dissimilar to MAP).

As before, proving Theorem~\ref{sbwtop:upcons} is a matter of figuring
out what can go wrong in the proof to Theorem~\ref{sbwtop}, after
removing one variable. Also as before, we get a logarithmic-size
best-case DPLL refutation.

\begin{cor}[SBW asymmetrical case, DPLL UB]\label{sbw:dpll:top}
  For $SBW_n^{n-2}$, there is a DPLL refutation of size $2*(2+\lceil
  log_2(n/3) \rceil)+1$.
\end{cor}

\begin{proof}
  From the proof to Theorem~\ref{sbwtop}, it follows directly that
  there is a DPLL refutation, in the shape of a line, with $2+\lceil
  log_2(n/3) \rceil$ non-failed search nodes. Adding the failed search
  nodes, this shows the claim.
\end{proof}

As in MAP, we consider it interesting, but found it extremely
difficult, to determine what the optimal backdoors are in general, in
particular at what point of the $k$ scale they become logarithmic. We
leave this topic open for future work.

%% file: SPH.tex
\subsection{SPH} 
\label{sd:sph}

In $SPH_n^k$, like in the classical pigeon hole problem the task is to
assign $n+1$ pigeons to $n$ holes. The difference lies in that there
is now one ``bad'' pigeon that requires $k$ holes, and $k-1$ ``good''
pigeons that can share a hole with the bad pigeon. The remaining
$n-k+1$ pigeons are normal, i.e., need one hole each. The range of $k$
is between $1$ and $n$. Independently of $k$, $n+1$ holes are needed
overall. In particular the CNF is unsatisfiable.

Precisely, the CNF is the following. The variables are $x\_y$ where
$x$ is a pigeon and $y$ is a hole: $x\_y$ set to $1$ means that $x$ is
assigned to $y$. The bad pigeon is $x=0$, the good pigeons are $x=1,
\dots, x=k-1$, the normal pigeons are $x=k, \dots, x=n$. The holes are
$y=1, \dots, n$. The clauses are:

\begin{itemize}
\item $\{x\_1, \dots, x\_n\}$ for all $x \neq 0$: all pigeons except
  the bad one need one hole.
\item $\{\neg x\_y, \neg x'\_y\}$ for all $x, x' \neq 0$, $x < x'$,
  and all $y$: no two normal or good pigeons can share a hole.
\item $\{\neg 0\_y, \neg x\_y\}$ for all $x \geq k$ and all $y$: none
  of the normal pigeons can share a hole with the bad pigeon.
\item $GEQ(\{0\_1, \dots, 0\_n\}, k)$: a set of clauses that is
  satisfiable iff at least $k$ variables out of the set $\{0\_1,
  \dots, 0\_n\}$ are set to $1$. We consider two options for the
  definition of $GEQ(\{0\_1, \dots, 0\_n\}, k)$, see below.
\end{itemize}

The most straightforward, {\em naive}, definition of $GEQ(\{0\_1,
\dots, 0\_n\}, k)$ is the set of clauses $\{0\_y \mid y \in Y\}$ for
all $Y \subseteq \{1, \dots, n\}, |Y| = n-k+1$. To explain this,
observe that the size of the sets $Y_0$ is chosen so that, when
removing the holes $Y_0$ from $Y$, only $k-1$ holes are left. So if
$p_0$ occupies $k$ holes then, out of each set $Y_0$, $p_0$ occupies
at least one hole. The other direction, if $p_0$ occupies only (at
most) $k-1$ holes, then the complement of these holes contains a set
$Y_0$ where the respective clause evaluates to $0$.

The drawback of the naive definition of $GEQ(\{0\_1, \dots, 0\_n\},
k)$ is that it can require an exponential number of clauses, e.g. if
one sets $k$ to $n/2$ and lets $n$ range. If one fixes $k$, or the
difference between $n$ and $k$, to some value, however (as e.g. at the
bottom/top ends of the $k$ scale where $k$ is fixed to $1$/$n-k$ is
fixed to $0$), then the number of clauses is polynomial (in $n$).  One
can obtain a definition of $GEQ(\{0\_1, \dots, 0\_n\}, k)$ that is
polynomial in both $n$ and $k$ by introducing additional auxiliary
variables, and connect them via appropriate clauses to implement a
counter of the holes the bad pigeon is assigned to. We discuss such an
encoding further below. For the moment, i.e.\ for the following formal
discussion, we assume the naive definition of $GEQ(\{0\_1, \dots,
0\_n\}, k)$ in the SPH formulas. Beside being more clear and easier to
understand, these formulas have the desirable property that they are a
generalization of the standard pigeon hole formula. If we set $k=1$
then we obtain exactly the standard formula using the single clause
$GEQ(\{0\_1, \dots, 0\_n\}, 1) = \{\{0\_1, \dots, 0\_n\}\}$ to ensure
that the bad pigeon is assigned to at least one hole.

For $k=n$, we obtain a formula that is inconsistent under UP:
$|Y|=n-k+1=1$ so we get the $n$ clauses $\{0\_1\} \dots \{0\_n\}$. In
words, the bad pigeon occupies all holes. By UP, the clause $\{n\_1,
\dots, n\_n\}$ becomes empty, i.e.\ there is no space for the last
remaining normal pigeon. For $k=n-1$, the CNF formula is consistent
under UP. Independently of $k$, the number of variables in $SPH_n^k$
is $(n+1)*n$. The constraint graph changes over $k$ (only) in that,
for all holes $1 \leq y \leq n$ and good pigeons $1 \leq g \leq k-1$,
there are no edges between $0\_y$ and $g\_y$, since the good and bad
pigeons can share holes (there are no exclusion clauses).
Independently of $k$, for any constant $b$, the $b$-cutset size is a
square function in $n$ -- linear in the total number of variables.
This is detailed in Appendix~\ref{cutsets}.

Since the $SPH_n^k$ formulas are much simpler than $MAP_n^k$ and
$SBW_n^k$, we were able to identify minimal backdoors across {\em the
entire range of $k$.} Precisely, the backdoor we identified, called
$SPH_n^k$, is the following:
 {\small
\medskip
\begin{tabbing}
$SPH_n^kB$ := \= $\{0\_y \mid 1 \leq y \leq n-1\} \; \; \cup$\\
\> $\{x\_y \mid k+2 \leq x \leq n, 1 \leq y \leq n-1\}$
\end{tabbing}}
\medskip
\noindent
In words, one selects $n-1$ of the holes (all but hole $n$) and
$n-k-1$ of the normal pigeons (all but numbers $k$ and $k+1$). The
variables talking about these holes and pigeons, plus the bad pigeon,
form the backdoor. Figure~\ref{XFIG/sph-BD.eps} illustrates
this.\footnote{We remark that, as one would expect, one can actually
select an arbitrary $n-1$ subset of the holes, and an arbitrary
$n-k-1$ subset of the normal pigeons, to form the backdoor. We fixed
one specific selection for simplicity of presentation.}

\widthfig{7}{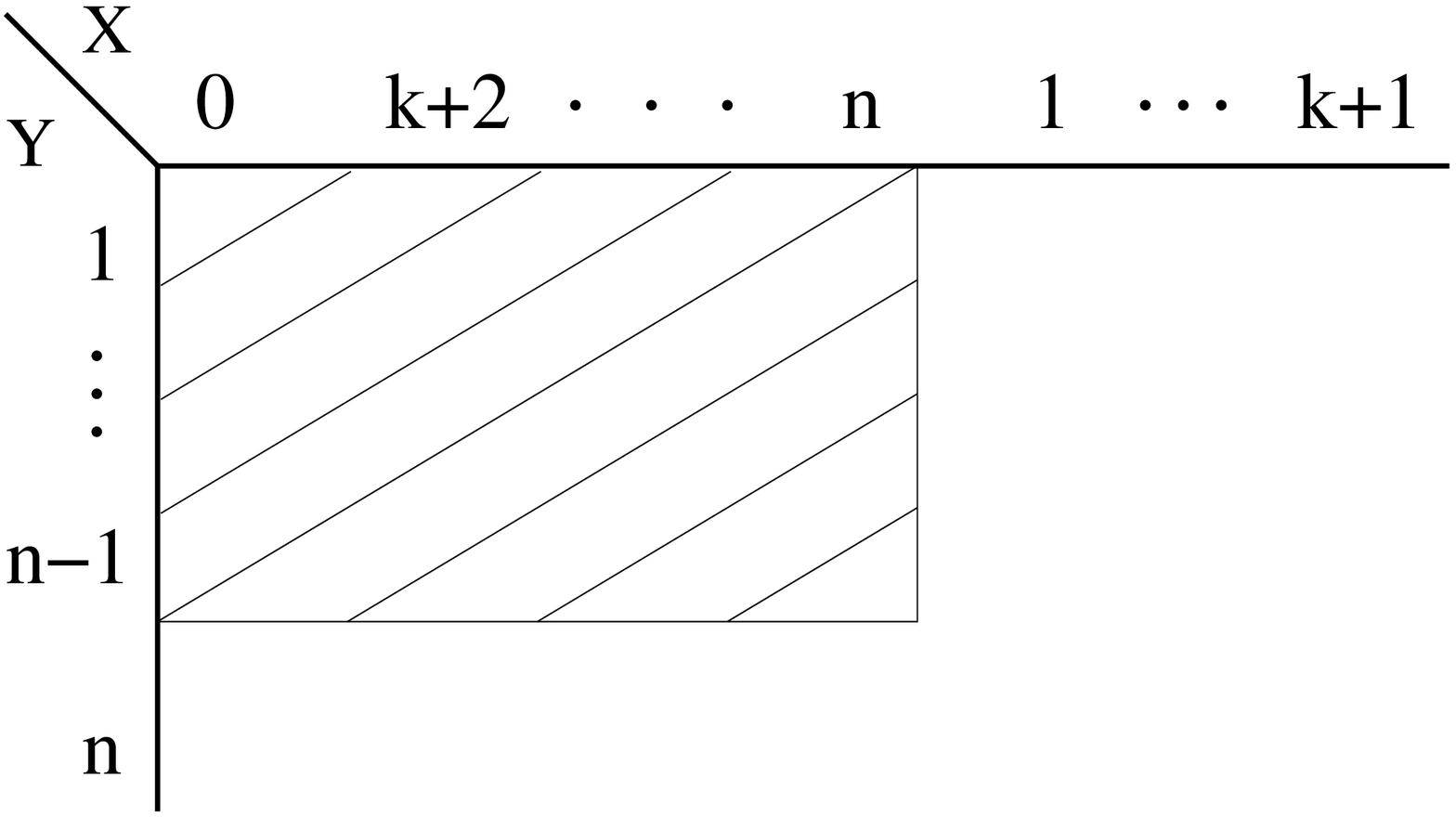}{An illustration of $SPH_n^kB$. The $x$ axis 
stands for the pigeons, the $y$ axis stands for the holes, and the
filled-in area indicates the set of backdoor variables (the
cross-product of the $n-1$ selected holes, and the $n-k-1$ selected
normal pigeons plus the bad one).}{-0.0cm}

\begin{thm}[SPH, BD]\label{sph}
  $SPH_n^kB$ is a backdoor in $SPH_n^k$.
\end{thm}

The size of $SPH_n^k$ is $(n-k)*(n-1)$. We conjecture that this is
also a lower bound on backdoor size; for $n \leq 5$, we verified this
empirically. The backdoor is minimal.

\begin{thm}[SPH, BD minimality]\label{sph:upcons}
  Let $B'$ be a subset of $SPH_n^kB$ obtained by removing one variable
  $x\_y$. Then, if $x \neq 0$, to the variables in $B'$ there are
  exactly $(n-2) * \dots * (k+1)$ UP-consistent assignments; if $x=0$
  then there are exactly $(n-2) * \dots * k$ UP-consistent
  assignments.
\end{thm}

\begin{sloppypar} 
Proving Theorem~\ref{sph} involves a number of case distinctions, but
is not overly difficult. The basis is the following ``efficiency
property'' of the naive encoding of $GEQ(\{0\_1, \dots, 0\_n\}, k)$:
{\em For any subset $Y'$, $|Y'| = n-k$, of holes, if $0\_y'$ is set to
  $0$ for all $y' \in Y'$, then $0\_y$ is set to 1 by UP for all $y
  \not \in Y'$.}  With this, it is easy to see that, after setting all
variables in the backdoor, at least $n-1$ holes are occupied by UP.
Then both remaining normal pigeons (those not selected for the
backdoor, numbers $k$ and $k+1$) are forced into the single remaining
hole, yielding an empty clause.  Proving Theorem~\ref{sph:upcons} is
largely a matter of combinatorics.  One first observes that a partial
assignment is UP-consistent iff it does not assign two pigeons (more
precisely, two pigeons except the bad and a good one) to the same
hole, and leaves at least two holes open. The rest of the proof is a
counting matter.
\end{sloppypar} 

As $k$ increases, the backdoor size goes down from a linear function
in the total number of variables -- $n(n+1)$ -- to a square root
function.  It is important to note that, between the DPLL refutations
induced by the backdoors (the DPLL refutations one gets when branching
over only the backdoor variables), there is an {\em exponential}
difference in size.  For $k=1$, the best-case refutation is
exponential in $n$ -- this is a special case of results by Buss and
Pitassi \cite{buss:pitassi:csl-97}. For $k=n-1$, we have the
following.

\begin{proposition}[SPH, DPLL UB]\label{sph:dpll:top}
  Any DPLL refutation of $SPH_n^{n-1}$, induced by $SPH_n^{n-1}B$, has
  size $2n-1$.
\end{proposition}

\begin{proof}
  For $k = n-1$, $SPH_n^{n-1}B = \{0\_y \mid 1 \leq y \leq n-1\}$. The
  bad pigeon needs $n-1$ holes, so as soon as one sets one of the
  backdoor variables to $0$, one gets a contradiction by UP (the bad
  pigeon is forced into the remaining $n-1$ holes, and the two normal
  pigeons are both forced into the single hole left open). So the DPLL
  tree contains non-failed nodes only for the search branches setting
  a (backdoor) variable to $1$, and the tree degenerates to a line, of
  length $n-1$, i.e., with $n-1$ non-failed search nodes.  Adding the
  failed search nodes, this shows the claim.
\end{proof}

Observe that the asymmetry behavior is unrelated to the number of
clauses: that number is low at the extreme ends ($k=1$ and $k=n-1$),
and increases/decreases (exponentially, in $n$) in between.

\begin{sloppypar}
  As opposed to the naive encoding of $GEQ(\{0\_1, \dots, 0\_n\}, k)$
  treated above, Bailleux and Boufkhad \cite{bailleux:boufkhad:cp-03}
  propose a polynomial-size encoding called $\Psi(\{0\_1, \dots,
  0\_n\},k,n)$ (in $\Psi(V,l,u)$, $u$ is an upper bound on the
  variables set to $1$). The encoding introduces $\Theta(nlogn)$ new
  variables, and $\Theta(n^2)$ clauses, to implement a tree-shaped
  counting procedure that creates a vector of $n$ output variables
  where all holes assigned to the bad pigeon are moved to one
  side.\footnote{I.e. in any satisfying assignment to $\Psi(\{0\_1,
    \dots, 0\_n\},k,n)$ the new variables behave in this way.} One can
  then simply check, by looking at that side of the output variables,
  if the bad pigeon occupies enough holes.  Theorem~\ref{sph} remains
  valid because, as proved by Bailleux and Boufkhad, $\Psi(\{0\_1,
  \dots, 0\_n\},k,n)$ has the same efficiency property as mentioned
  above for the naive encoding. The question is whether the backdoors
  can become smaller. We empirically found that this can indeed be the
  case. For the case $n=4$, $k=1$, our enumeration procedure found
  backdoors of size $8$, rather than the $9$ variables necessary in
  the naive encoding. The question is, exactly what shape do the new
  optimal backdoors have? We performed a few tests and came up with
  some hypotheses, but didn't get very far because the interesting
  things happened only for values of $n$ and $k$ far beyond the reach
  of complete enumeration. We leave this topic open for future work.
\end{sloppypar}

%% file: empirical.tex
\section{Empirical Behavior in Synthetic Domains}
\label{empirical}

\vspace{-0.0cm}
\begin{figure*}
  \begin{center}
    \begin{tabular}{cc}\vspace{-1.1cm}
    \hspace{-1.3cm}  \includegraphics[width=9.5cm]{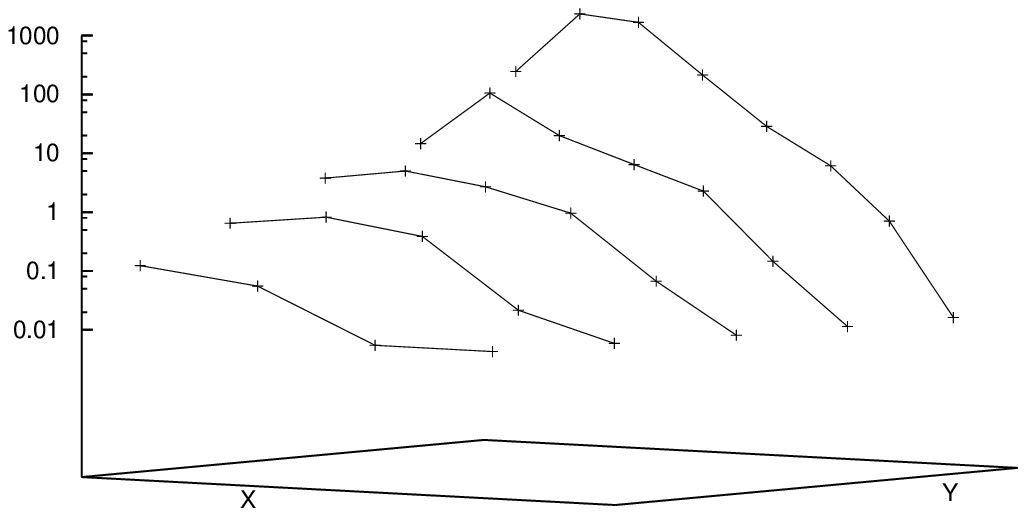} \hspace{-2.4cm}&\hspace{-2.4cm} \includegraphics[width=9.5cm]{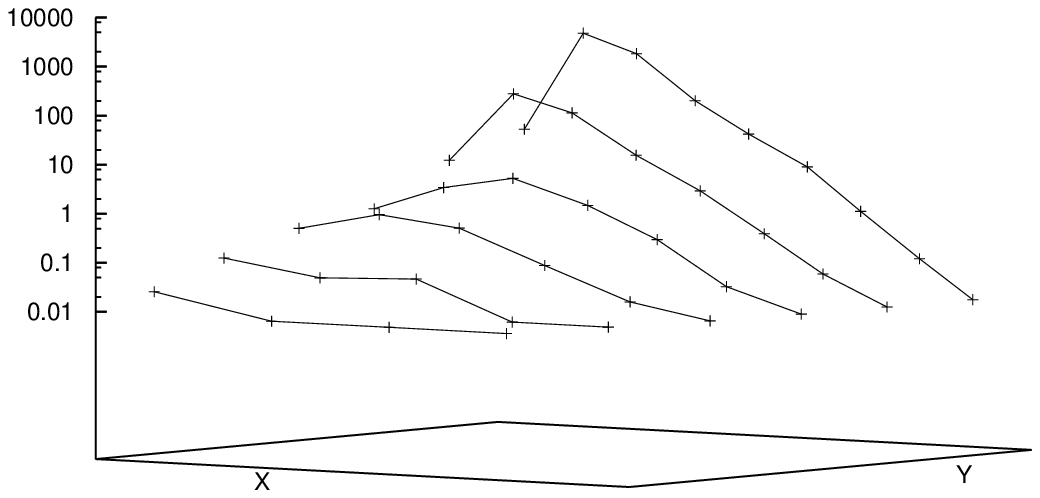}\\\vspace{-0.4cm}
 \hspace{-1.3cm}     (a) \zchaff\ in MAP \hspace{-2.4cm}&\hspace{-2.4cm} (b) \minisat\ in MAP\\\vspace{-1.1cm}
\hspace{-1.3cm}\includegraphics[width=9.5cm]{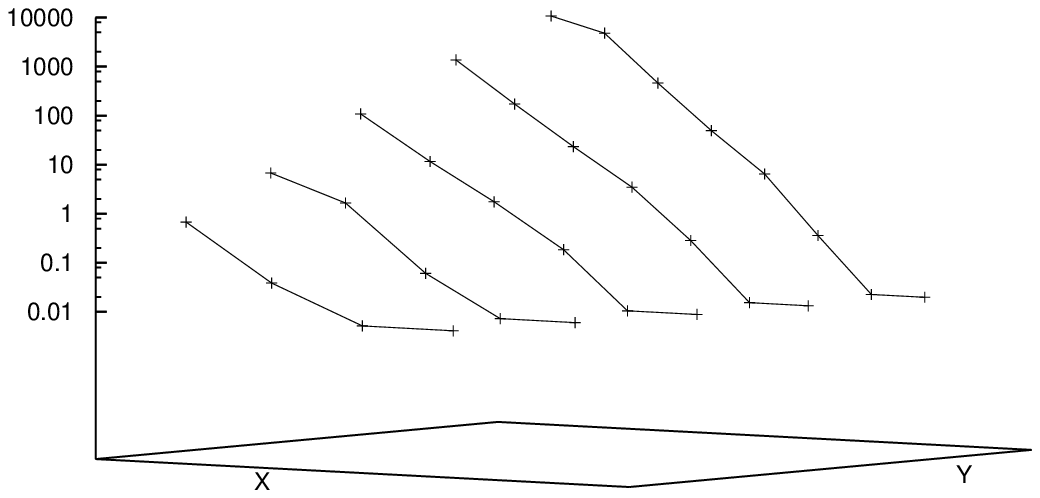} \hspace{-2.4cm}&\hspace{-2.4cm} \includegraphics[width=9.5cm]{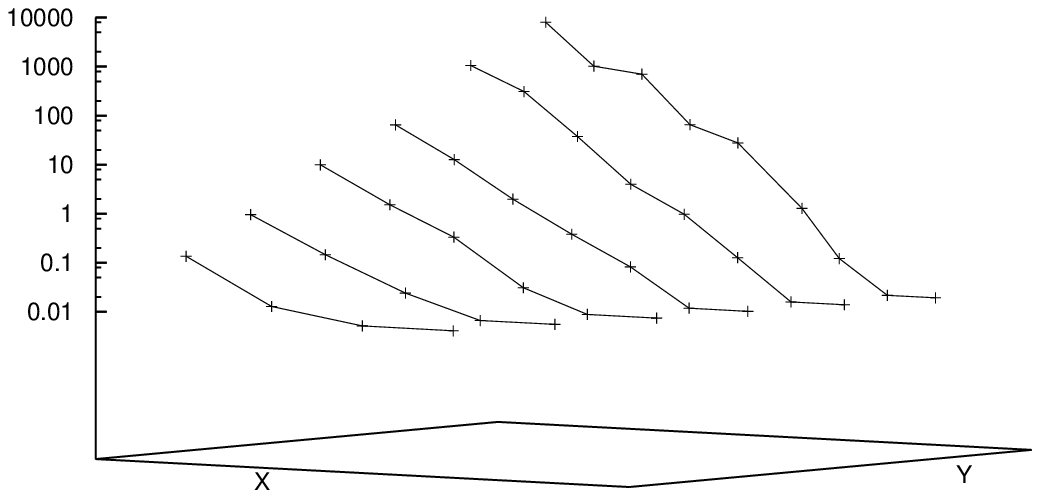}\\\vspace{-0.4cm}
\hspace{-1.3cm}      (c) \zchaff\ in SBW \hspace{-2.4cm}&\hspace{-2.4cm} (d) \minisat\ in SBW\\\vspace{-1.1cm}
\hspace{-1.3cm} \includegraphics[width=9.5cm]{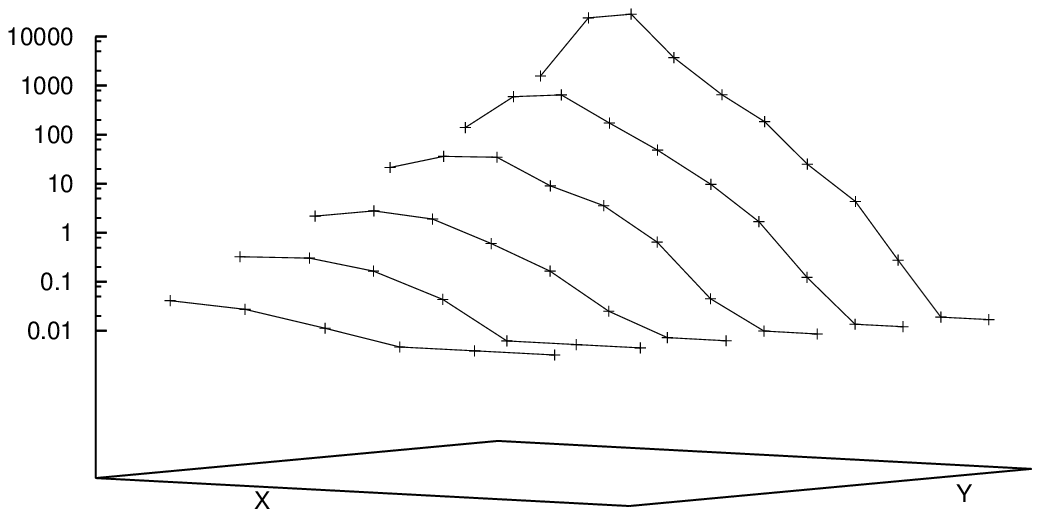} \hspace{-2.4cm}&\hspace{-2.4cm} \includegraphics[width=9.5cm]{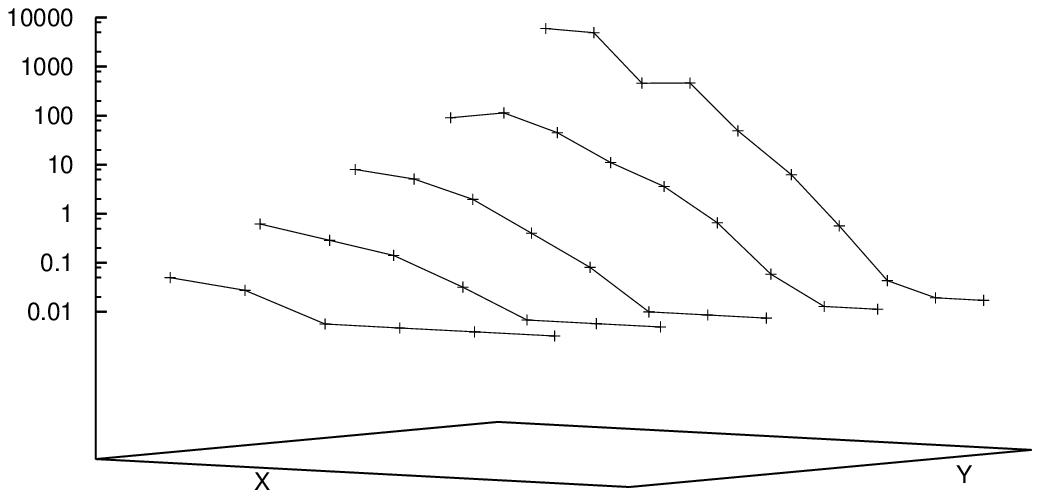}\\
\hspace{-1.3cm}      (e) \zchaff\ in SPH \hspace{-2.4cm}&\hspace{-2.4cm} (f) \minisat\ in SPH\\
    \end{tabular}
  \end{center}
  \vspace{-0.0cm}
  \caption{\label{sd:empirical}Runtime of \zchaff\ and \minisat,
    log-plotted against $\ASratio$ (axes labelled ``X'') and $n$ (axes
    labelled ``Y''), in our synthetic CNFs. The range of $n$ depends
    on domain and SAT solver, see text; the value grows from front to
    back. The range of $\ASratio$ is $[0;1]$ in all plots, growing
    from left to right. For each value of $n$, the data points shown
    on the X axis correspond to the $\ASratio$ values for all possible
    settings of $k$.}
\end{figure*}

Regarding the analysis of synthetic domains, it remains to verify
whether the theoretical observations carry over to practice -- i.e.,
to verify whether state-of-the-art SAT solvers do indeed behave as
expected. To confirm this, we ran \zchaff\
\cite{moskewicz:etal:dac-01} and \minisat\
\cite{een:soerensson:sat-03,een:biere:sat-05} on our synthetic CNFs.
The results are in Figure~\ref{sd:empirical}. 

The axes labelled ``X'' in Figure~\ref{sd:empirical} refer to
$\ASratio$. The axes labelled ``Y'' refer to values of $n$. The range
of $\ASratio$ is $0$ to $1$ in all pictures, growing on a linear scale
from left to right. For each value of $n$, the data points shown on
the X axis correspond to the $\ASratio$ values for all possible
settings of $k$. The range of $n$ depends on domain and SAT solver; in
all pictures, $n$ grows on a linear scale from front to back. We
started each $n$ range at the lowest value yielding non-zero runtimes,
and scaled until the first time-out occurred, which we set to two
hours (we used a PC running at 1.2GHz with 1GB main memory and 1024KB
cache running Linux). The plots in Figure~\ref{sd:empirical} include
all values of $n$ for which {\em no} time-out occurred. With this
strategy, in MAP, for \zchaff\ we got data for $n=5, \dots, n=9$, and
for \minisat\ we got data for $n=5, \dots, n=10$. In SBW, for \zchaff\
we got data for $n=6, \dots, n=10$, and for \minisat\ we got data for
$n=6, \dots, n=11$. In SPH, for \zchaff\ we got data for $n=7, \dots,
n=12$, and for \minisat\ we got data for $n=7, \dots, n=11$. (Note
that \minisat\ outperforms \zchaff\ in MAP and SBW, while \zchaff\ is
better in SPH.)

All in all, the empirical results comply very well with our analytical
results, meaning that the solvers do succeed, at least to some extent,
in exploiting the problem structure and finding short proofs. When
walking towards the back on a parallel to the $Y$ axis -- when
increasing $n$ -- as expected runtime grows exponentially in all but
the most asymmetric (rightmost) instances, for which we proved the
existence of polynomial (logarithmic, in MAP and SBW) proofs. Also,
when walking to the left on a parallel to the $X$ axis -- when
decreasing $\ASratio$ -- as expected runtime grows exponentially.
There are a few remarkable exceptions to the latter, particularly for
\zchaff\ and \minisat\ in MAP, and for \zchaff\ in SPH: there, the
most symmetric (leftmost) instances are solved faster than their
direct neighbors to the right. There actually also is such a
phenomenon in SBW. This is not visible in Figure~\ref{sd:empirical}
since $k=0$ and $k=1$ lead to the same $\ASratio$ value in SBW, and
Figure~\ref{sd:empirical} uses $k=0$; the $k=1$ instances take about
$20\%-30\%$ more runtime than the $k=0$ instances. So, the SAT solvers
often find the completely symmetric cases easier than the only
slightly asymmetric ones. We can only speculate what the reason for
that is. Maybe the phenomenon is to do with the way these SAT solvers
perform clause learning; maybe the branching heuristics become
confused if there is only a very small amount of asymmetry to focus
on. It seems an interesting topic to explore this issue -- in the
``slightly asymmetric'' cases, significant runtime could be saved.

%% file: strawman.tex
\section{A Red Herring}
\label{strawman}

Being defined as a simple cost ratio -- even one that involves
computing optimal plan lengths -- $\ASratio$ cannot be fool-proof. We
mentioned in Section~\ref{real} already that one can replace $G$ with
a single goal $g$, and an additional action with precondition $G$ and
add effect $\{g\}$, thereby making $\ASratio$ devoid of information.
Such ``tricks'' could be dealt with by defining $\ASratio$ over
``necessary sub-goals'' instead, as explained in the next
section. More importantly perhaps, of course there are examples of
(parameterized) planning tasks where $\ASratio$ does {\em not}
correlate with DPLL proof size. One way to construct such examples is
by ``hiding'' a relevant phenomenon behind an irrelevant phenomenon
that controls $\ASratio$. Figure~\ref{XFIG/strawman.eps} provides the
construction of such a case, based on the MAP domain.

\widthfig{11}{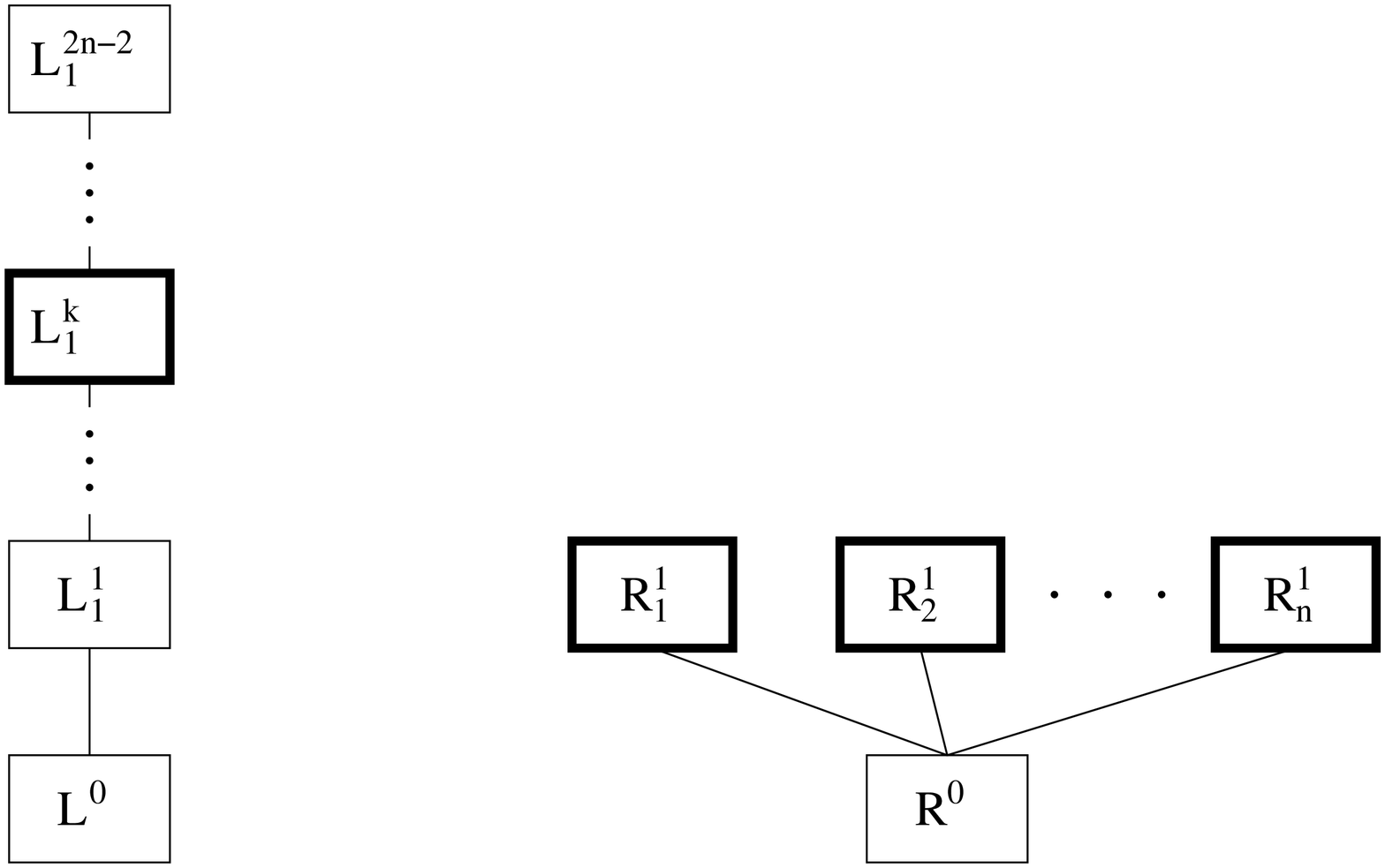}{The red herring example, consisting
  of two parallel maps on which nodes must be visited. Parallel moves
  on separate maps are allowed, thus plan length is $2n-1$
  irrespectively of $k$. Plan length bound for the CNFs is $2n-2$, the
  left map is solvable within $2n-2$ steps, the right map is not.
  $\ASratio$ is controlled by the left map, but to prove
  unsatisfiability one has to deal with the right map, which does not
  change over $k$ and demands exponentially long resolution
  proofs.}{-0.0cm}

The actions in Figure~\ref{XFIG/strawman.eps} are the same as in MAP,
moving along edges in the graph, precondition $\{at\mbox{-}x\}$, add
effect $\{at\mbox{-}y, visited\mbox{-}y\}$, delete effect
$\{at\mbox{-}x\}$.  The difference is that we now have {\em two} maps
(graphs), and we assume that we can move on both in parallel, i.e.,
our CNFs allow parallel move actions on separate graphs (which, for
example, the Graphplan-based encoding indeed does). Initially one is
located at $L^0$ and $R^0$. The goal is to visit the locations shown
in bold face in Figure~\ref{XFIG/strawman.eps}: $L_1^k$ and all of
$R_1^1, \dots, R_n^1$, where $k$ ranges between $1$ and
$2n-2$. Independently of $k$, the optimal plan length is $2n-1$, and
our CNF encodes $2n-2$ plan steps.  The left map is solvable within
this number of steps, so unsatisfiability must be proved on the right
map, which does not change over $k$ and requires exponentially long
resolution proofs, c.f.\ Theorem~\ref{the:map-base-lb}. $\ASratio$,
however, is $k/(2n-1)$.

In the red herring example, $\ASratio$ scales as the ratio between $k$
and $n$, as before, but the best-case DPLL proof size remains
constant. One can make the situation even ``worse'' by making the
right map more complicated. Say we introduce additional locations
$R_1^2, \dots, R_n^2$, each $R_i^2$ being linked to $R_i^1$. We
further introduce the location $R_1^3$ linked to $R_1^2$. The idea is
to introduce a varying number of goal nodes that lie ``further
out''. The larger the number of such further out goal nodes, the
easier we expect it to be (from our experience with the MAP domain) to
prove the right map unsatisfiable.  We can always ``balance'' the goal
nodes in a way keeping the optimal plan length constant at $2n-1$. We
then {\em invert} the correlation between $\ASratio$ and DPLL
performance by introducing a lot of further out goals when $k$ is
small, and less further out goals when $k$ is large. Concretely, with
maximum $k$ we take the goal to be the same as before, to visit
$R_1^1, \dots, R_n^1$. With minimum $k$, for simplicity say $n =
2m$. Then our goal will be to visit $R_1^3, R_2^2, \dots,
R_{m}^2$. The optimal plan here first visits all of $R_2^2, \dots,
R_{m}^2$, taking $4*(m-1) = 4*(n/2-1) = 2n-4$ steps ($4$ steps each
since one needs to go forth and back). Then $R_1^3$ is visited, taking
another $3$ steps.  All in all, with this construction, $\ASratio$ is
still dominated by the left map. But the right map becomes {\em
harder} to prove unsatisfiable as $k$ increases.

The example contains {\em completely separate} sub-problems, the left
and right maps. The complete separation invalidates the connection
between $\ASratio$ and our intuitions about the relevant ``problem
structure''. As explained in the introduction, the general intuition
is that (1) in the symmetrical case (low $\ASratio$) there is a fierce
competition of many sub-problems (goals) for the available resources,
and (2) in the asymmetrical case (high $\ASratio$) a single
sub-problem uses most of the resources on its own. In the red herring
example, (1) is still valid, but (2) is not since the asymmetric
sub-problem in question has access to its own resources -- the
parallel moves on the left map. So, in this case, $\ASratio$ fails to
characterize the intuitive ``degree of sub-problem interactions''. The
mistake lies in the maximization over individual goal fact costs,
which disregards to what extent the respective sub-problems are
actually interconnected. In this sense, the red herring is a
suggestion to research more direct measures of sub-problem
interactions, identifying their causes rather than their effects.


%% file: conclusion.tex
\section{Conclusions and Further Directions} 
\label{conclusion}

We defined a concrete notion of what problem structure is in Planning,
and we revealed empirically that this structure often governs SAT
solver performance. Our analytical results provide a detailed case
study of how this phenomenon arises. In particular, we show that the
phenomenon can make a doubly exponential difference, and we identify
some prototypical behaviors that appear also in planning competition
benchmarks.

Our parameterized CNFs can be used as a set of benchmarks with a very
``controlled'' nature. This is interesting since it allows very clean
and detailed tests of how good a SAT solver is at exploiting this
particular kind of problem structure. For example, our experiments
from Section~\ref{empirical} suggest that \zchaff\ and \minisat\ have
difficulties with ``slightly asymmetric'' instances.

There are some open topics to do with the definition of
$\ASratio$. First, one should try to define a version of $\ASratio$
that is more stable with respect to the ``sub-goal structure''. As
mentioned, our current definition, $\ASratio = max_{g \in G}
\cost(g)/\cost(\bigwedge_{g \in G} g)$, can be fooled by replacing $G$
with a single goal $g$, and introducing an additional action with
precondition $G$ and add effect $\{g\}$. A more stable approach would
be to identify a hierarchy of layers of ``necessary sub-goals'' --
facts that must be achieved along any solution path. Such facts were
termed ``landmarks'' in recent research \cite{hoffmann:etal:jair-04}.
In a nutshell, one could proceed as follows. Build a sequence of
landmark ``layers'' $G_0, \dots, G_m$.  Start with $G_0 := G$.
Iteratively, set $G_{i+1} := \bigcup_{g \in G_i} \bigcap_{a: g \in
add(a)} pre(a)$, until $G_{i+1}$ is contained in the previous
layers. Set $m := i$, i.e., consider the layers up to the last one
that brought a new fact.  It has been shown that such a process
results in informative problem decompositions in many benchmark
domains \cite{hoffmann:etal:jair-04}. For an alternative definition of
$\ASratio$, the idea would now be to select one layer $G_i$, and
define $\ASratio$ based on that, for example as $max_{g \in G_i}
\cost(g)+i$ divided by $\cost(\bigwedge_{g \in G} g)$. A good
heuristic may be to select the layer $G_i$ that contains the largest
number of facts. This approach could not be fooled by replacing $G$
with a single goal. It remains an open question in what kinds of
domains the approach would deliver better (or worse) information than
our current definition of $\ASratio$. We remark that the approach does
not solve the red herring example.  There, $G_1 =
\{at\mbox{-}L_1^{k-1}, at\mbox{-}R^0\}$, and $G_i =
\{at\mbox{-}L_1^{k-i}\}$ for $2 \leq i \leq k$. So, for all $i$,
$max_{g \in G_i} \cost(g)+i$ is the same as $max_{g \in G}
\cost(g)$. Motivated by this, one could try to come up with more
complex definitions of $\ASratio$, taking into account to what extent
the sub-problems corresponding to the individual goals are
interconnected.  Interconnection could, e.g., be approximated using
shared landmarks. A hard sub-problem that is tightly interconnected
would have more weight than one that is not.

A potentially more important topic is to explore whether it is
possible to define more direct measures of ``the degree of sub-problem
interactions''. A first option would be to explore this in the context
of the above landmarks.  Previous work \cite{hoffmann:etal:jair-04}
has already defined what ``interactions'' between landmarks could be
taken to mean. A landmark $L$ is said to interact with a landmark $L'$
if it is known that $L$ has to be true at some point after $L'$, {\em
and} that achieving $L'$ involves deleting $L$.  This means that, if
$L$ is achieved before $L'$, then $L$ must be re-achieved after
achieving $L'$ -- hence, $L$ should be achieved {\em before} $L'$. One
can thus define an order over the landmarks, based on their
interactions.  It is unclear, however, how this sort of interactions
could be turned into a number that stably correlates with performance
across a range of domains.

Alternative approaches to design more direct measures of sub-problem
interaction could be based on (1) the ``fact generation trees''
examined previously to draw conclusions about the quality of certain
heuristic functions \cite{hoffmann:jair-05}, or (2) the
``criticality'' measures proposed previously to design problem
abstractions \cite{bundy:etal:ai-96}. As for (1), the absence of a
certain kind of interactions in a fact generation tree allows us to
conclude that a certain heuristic function returns the precise goal
distance (as one would expect, this is not the case in any but the
most primitive examples). Like for the landmarks, it is unclear how
this sort of observation could be turned into a number estimating the
``degree'' of interaction. As for (2), this appears promising since,
in contrast to the previous two ideas, ``criticalities'' already are
numbers -- estimating how ``critical'' a fact is for a given set of
actions. The basic idea is to estimate how many alternative ways there
are to achieve a fact, and, recursively, how critical the
prerequisites of these alternatives are. The challenge is that, in
particular, the proposed computation of criticalities disregards
initial state, goal, and delete effects -- which is fine in the
original context, but not in our context here. All three (initial
state, goal, and delete effects) must be integrated into the
computation in order to obtain a measure of ``sub-problem
interaction'' in our sense. It is also unclear how the resulting
measure for each individual fact would be turned into a measure of
interactions between (several) facts.

We emphasize that our synthetic domains are relevant no matter how one
chooses to characterize ``the degree of sub-problem
interactions''. The intuition in the domains is that, on one side, we
have a CNF whose unsatisfiability arises from interactions between
many sub-problems, while on the other side we have a CNF whose
unsatisfiability arises mostly from the high degree of constrainedness
of a single sub-problem. As long as it is this sort of structure one
wants to capture, the domains are representative examples. In fact,
they can be used to test quickly whether or not a candidate definition
is sensible -- if the definition does not capture the transition on
our $k$ scales, then it can probably be discarded.

In some ways, our empirical and analytical observations could be used
to tailor SAT solvers for planning. A few insights obtained from our
analysis of backdoors are these. First, one should branch on actions
serving, recursively over add effects and preconditions, to support
the goals; such actions can be found by simple approximative backward
chaining.  Second, the number of branching decisions related to the
individual goal facts should have a similar distribution as the cost
of these facts. Third, the branching should be distributed evenly
across the time steps, with not much branching in directly neighbored
time steps (since decisions at $t$ typically affect $t-1$ and $t+1$ a
lot anyway). To the best of our knowledge, such techniques have hardly
been scratched so far in the (few) existing planning-specific
backtracking solvers
\cite{rintanen:kr-98,baioletti:etal:aips-00,hoffmann:geffner:icaps-03}.
The only similar technique we know of is used in ``BBG''
\cite{hoffmann:geffner:icaps-03}; this system branches on actions
occurring in a ``relaxed plan'', which is similar to the requirement
to support the goals.

A perhaps more original topic is to approximate $\ASratio$, and take
decisions based on that. Due to the wide-spread use of heuristic
functions in Planning, the literature contains a wealth of
well-researched techniques for approximating the number of steps
needed to achieve a goal, e.g.\
\cite{blum:furst:ai-97,bonet:geffner:ai-01,hoffmann:nebel:jair-01,edelkamp:ecp-01,gerevini:etal:jair-03,helmert:icaps-04,haslum:etal:aaai-05}.
We expect that (a combination of) these techniques could, with some
experimentation, be used to design a useful approximation of
$\ASratio$.

There are various possible uses of approximated $\ASratio$. First, one
could design specialized branching heuristics, and choose one
depending on the (approximated) value of $\ASratio$. The very
different forms of our identified backdoors in the symmetrical and
asymmetrical cases offer a rich source of ideas for such
heuristics. For example, it seems that a high $\ASratio$ allows for
large ``holes'' in the time steps branched upon, while a low
$\ASratio$ asks for more close decisions. Also, it seems that NOOP
variables are not important with high $\ASratio$, but do play a
crucial role with low $\ASratio$; this makes intuitive sense since, in
the presence of subtle interactions, it may be more important to
preserve the truth value of some fact through time steps. Perhaps most
importantly, with some more work approximated $\ASratio$ could
probably be made part of a successful runtime predictor. If not in
general -- across domains -- it is reasonable to expect that this will
work within some fixed domain (or class of domains) of interest.

Last but not least, we would like to point out that our work is merely
a first step towards understanding practical problem hardness in
optimal planning. Some important limitations of the work, in its
current state, are:
\begin{enumerate}[$\bullet$]
\item The experiments reported in Section~\ref{real} are performed
only for a narrow range of instance size parameters, for each
domain. It is important to determine how the behavior of $\ASratio$
changes as a function of changing parameter settings. We have merely
scratched this direction for Driverlog and Logistics; future work
should address it in full detail for some selected domains. The
problem in performing such a research lies in the large numbers of
domain parameters, and the exorbitant amount of computation time
needed to explore even a single parameter setting -- in our
experiments, often a single case required several months of
computation time.
\item Step-optimality, which we explore herein, is often not the most
relevant optimization criterion. In practice, one typically wants to
minimize makespan -- the end time point of a concurrent durational
plan -- or the summed up cost of all performed actions.\footnote{Note
that this is a problem of present-day optimal AI planning in general,
not only of our research. Planning research has only quite recently
started to develop techniques addressing makespan and, in particular,
cost optimization.}  While there are obvious extensions of $\ASratio$
to these settings, it is unclear if the phenomena observed herein will
carry over.
\item The analyzed synthetic domains are extremely simplistic; while
they provide clear intuitions, their relevance for complex practical
scenarios -- in particular for scenarios going beyond the planning
competition benchmarks -- is unclear.
\end{enumerate}
To sum up, we are far from claiming that our research ``solves'' the
addressed problems in a coherent way. Rather, we hope that our work
will inspire other researchers to work on similar topics. Results on
practical problem structure are difficult to obtain, but they are of
vital importance for understanding combinatorial search and its
behavior in practice. We believe that this kind of research should be
given more attention.



%% file: proofs.tex
\section{Proofs}
\label{proofs}

We first prove the resolution lower bound for MAP, then we prove the
theorems regarding existence of backdoors of a certain size, and of
their minimality.

\input{lbproofs.tex}

\bigskip

We now consider the backdoors of MAP, SBW, and SPH in turn.  The
reader is advised to read the proof to Theorem~\ref{mapbottom} first.
It explains the various arguments used in most detail, and some of the
general argumentation chains, notations, and terminology are re-used
in the subsequent proofs.

\input{MAP-proofs.tex}

\input{SBW-proofs.tex}

\input{SPH-proofs.tex}

%% file: lbproofs.tex
\subsection{MAP Resolution Lower Bound}
\label{lbproofs}

\begin{sloppypar}
By $PHP_n$, we denote the standard pigeon hole problem formula with
$n$ pigeons, i.e., the formula $SPH_n^1$. By $fPHP_n$, we denote the
{\em functional} pigeon hole problem with $n$ pigeons, i.e., the
formula $PHP_n \wedge \bigwedge_{1 \leq x \leq n, 1 \leq y \neq y'
  \leq n-1} \{\neg x\_y, \neg x\_y'\}$: the pigeon hole formula plus
clauses requiring that each pigeon is assigned to at most one hole.
(Throughout the section, we will denote pigeons as ``$x$'' and holes
as ``$y$''.) By $oPHP_n$, we denote the {\em onto} pigeon hole problem
with $n$ pigeons, i.e., the formula $PHP_n \wedge \bigwedge_{1 \leq y
  \leq n-1} \{1\_y, \dots, n\_y\}$: the pigeon hole formula plus
clauses requiring that at least one pigeon is assigned to each hole.
By $ofPHP_n$ we denote the combination of the two, i.e., the formula
including all the above clauses. It was recently proved that every
resolution proof of $ofPHP_n$ must have size $exp(\Omega(\frac{n}{(log
  (n+1))^2}))$ \cite{razborov:jcss:04}.
\end{sloppypar}

Our planning CNFs can be reduced to formulas close to the following
formula, that we call the {\em temporal} PHP, short $TPHP_n$.
\begin{enumerate}
\item The usual exclusion clauses: $\{\neg x\_y, \neg x'\_y\}$ for all
$1 \leq x \neq x' \leq n$ and $1 \leq y \leq n-1$.
\item A temporal version of the $\{x\_1, \dots, x\_(n-1)\}$ clauses: for
each $1 \leq x \leq n$,
\begin{enumerate}
\item $\{x\_1, \neg px\_2\}$
\item For each $2 \leq y \leq n-2$: $\{x\_y, px\_y, \neg px\_(y+1)\}$
\item $\{x\_(n-1), px\_(n-1)\}$
\end{enumerate}
\end{enumerate}
Here, the $px\_y$ are new variables whose intended meaning is that
$px\_y$ is set to $1$ iff $x$ is assigned to some hole $y' < y$. These
variables correspond to the NOOP actions in our encodings. By
$oTPHP_n$, we denote the {\em onto} version of this problem, i.e., the
$TPHP_n$ formula plus the clauses $\bigwedge_{1 \leq y \leq n-1}
\{1\_y, \dots, n\_y\}$.

Note that, by resolving over the $px\_y$ variables, $TPHP_n$ can be
reduced to $PHP_n$ in only $n(n-2)$ resolution steps. So any
resolution proof for $PHP_n$ can easily be turned into a proof for
$TPHP_n$. However, to prove lower bounds for $TPHP_n$, we need the
reduction the other way around, i.e., turn a resolution proof for
$TPHP_n$ into a proof for $PHP_n$.  We first show that $fPHP_n$ can be
polynomially reduced to $TPHP_n$.

\begin{lemma}\label{lem:fPHP2TPHP}
Let $P$ be a resolution proof for $TPHP_n$. Then a resolution proof
for $fPHP_n$ can be constructed from $P$ by replacing each resolution
step in $P$ with at most $n^2+n$ new resolution steps.
\end{lemma}

\begin{proof}
We define a translation function $t$ on literals as follows. The
literals $x\_y$ and $\neg x\_y$ remain untouched. For $px\_y$, we set
$t(px\_y)$ to $\{x\_1, \dots, x\_(y-1)\}$; we set $t(\neg px\_y)$ to
$\{x\_y, \dots, x\_(n-1)\}$. We extend $t$ to clauses by applying it
to the individual literals in the clause and taking the union of the
result. We will show that, for any clause $C$ that occurs in $P$, we
can derive the clause $t(C)$ by resolution steps in $fPHP$. Since
$t(\emptyset) = \emptyset$, this proves that we can turn $P$ into a
proof for $fPHP$. The claim follows from {\em how} we define the
resolution steps in $fPHP$. The proof is by induction over the
resolution steps in $P$.

For the base case in the induction, observe that $t(C) \in fPHP$ for
every $C \in TPHP$. This is obviously true for the exclusion clauses.
As for the other clauses, for each $x$ every one of them is mapped
onto the clause $\{x\_1, \dots, x\_(n-1)\}$.

\medskip

Now, say $\frac{C_1 \; \; C_2}{C_3}$ is a resolution step in $P$, so that we
already derived the clauses $t(C_1)$ and $t(C_2)$. We distinguish two
cases.

\medskip

{\bf Case 1, $\frac{C_1 \; \; C_2}{C_3}$ resolves over a variable
$x\_y$.} Say $x\_y \in C_1$ and $\neg x\_y \in C_2$. But then,
obviously, $x\_y \in t(C_1)$ and $\neg x\_y \in t(C_2)$. So
$\frac{t(C_1) \; \; t(C_2)}{C}$ is a legal resolution step, where $C =
t(C_3)$.

\medskip

{\bf Case 2, $\frac{C_1 \; \; C_2}{C_3}$ resolves over a variable
$px\_y$.} Say $px\_y \in C_1$ and $\neg px\_y \in C_2$. We have $C_1 =
A \cup \{px\_y\}$, $C_2 = B \cup \{\neg px\_y\}$, and $C_3 = A \cup
B$. We have $t(C_1) = t(A) \cup \{x\_1, \dots, x\_{y-1}\}$, and
$t(C_2) = t(B) \cup \{x\_y, \dots, x\_(n-1)\}$. We now derive the
clauses $tC_1^j = t(A) \cup \{\neg x\_j\}$, for $y \leq j \leq
n-1$. These can then be iteratively resolved with $t(C_2)$ to obtain
the clause $t(A) \cup t(B) = t(C_3)$. Each $tC_1^j$ is derived as
follows. We first resolve $t(C_1)$ with $\{\neg x\_1, \neg x\_j\}$,
yielding the clause $t(A) \cup \{\neg x\_j, x\_2, \dots,
x\_(y-1)\}$. We then resolve this with $\{\neg x\_2, \neg x\_j\}$ to
get to the clause $A^* \cup \{\neg x\_j, x\_3, \dots, x\_(y-1)\}$, and
so on until we get rid of $x\_(y-1)$. Obviously, the number of
resolution steps it takes to derive $t(C_3)$ is bounded by
$n^2+n$. This concludes the argument.
\end{proof}

We can immediately conclude that a similar statement holds for
$ofPHP_n$ and $oTPHP_n$.

\begin{cor}\label{cor:ofPHP2oTPHP}
Let $P$ be a resolution proof for $oTPHP_n$. Then a resolution proof
for $ofPHP_n$ can be constructed from $P$ by replacing each resolution
step in $P$ with at most $n^2+n$ new resolution steps.
\end{cor}

\begin{proof}
  Leaving the translation function $t$ from the proof to
  Lemma~\ref{lem:fPHP2TPHP} as it is, the only observation we need to
  make is that the additional clauses are not affected by $t$, i.e.,
  $t(\{1\_y, \dots, n\_y\}) = \{1\_y, \dots, n\_y\}$ for all $y$. The
  other arguments of the proof to Lemma~\ref{lem:fPHP2TPHP} remain
  valid.
\end{proof}

We next make a basic observation about reductions between CNF formulas
that are harmless in the sense that they preserve the existence and
size of resolution proofs. Let $\phi$ be a CNF formula with variable
set $V$. Then a {\em reduction function} $r$ maps $V$ into $V \cup
\overline{V} \cup \{0,1\}$, where $\overline{V} := \{\neg v \mid v \in
V\}$. That is, $r$ is a composition of assignments that either set a
variable $v \in V$ to $0$ or $1$, or that replace a variable $v \in V$
with (the negation of) a potentially different variable $v' \in V$. If
$v$ is not affected by $r$ then $r(v) = v$. We require that, for each
$v \in V$, if $r(v) \neq v$ then there is no $v'$ so that $r(v') = v$
or $r(v') = \neg v$. This forbids transitive replacements. We extend
$r$ to clauses by applying it to the individual variables, computing
negation (of negation) if required, and taking the union of the
result. We extend $r$ to CNF formulas by applying it to the individual
clauses and taking the union of the result. Note that, for sets $A$
and $B$ of literals, $r(A \cup B) = r(A) \cup r(B)$. For a formula
$\phi$ and reduction function $r$, by $r^*(\phi)$ we denote the
formula that results from $r(\phi)$ by removing literals assigned to
$0$, removing clauses that contain a literal assigned to $1$, and
removing clauses that contain both a literal and its negation. Given
formulas $\phi$ and $\psi$, we say that $\phi$ is {\em subsumed} by
$\psi$ if, for every $C \in \phi$, there exists $C^* \in \psi$ such
that $C \supseteq C^*$.

\begin{lemma}\label{lem:P2rP}
  Let $\phi$ and $\psi$ be unsatisfiable CNF formulas for which there
  exists a reduction function $r$ so that $r^*(\phi)$ is subsumed by
  $\psi$.  Let $P$ be a resolution proof for $\phi$. Then a resolution
  proof for $\psi$ can be constructed from $P$ by replacing each
  resolution step in $P$ with at most one new resolution step.
\end{lemma}

\begin{proof}
We will show that, for any clause $C$ that occurs in $P$, we either
have $1 \in r(C)$, or we have some $w$ with $\{w, \neg w\} \subseteq
r(C)$, or we can derive a clause $C^* \subseteq r(C)$ by resolution
steps in $\psi$. Since $r(\emptyset) = \emptyset$, this proves that we
can turn $P$ into a proof for $\psi$. The claim follows from {\em how}
we define the resolution steps in $\psi$. The proof is by induction
over the resolution steps in $P$.

For the base case of the induction, observe that the clauses $C \in
\phi$ satisfy the requirements by prerequisite. Since $r^*(\phi)$ is
subsumed by $\psi$, $r(C)$ either contains a true literal, or a
literal and its negation, or is a superset of some $C^* \in \psi$.

\medskip

Now, say $\frac{C_1 \; \; C_2}{C_3}$ is a resolution step in $P$,
resolving over a variable $v$. We denote $C_1 = A \cup \{v\}$ and $C_2
= B \cup \{\neg v\}$. We have $C_3 = A \cup B$, and we prove that
either $1 \in r(C_3)$, or there is some $w$ with $\{w, \neg w\}
\subseteq r(C_3)$, or we already derived a clause $C_3^* \subseteq
r(C_3)$, or we can derive a clause $C_3^* \subseteq r(C_3)$ with a
single resolution step. We have that $r(C_1) = r(A) \cup \{r(v)\}$,
that $r(C_2) = r(B) \cup \{\neg r(v)\}$, and that $r(C_3) = r(A) \cup
r(B)$. By induction hypothesis, we know, for each $i \in \{1, 2\}$,
that either $1 \in r(C_i)$, or there is some $w$ with $\{w, \neg w\}
\subseteq r(C_i)$, or we have derived a clause $C_i^* \subseteq
r(C_i)$. We distinguish three cases.

\medskip

{\bf Case 1, $1 \in r(C_1) \cup r(C_2)$.} Note that $r(C_1) \cup
r(C_2) = r(A) \cup \{r(v)\} \cup$ $r(B) \cup \{\neg r(v)\}$. First, if
$1 \in r(A) \cup r(B) = r(A \cup B) = r(C_3)$ then we are
done. Second, say $1 \not \in r(A) \cup r(B)$, and $r(v) = 1$. Then
$\neg r(v) = 0$, so $1 \not \in r(C_2)$, so by induction hypothesis we
have derived a clause $C_2^* \subseteq r(C_2)$. But, since $\{\neg
r(v)\} = \emptyset$, this means that $C_2^* \subseteq r(B)$, and we
are done with $r(B) \subseteq r(A \cup B)$.  Third, $1 \not \in r(A)
\cup r(B)$, and $r(v) = 0$. Then, similarly, we get that a clause
$C_1^* \subseteq r(A)$ has been derived.

\medskip

{\bf Case 2, there is a $w$ so that $\{w, \neg w\} \subseteq r(C_1)$
or $\{w, \neg w\} \subseteq r(C_2)$.}  First, observe that we are done
if $w \neq r(v)$. Say $\{w, \neg w\} \subseteq r(C_1)$, $w \neq
r(v)$. Then $r(C_1)$ has the form $r(A) \cup \{r(v)\}$ where $\{w,
\neg w\} \subseteq r(A)$, and we are done with $r(A) \subseteq
r(C_3)$. The symmetrical argument applies if $\{w, \neg w\} \subseteq
r(C_2)$, $w \neq r(v)$ (i.e., we get $\{w, \neg w\} \subseteq r(B)
\subseteq r(C_3)$. Say $w=r(v)$. If $\{r(v), \neg r(v)\}$ is contained
in both of $r(C_1)$ and $r(C_2)$, then $\neg r(v) \in r(A)$, and $r(v)
\in r(B)$, so $\{v, \neg v\} \subseteq r(A) \cup r(B)$, and we are
done. Say $w=r(v)$, and $\{r(v), \neg r(v)\}$ is contained only in
$r(C_1)$, not in $r(C_2)$. Then $\neg v \in r(A)$ and thus $\neg v \in
r(A) \cup r(B)$. But then, because $r(C_2) = r(B) \cup \{\neg v\}$,
$r(C_2) \subseteq r(A) \cup r(B)$, and we are done. If, finally,
$w=r(v)$ and $\{r(v), \neg r(v)\}$ is contained only in $r(C_2)$, then
$v \in B$ and, similarly, we are done with $r(C_1) \subseteq r(A) \cup
r(B)$.

\medskip

{\bf Case 3, $1 \not \in r(C_1) \cup r(C_2)$, and there is no $w$ so
  that $\{w, \neg w\} \subseteq C_1$ or $\{w, \neg w\} \subseteq
  C_2$.}  Then we know by induction hypothesis that clauses $C_1^*
\subseteq r(C_1)$ and $C_2^* \subseteq r(C_2)$ have been derived. If
$v \in C_1^*$ and $\neg v \in C_2^*$, then $\frac{C_1^* \; \;
  C_2^*}{C_3^*}$ is a legal resolution step, and we obviously have
$C_3^* \subseteq r(C_3)$. Say $v \not \in C_1^*$ or $\neg v \not \in
C_2^*$. We make a case distinction over the possible values of $r(v)$.
Note that $r(v)$ can not be $0$ or $1$ because that would be a
contradiction to $1 \not \in r(C_1) \cup r(C_2) = r(A) \cup \{r(v)\}
\cup$ $r(B) \cup \{\neg r(v)\}$. Now, first, say $r(v) = v$. Then, if
$v \not \in C_1^*$, we have $C_1^* \subseteq r(A)$ (remember that
$r(C_1) = r(A) \cup \{r(v)\}$) which suffices; if $\neg v \not \in
C_2^*$, we have $C_2^* \subseteq r(B)$ which suffices. Since one of
the two must be true this case is treated. Second, say $r(v) = v' \neq
v$. This case is treated with exactly the same argument (except
replacing $v$ with $v'$): if $v' \in C_1^*$ and $\neg v' \in C_2^*$,
then $\frac{C_1^* \; \; C_2^*}{C_3^*}$ is a legal resolution step;
else, either $C_1^* \subseteq r(A)$ or $C_2^* \subseteq r(B)$ follows.
Third, say $r(v) = \neg v'$. If $\neg v' \in C_1^*$ and $v' \in
C_2^*$, then $\frac{C_1^* \; \; C_2^*}{C_3^*}$ is a legal resolution
step, and we obviously have $C_3^* \subseteq r(C_3)$. Else, either
$\neg v' \not \in C_1^*$, or $v' \not \in C_2^*$ (or both). In the
former case, $C_1^* \subseteq r(A)$ follows and we are done. In the
latter case, $C_2^* \subseteq r(B)$ follows and we are done. This
concludes the argument.
\end{proof}

\bigskip

\noindent {\bf Theorem~\ref{the:map-base-lb}.} {\em Every resolution
  proof of $MAP_n^1$ must have size exponential in $n$.}

\begin{proof}
  We define a reduction function $r$ that reduces $MAP_n^1$ to a
  formula subsumed by $oTPHP_n$, i.e., so that $r^*(MAP_n^1)$ is
  subsumed by $oTPHP_n$. This suffices with Lemma~\ref{lem:P2rP} and
  Corollary~\ref{cor:ofPHP2oTPHP} since every resolution proof of
  $ofPHP_n$ must have size $exp(\Omega(\frac{n}{(log (n+1))^2}))$
  \cite{razborov:jcss:04}. It is advisable to have a look at the
  definition of $MAP_n^1$, i.e. Section~\ref{sd:map} for the
  definition of the underlying planning task and Section~\ref{prelim}
  for the definition of the CNF encoding, before reading the proof
  below.

  For the purpose of the proof, by $mapTPHP_n$ we denote the formula
\[
TPHP \wedge \bigwedge_{1 \leq x \leq n, 1 \leq y \leq n-2} \{\neg
x\_(y+1), 1\_y, \dots, n\_y\},
\]
which will be the precise formula that we reduce $MAP_n^1$ to. That
is, we will get $r^*(MAP_n^1) = mapTPHP_n$. Obviously, $mapTPHP_n$ is
subsumed by $oTPHP_n$.

\begin{sloppypar}
  The variables in $MAP_n^1$ are the following:
  \begin{enumerate}
  \item $move\mbox{-}L^0\mbox{-}L_i^1(t)$ for $1 \leq i \leq n$ and $1
  \leq t \leq 2n-2$,
  \item $move\mbox{-}L_i^1\mbox{-}L^0(t)$ for $1 \leq i \leq n$ and $2
  \leq t \leq 2n-2$,
  \item $move\mbox{-}L_1^{j-1}\mbox{-}L_1^{j}(t)$ for $2 \leq j \leq
  2n-3$ and $j \leq t \leq 2n-2$,
  \item $move\mbox{-}L_1^{j}\mbox{-}L_1^{j-1}(t)$ for $2 \leq j \leq
  2n-3$ and $j+1 \leq t \leq 2n-2$,
  \item $NOOP\mbox{-}at\mbox{-}L^0(t)$ for $1 \leq t \leq 2n-2$,
  \item $NOOP\mbox{-}at\mbox{-}L_i^1(t)$ for $1 \leq i \leq n$ and $2
  \leq t \leq 2n-2$,
  \item $NOOP\mbox{-}at\mbox{-}L_1^j(t)$ for $2 \leq j \leq 2n-3$ and
  $j+1 \leq t \leq 2n-2$,
  \item $NOOP\mbox{-}visited\mbox{-}L^0(t)$ for $3 \leq t \leq 2n-2$,
  \item $NOOP\mbox{-}visited\mbox{-}L_i^1(t)$ for $1 \leq i \leq n$
  and $2 \leq t \leq 2n-2$,
  \item $NOOP\mbox{-}visited\mbox{-}L_1^j(t)$ for $2 \leq j \leq 2n-3$
  and $j+1 \leq t \leq 2n-2$.
  \end{enumerate}
  The other combinations of actions and time steps are not present.
  The idea behind the reduction function is to introduce constraints
  of that we know that they would be satisfied in any optimal plan for
  the underlying planning task. Note that this is just a rationale to
  make the following understandable. The formal property that we
  need/that suffices is that $r^*(MAP_n^1) = mapTPHP_n$, which we will
  show below. Our reduction function is defined as follows:
  \begin{enumerate}
  \item We set $r(move\mbox{-}L^0\mbox{-}L_i^1(t)) := 0$ for $1 \leq i
  \leq n$ and $t \in \{2, 4, \dots, 2n-2\}$. Any optimal plan will
  iterate between consecutive moves from $L^0$ to $L_1^i$ and back
  (except in the last step), starting at the first step. So we will
  never move from $L^0$ to $L_1^i$ in an even time step.
  \item
  \begin{enumerate}
  \item We set $r(move\mbox{-}L_i^1\mbox{-}L^0(t)) := 0$ for $1 \leq i
  \leq n$ and $t \in \{3, \dots, 2n-3\}$. Same rationale as above.
  \item We set $r(move\mbox{-}L_i^1\mbox{-}L^0(t)) :=$
  $move\mbox{-}L^0\mbox{-}L_i^1(t-1)$ for $1 \leq i \leq n$ and $t \in
  \{2, 4, \dots, 2n-2\}$. Same rationale as above.
  \end{enumerate}
  \item We set $r(move\mbox{-}L_1^{j-1}\mbox{-}L_1^{j}(t)) := 0$ for
  $2 \leq j \leq 2n-3$ and $j \leq t \leq 2n-2$. No optimal plan will
  move along these locations.
  \item We set $r(move\mbox{-}L_1^{j}\mbox{-}L_1^{j-1}(t)) := 0$ for
  $2 \leq j \leq 2n-3$ and $j+1 \leq t \leq 2n-2$. See above.
  \item We set $r(NOOP\mbox{-}at\mbox{-}L^0(t)) := 0$ for $1 \leq t
  \leq 2n-2$. We don't have the time to be standing still.
  \item We set $r(NOOP\mbox{-}at\mbox{-}L_i^1(t)) := 0$ for $1 \leq i
  \leq n$ and $2 \leq t \leq 2n-2$. See above.
  \item We set $r(NOOP\mbox{-}at\mbox{-}L_1^j(t)) := 0$ for $2 \leq j
  \leq 2n-3$ and $j+1 \leq t \leq 2n-2$. See above.
  \item We set $r(NOOP\mbox{-}visited\mbox{-}L^0(t)) := 0$ for $3 \leq
  t \leq 2n-2$. It does not matter whether or not we have already
  visited (moved to) $L^0$.
  \item
  \begin{enumerate}
  \item We set $r(NOOP\mbox{-}visited\mbox{-}L_i^1(2n-2)) := 1$ for $1
  \leq i \leq n$. Since the $move\mbox{-}L^0\mbox{-}L_i^1(2n-2)$
  variables are set to $0$, we will have to visit these locations
  earlier.
  \item We set $r(NOOP\mbox{-}visited\mbox{-}L_i^1(t)) :=$
  $NOOP\mbox{-}visited\mbox{-}L_i^1(t+1)$ for $1 \leq i \leq n$ and $t
  \in \{2, 4, \dots, 2n-4\}$. Since the
  $move\mbox{-}L^0\mbox{-}L_i^1(t)$ variables at these $t$ are set to
  $0$, having visited $L_i^1$ by time $t$ is the same as having
  visited it by time $t+1$.
  \end{enumerate}
  \item We set $r(NOOP\mbox{-}visited\mbox{-}L_1^j(t)) := 0$ for $2
  \leq j \leq 2n-3$ and $j+1 \leq t \leq 2n-2$. It does not matter
  whether or not we have already visited any of these locations.
  \end{enumerate}
  After doing these replacements, the only variables that remain in
  the formula, i.e., where $r(v) = v$, have the form
  $move\mbox{-}L^0\mbox{-}L_i^1(t)$ and
  $NOOP\mbox{-}visited\mbox{-}L_i^1(t)$, for $1 \leq t \leq n$ and $t
  \in \{1, 3, \dots, 2n-3\}$. The intuition is the following. Imagine
  to rename the former to $i\_t$, i.e. $x\_y$ where $x=i$ and $y=t$,
  and the latter to $pi\_t$, i.e. $px\_y$ where $x=i$ and $y=t$. The
  remaining formula is exactly $mapTPHP_n$, except that the range of
  $t$ is $\{1, \dots, 2n-3\}$ instead of $\{1, \dots,
  n-1\}$. Obviously, the necessary additional renaming is $(t+1)/2$
  for $t$. To see that we indeed get down to $mapTPHP_n$ in this way,
  let us consider what happens to the clauses in $MAP_n^1$.

  First, observe that the GC clauses,
  $\{move\mbox{-}L^0\mbox{-}L_i^1(2n-2),$
  $NOOP\mbox{-}visited\mbox{-}L_i^1(2n-2)\}$ (plus
  $move\mbox{-}L_1^2\mbox{-}L_i^1(2n-2)$, for $i=1$) are removed
  because $NOOP\mbox{-}visited\mbox{-}L_i^1(2n-2)$ is set to $1$.
\end{sloppypar}

Next, observe that the only remaining EC clauses, where none of the
two variables is set to $0$, have the form $\{\neg
move\mbox{-}L^0\mbox{-}L_i^1(t),$ $\neg
move\mbox{-}L^0\mbox{-}L_i'^1(t)\}$, for $1 \leq i \neq i' \leq n$ and
$t \in \{1, 3, \dots, 2n-3\}$. To see this, simply consider that
$NOOP\mbox{-}visited$ variables do not participate in any EC clauses,
that $NOOP\mbox{-}visited$ variables are the only ones set to $1$ by
$r$, and that the only non-$NOOP\mbox{-}visited$ variables not removed
(set to $0$ or to another variable) by $r$ are the
$move\mbox{-}L^0\mbox{-}L_i^1(t)$ variables, for $t \in \{1, 3, \dots,
2n-3\}$.

  So the GC clauses get removed from the formula, and the EC clauses
  boil down to precisely the exclusion clauses in $mapTPHP_n$. It
  remains to show that the simplified AC clauses are equal to the
  other clauses in $mapTPHP_n$. These latter clauses are, for each $1
  \leq i \leq n$:
\begin{enumerate}
\item $\{\neg pi\_3, i\_1\}$,
\item for each $t \in \{3, 5, \dots, 2n-5\}$: $\{\neg pi\_(t+2), i\_t,
pi\_t\}$,
\item $\{i\_(2n-3), pi\_(2n-3)\}$,
\item for each $t \in \{1, 3, \dots, 2n-5\}$: $\{\neg i\_(t+2), 1\_t,
\dots, n\_t\}$.
\end{enumerate}
The indexing with ``$i$'' and ``$t$'' instead of ``$x$'' and ``$y$''
is chosen to improve readability. Remember that the intended
correspondence is $i\_t$ for $move\mbox{-}L^0\mbox{-}L_i^1(t)$ and
$pi\_t$ for $NOOP\mbox{-}visited\mbox{-}L_i^1(t)$. (To avoid
confusion, here the range $\{1, \dots, n-1\}$ for $y$ is replaced by
the range $\{3, 5, \dots, 2n-3\}$ for $t$; we also did a little
re-ordering of literals to have an immediate match with the clauses
listed below.)

Remember that the AC clauses for an action $a$ at step $t$ have the
form $\{\neg a(t), a_1(t-1), \dots, a_l(t-1)\}$, where $a_1, \dots,
a_l$ are all achievers of a precondition $p$ of $a$ that are present
at time $t-1$. We have such a clause for every action $a$ present at a
time step $t>1$, and for every precondition of $a$. In our case,
i.e. in the MAP domain, each action has only a single precondition so
there is only a single AC clause. 

First, observe that many AC clauses are actually removed, because the
respective variables are set to $0$ by $r$:
\begin{itemize}
\item the AC clause of every $NOOP\mbox{-}at$ action,
\item the AC clause of every $move\mbox{-}L_1^j\mbox{-}L_1^{j+1}$
action for $j \geq 1$,
\item the AC clause of every $move\mbox{-}L_1^{j+1}\mbox{-}L_1^{j}$
action for $j \geq 1$,
\item the AC clause of every $move\mbox{-}L^0\mbox{-}L_i^1(t)$
variable for $t \in \{2, 4, \dots, 2n-2\}$,
\item the AC clause of every $move\mbox{-}L_i^1\mbox{-}L^0(t)$
variable for $t \in \{1, 3, \dots, 2n-3\}$.
\end{itemize}
\begin{sloppypar}
The AC clause of every $move\mbox{-}L_i^1\mbox{-}L^0(t)$ variable for
$t \in \{2, 4, \dots, 2n-2\}$ is removed because it has the form
$\{\neg move\mbox{-}L_i^1\mbox{-}L^0(t),$
$move\mbox{-}L^0\mbox{-}L_i^1(t-1),$
$NOOP\mbox{-}at\mbox{-}L_i^1(t-1)\}$. This is mapped by $r$ onto
$\{\neg move\mbox{-}L^0\mbox{-}L_i^1(t-1),$
$move\mbox{-}L^0\mbox{-}L_i^1(t-1), 0\}$ and removed by $r^*$ because
it contains both a literal and its negation.
\end{sloppypar}

What remains are the AC clauses of the following actions, for each $1
\leq i \leq n$ (compare the list of the remaining $mapTPHP_n$ clauses
above):
\begin{sloppypar}
\begin{enumerate}
\item $NOOP\mbox{-}visited\mbox{-}L_i^1(3)$, whose AC clause has the
form $\{\neg NOOP\mbox{-}visited\mbox{-}L_i^1(3),$
$move\mbox{-}L^0\mbox{-}L_i^1(1)\}$,
\item for each $t \in \{3, 5, \dots, 2n-5\}$:
$NOOP\mbox{-}visited\mbox{-}L_i^1(t+2)$, whose AC clause has the form
$\{\neg NOOP\mbox{-}visited\mbox{-}L_i^1(t+2),$
$move\mbox{-}L^0\mbox{-}L_i^1(1),$
$NOOP\mbox{-}visited\mbox{-}L_i^1(t)\}$,
\item $NOOP\mbox{-}visited\mbox{-}L_i^1(2n-2)$, whose AC clause has
the form $\{\neg NOOP\mbox{-}visited\mbox{-}L_i^1(2n-2),$
$move\mbox{-}L^0\mbox{-}L_i^1(2n-3),$
$NOOP\mbox{-}visited\mbox{-}L_i^1(2n-3)\}$, where
$NOOP\mbox{-}visited\mbox{-}L_i^1(2n-2)$ is set to $1$ by $r$,
\item for each $t \in \{1, 3, \dots, 2n-5\}$:
  $move\mbox{-}L^0\mbox{-}L_i^1(t+2)$, whose AC clause has the form
  $\{\neg move\mbox{-}L^0\mbox{-}L_i^1(t+2),$
  $NOOP\mbox{-}at\mbox{-}L^0(t+1),$
  $move\mbox{-}L_1^1\mbox{-}L^0(t+1), \dots, $
  $move\mbox{-}L_n^1\mbox{-}L^0(t+1)\}$, which is mapped by $r$ onto
  $\{\neg move\mbox{-}L^0\mbox{-}L_i^1(t+2),$ $0,$
  $move\mbox{-}L^0\mbox{-}L_1^1(t), \dots,$
  $move\mbox{-}L^0\mbox{-}L_n^1(t)\}$.
\end{enumerate}
\noindent
Altogether, the remaining clauses we get are exactly the same as those
in $mapTPHP_n$. This concludes the argument.
\end{sloppypar}
\end{proof}

%% file: MAP-proofs.tex
\subsection{MAP  Backdoors}
\label{proofs:map}

We first consider the symmetrical case, then the asymmetrical case.

\subsubsection{Symmetrical Case}
\label{proofs:map:bottom}

Recall that the clauses encoding action precondition support are
referred to as {\em action precondition (AC) clauses}, the clauses
encoding goal support are referred to as {\em goal (GC) clauses}, and
the clauses encoding action incompatibility are referred to as {\em
exclusion (EC) clauses}. Recall also that every pair of non-NOOP
actions is incompatible in our encodings -- in particular, there is an
EC clause, at any time step, for any two $move$ actions in MAP.

As stated in Section~\ref{sd:map}, the backdoor we identify for
$MAP_n^1$, denoted by $MAP_n^1B$, is defined as follows (compare
Figure~\ref{XFIG/map-base-newBD.eps}), where $T := \{3, 5, \dots, 2n -
3\}$: {\small
\medskip
\begin{tabbing}
$MAP_n^1B$ := \= $\{ move\mbox{-}L^0\mbox{-}L_i^1(t) \mid t \in T, 2
\leq i \leq n \} \; \; \cup$\\
\> $\{ NOOP\mbox{-}visited\mbox{-}L_i^1(t) \mid t \in T, 3
  \leq i \leq n \} \; \; \cup$\\
\> $\{ NOOP\mbox{-}at\mbox{-}L^0(1) \} \; \; \cup$\\
\> $\{ move\mbox{-}L^0\mbox{-}L_1^1(t) \mid t \in T \setminus \{2n-5, 2n-3\} \}$
\end{tabbing}}

\bigskip

\noindent {\bf Theorem~\ref{mapbottom}.} {\em $MAP_n^1B$ is a backdoor
  for $MAP_n^1$.}

\begin{proof}
Throughout the proof, to simplify the notation, we denote $\phi :=
MAP_n^1$. The main idea behind the proof is to look at any assignment
$a$ to the variables in $MAP_n^1B$ in a regression-style nature. In
the last time step of $\phi$, $t=2n-2$, there are the $n$ GC clauses,
the ``goal constraints'' requiring, for each branch $i$, to either
visit $L_i^1$ right here, or to include a NOOP indicating that $L_i^1$
has been visited earlier already. Now, if we assign values to all
$MAP_n^1B$ variables at time step $2n-3$, then we will find that at
least $n-1$ of the goal constraints have ``survived'', i.e., after
unit propagation we will have at least $n-1$ similar clauses in time
step $2n-4$. Iterating this argument, we end up with a non-empty set
of goal constraints at time step $2$. From there, the backdoor
property follows with a number of subtle observations and case
distinctions.

To formalize the ``regression steps'', we will need another notation.
For a formula $\psi$ (i.e., for a partially instantiated version of
$\phi$) we denote
\[
G^t(\psi) := \{ i \mid \{NOOP\mbox{-}visited\mbox{-}L_i^1(t),
  move\mbox{-}L^0\mbox{-}L_i^1(t)\} \in \psi \}.
\]
That is, $G^t(\psi)$ denotes the set of the indices of the branches
for which goal constraints are present in $\psi$ at time
$t$.

First, observe that $G^{2n-2}(\phi) = \{2, \dots, n\}$. This is just
due to the GC clauses contained in the original formula at the last
time step. Note that the GC clause for branch $1$ is different because
$L_1^1$ can also be reached from $L_1^2$. Precisely, the goal
constraint for branch $1$ is
$\{NOOP\mbox{-}visited\mbox{-}L_i^1(2n-2),$
$move\mbox{-}L^0\mbox{-}L_i^1(2n-2),$
$move\mbox{-}L_1^2\mbox{-}L_i^1(2n-2)\}$; we will dedicate a special
case treatment to it further below. That special case treatment will
also make use of the $|T|-2$ $move\mbox{-}L^0\mbox{-}L_1^1$
variables in $MAP_n^1B$; until then, we ignore (don't need to
consider) these variables.

From now on, we assume we are given any truth value assignment $a$ to
the variables in $MAP_n^1B$. Our task is to show that $UP(\phi_a)$
contains a contradiction, i.e., contains the empty clause. Since, as
stated, we will be looking at $a$ in regression-style, we denote with
$a^{\geq t}$ the restriction of $a$ to the subset of variables in
$MAP_n^1B$ at time steps $t' \geq t$.

Before we formalize the regression steps, it is instructive to observe
that, if $G^2(\psi) \geq 2$, then branching over
$NOOP\mbox{-}at\mbox{-}L^0(1)$ yields a contradiction. More
precisely:

\begin{sloppypar}
\begin{itemize}
\item {\bf Observation 1.} {\em Let $\psi$ result from $\phi$ by 
executing all value assignments in $a$ at time steps $t' \geq 3$, and
applying UP, i.e., $\psi = UP(\phi_{a^{\geq 3}})$. If $G^2(\psi)
\supseteq \{g, g'\}$ for some $g \neq g'$, then after assigning a
value to $NOOP\mbox{-}at\mbox{-}L^0(1)$ UP yields the empty clause.}

\medskip

Case 1, $a$ sets $NOOP\mbox{-}at\mbox{-}L^0(1)$ to $0$. Then UP sets
$move\mbox{-}L^0\mbox{-}L_g^1(2)$ and
$move\mbox{-}L^0\mbox{-}L_{g'}^1(2)$ to $0$ due to these variables'
AC clauses. In effect, UP over the two goal constraints sets
$NOOP\mbox{-}visited\mbox{-}L_g^1(2)$ and
$NOOP\mbox{-}visited\mbox{-}L_{g'}^1(2)$ to $1$. Over AC clauses, UP
now sets both $move\mbox{-}L^0\mbox{-}L_g^1(1)$ and
$move\mbox{-}L^0\mbox{-}L_{g'}^1(1)$ to $1$, resulting in an empty
EC clause.

\medskip

Case 2, $a$ sets $NOOP\mbox{-}at\mbox{-}L^0(1)$ to $1$. Then UP sets
$move\mbox{-}L^0\mbox{-}L_g^1(1)$ and
$move\mbox{-}L^0\mbox{-}L_{g'}^1(1)$ to $0$ [EC clauses], sets
$NOOP\mbox{-}visited\mbox{-}L_g^1(2)$ and
$NOOP\mbox{-}visited\mbox{-}L_{g'}^1(2)$ to $0$ [AC clauses], and sets
$move\mbox{-}L^0\mbox{-}L_g^1(2)$ and
$move\mbox{-}L^0\mbox{-}L_{g'}^1(2)$ to $1$ [goal constraints],
yielding again an empty EC clause.
\end{itemize}
\end{sloppypar}

There are initially $n-1$ goal constraints, $G^{2n-2}(\phi) = \{2,
\dots, n\}$. There are $n-2$ time steps in $T = \{3, 5, \dots, 2n-3\}$.
As we move over the time steps $t \in T$, as we will show we get rid
of at most one of these per time step. Precisely, at each step $t \in
T$ we are in the following situation. We have $G^{t+1} :=
G^{t+1}(UP(\phi_{a^{\geq t+2}}))$, and we want to show something about
$G^{t-1} := G^{t-1}(UP(\phi_{a^{\geq t}}))$. Note that, for $t=2n-3$,
i.e. for the highest step in $T$, $a^{\geq t+2} = a^{\geq 2n-1}$ is
the empty assignment, and $G^{t+1} =$ $G^{t+1}(UP(\phi_{a^{\geq
t+2}})) =$ $G^{2n+2}(\phi) =$ $\{2, \dots, n\}$. The possible
regression steps are:

\begin{sloppypar}
\begin{itemize}
\item {\bf Regression step A.} {\em There is an $x^t \in \{2, \dots, n\}$ so 
that $a$ sets $move\mbox{-}L^0\mbox{-}L_{x^t}^1(t)$ to $1$. Then
$G^{t-1} = G^{t+1} \setminus \{x^t\}$.}

\medskip

After setting $move\mbox{-}L^0\mbox{-}L_i^1(t)$ to $1$, UP does the
following assignments: $move\mbox{-}L^0\mbox{-}L_{i}^1(t) = 0$ for
$i \neq x^t$ [EC clauses; we get a contradiction if one of these
variables is is set to $1$ by $a$];
$move\mbox{-}L_i^1\mbox{-}L^0(t) = 0$ for $i \neq x_t$ [EC
clauses]; $NOOP\mbox{-}at\mbox{-}L^0(t) = 0$ [EC clause];
$move\mbox{-}L^0\mbox{-}L_i^1(t+1) = 0$ for all $i$ [AC clauses];
$NOOP\mbox{-}visited\mbox{-}L_g^1(t+1) = 1$ for $g \in G^{t+1}$ [goal
constraints]; $NOOP\mbox{-}visited\mbox{-}L_g^1(t) = 1$ for $g \in
G^{t+1} \setminus
\{ x^t \}$ [AC clause for $NOOP\mbox{-}visited\mbox{-}L_g^1(t)$; the 
respective move actions are set to $0$ already; if a
$NOOP\mbox{-}visited\mbox{-}L_g^1(t)$ is set to $0$ by $a$, we get a
contradiction]. The last step proves the claim: for each $g \in
G^{t+1} \setminus \{ x^t \}$ we get, from the AC clause to
$NOOP\mbox{-}visited\mbox{-}L_g^1(t)$, the clause
$\{NOOP\mbox{-}visited\mbox{-}L_g^1(t-1),
move\mbox{-}L^0\mbox{-}L_g^1(t-1)\}$.

\medskip

\item {\bf Regression step B.} {\em All $move\mbox{-}L^0\mbox{-}L_i^1(t)$ 
variables, $2 \leq i \leq n$, are set to $0$ by $a$, and there is an
$x^t \in \{3, \dots, n\}$ so that $a$ sets
$NOOP\mbox{-}visited\mbox{-}L_{x^t}^1(t)$ to $0$. Then
$move\mbox{-}L^0\mbox{-}L_{x^t}^1(t+1)$ is set to $1$ by UP, and
$G^{t-1} = G^{t+1} \setminus
\{x^t\}$.}

\medskip

After $NOOP\mbox{-}visited\mbox{-}L_{x^t}^1(t)$ is set to $0$, UP does
the following assignments: $NOOP\mbox{-}visited\mbox{-}L_{x^t}^1(t+1)
= 0$ [AC clause]; $move\mbox{-}L^0\mbox{-}L_{x^t}^1(t+1) = 1$ [goal
constraint for $x^t$]; $move\mbox{-}L^0\mbox{-}L_i^1(t+1) = 0$ for
$i \neq x^t$; $NOOP\mbox{-}visited\mbox{-}L_g^1(t+1) = 1$ for $g \in
G^{t+1} \setminus \{ x^t \}$ [goal constraints];
$NOOP\mbox{-}visited\mbox{-}L_g^1(t) = 1$ for $g \in G^{t+1} \setminus
\{ x^t \}$ [AC clause for $NOOP\mbox{-}visited\mbox{-}L_g^1(t)$; the 
respective move actions are assigned to $0$ by $a$; if a
$NOOP\mbox{-}visited\mbox{-}L_g^1(t)$ is set to $0$ by $a$, we get a
contradiction]. The claim now follows with the same argument as above.

\medskip

\item {\bf Regression step C.} {\em All $move\mbox{-}L^0\mbox{-}L_i^1(t)$ 
variables, $2 \leq i \leq n$, are set to $0$ by $a$, and all
$NOOP\mbox{-}visited\mbox{-}L_{i}^1(t)$ variables, $3 \leq i \leq n$,
are set to $1$ by $a$. Then $G^{t-1} = G^{t+1} \setminus
\{x^t\}$ where $x_t := 2$.}

\medskip

Having set all the $NOOP\mbox{-}visited\mbox{-}L_{i}^1(t)$ variables
to $1$, from the AC clauses to these variables for $i \in G^{t+1}$ we
immediately get the claim.
\end{itemize}
\end{sloppypar}

What the above immediately shows is that, as we move along $a$ from
the top to the bottom of the time steps, at each $t \in T$ we either
get an empty clause by UP, or there exists some $x^t \in \{2, \dots,
n\}$ so that $G^{t-1} = G^{t+1} \setminus \{x^t\}$. Since $|T| = n-2$,
this means that $|G^2| \geq 1$. Observation 1 above tells us that we
get a contradiction if $|G^2| \geq 2$. Clearly, $|G^2|$ is $1$ only if
all the $x^t$, $t \in T$, are pairwise different -- that is, if at
each regression step we get rid of a different goal constraint. (Also,
$a$ must assign $0$ to each variable
$NOOP\mbox{-}visited\mbox{-}L_{x^t}^1(t')$, for $t' < t$, in order to
not re-introduce the goal constraint for $x^t$.)

At this point, we must do a little forward-style reasoning. Let $t_C$
be the time point in $T$ at which $a$ ``uses'' regression step C (with
the above, there is exactly one such time point, or we get an empty
clause by UP). We will now show that, by UP,
$NOOP\mbox{-}visited\mbox{-}L_{2}^1(t_C)$, i.e. the NOOP variable not
in $MAP_n^1$ at that time, is set to $0$ by UP. This then implies that
the regression step at $t_C$ is exactly identical to regression step
B; in particular, a move variable ($move\mbox{-}L^0\mbox{-}L_{2}^1$)
is set to $1$ by UP at time $t_C+1$. We will need this for our final
case distinction below.

\begin{sloppypar}
We know that $|G^2| = 1$. Let's denote the element of $G^2$ with $x^1$
(the naming is chosen to be consistent with the $t$ indices of the
other $x^t$, increasing by $2$ between variables). If $a$ assigns
$NOOP\mbox{-}at\mbox{-}L^0(1)$ to $0$, then UP sets
$move\mbox{-}L^0\mbox{-}L_{x^1}^1(2)$ to $0$ [AC clause],
$NOOP\mbox{-}visited\mbox{-}L_{x^1}^1(2)$ to $1$ [goal constraint],
and $move\mbox{-}L^0\mbox{-}L_{x^1}^1(1)$ to $1$ [AC clause].  If
$a$ assigns $NOOP\mbox{-}at\mbox{-}L^0(1)$ to $1$, then UP sets
$move\mbox{-}L^0\mbox{-}L_{x^1}^1(1)$ to $0$ [EC clause],
$NOOP\mbox{-}visited\mbox{-}L_{x^1}^1(2)$ to $0$ [AC clause], and
$move\mbox{-}L^0\mbox{-}L_{x^1}^1(2)$ to $1$ [goal constraint]. So,
no matter what value is assigned to $NOOP\mbox{-}at\mbox{-}L^0(1)$,
after UP all move variables (except those for $x^1$) are forced out in
time steps $1$ and $2$; in particular,
$move\mbox{-}L^0\mbox{-}L_{2}^1$ is forced out.
\end{sloppypar}

Considering the next higher time step $t=3 \in T$, if $a$ uses
regression step A at $t$, then all move variables are forced out at
time $t$ by EC clauses, and all $move\mbox{-}L^0\mbox{-}L_{i}^1$
variables are forced out at time $t+1$ by $AC$ clauses (see the proof
to regression step A above). If $a$ uses regression step B at $t$, then
all $move\mbox{-}L^0\mbox{-}L_{i}^1$ variables, $2 \leq i \leq n$,
are set to $0$ at time $t$ by $a$, and all move variables are forced
out at time $t+1$ by EC clauses. We can iterate the argument upwards
over the $t \in T$, and find that, in particular,
$move\mbox{-}L^0\mbox{-}L_{2}^1$ is forced out at all time steps $t
< t_C$. By AC clause, obviously this causes
$NOOP\mbox{-}visited\mbox{-}L_{2}^1(t_C)$ to be set to $0$, which we
wanted to show.

We now know that, in each pair of time steps $\{1, 2\},$ $\dots,$
$\{2n-3, 2n-2\}$, exactly one move variable is set to $1$ after UP.
There are $n-1$ such pairs, and the goal constraints for each of the
$n-1$ branches $\{2, \dots, n\}$ have been accommodated. We now show
that UP yields a contradiction due to the remaining open goal
constraint for branch $1$, and the need to assign values to the
$|T|-2$ $move\mbox{-}L^0\mbox{-}L_1^1$ variables in $MAP_n^1$. For
consistent terminology, we say that regression step A is used at $t=1$
if $move\mbox{-}L^0\mbox{-}L_{x^1}^1$ is forced to $1$ at time $1$;
otherwise (i.e, if $move\mbox{-}L^0\mbox{-}L_{x^1}^1$ is forced to
$1$ at time $2$), we say that regression step B is used at $t=1$. We
will be able to prove the theorem once we made the following
observation:

\begin{sloppypar}
\begin{itemize}
\item {\bf Observation 2.} {\em There is a time point $t_0 \in 1 \cup
    T \cup \{2n-1\}$ so that, at all time steps $t < t_0$, regression
    step A is used, while, at all time steps $t \geq t_0$, regression
    step B is used. (Or else we get a contradiction by UP.)}

\medskip

The reason for this is, simply, that UP yields an empty clause if
regression step B is used at $t$, and regression step A is used at
$t+2$. Namely, using regression step $B$ at $t$ enforces a
$move\mbox{-}L^0\mbox{-}L_i^1$ action at time $t+1$, and using
regression step $A$ at $t+2$ enforces a
$move\mbox{-}L^0\mbox{-}L_j^1$ action at time $t+2$. The former
action (variable setting) forces, by UP over EC clauses, all
precondition achievers of the latter action to be set to $0$ at time
$t+1$, yielding an empty AC clause.
\end{itemize}
\end{sloppypar}

Figure~\ref{XFIG/map-base-3cases.eps} illustrates the three
possible situations. We now perform a case distinction over these
situations, starting with the two extreme cases.

\widthfig{14}{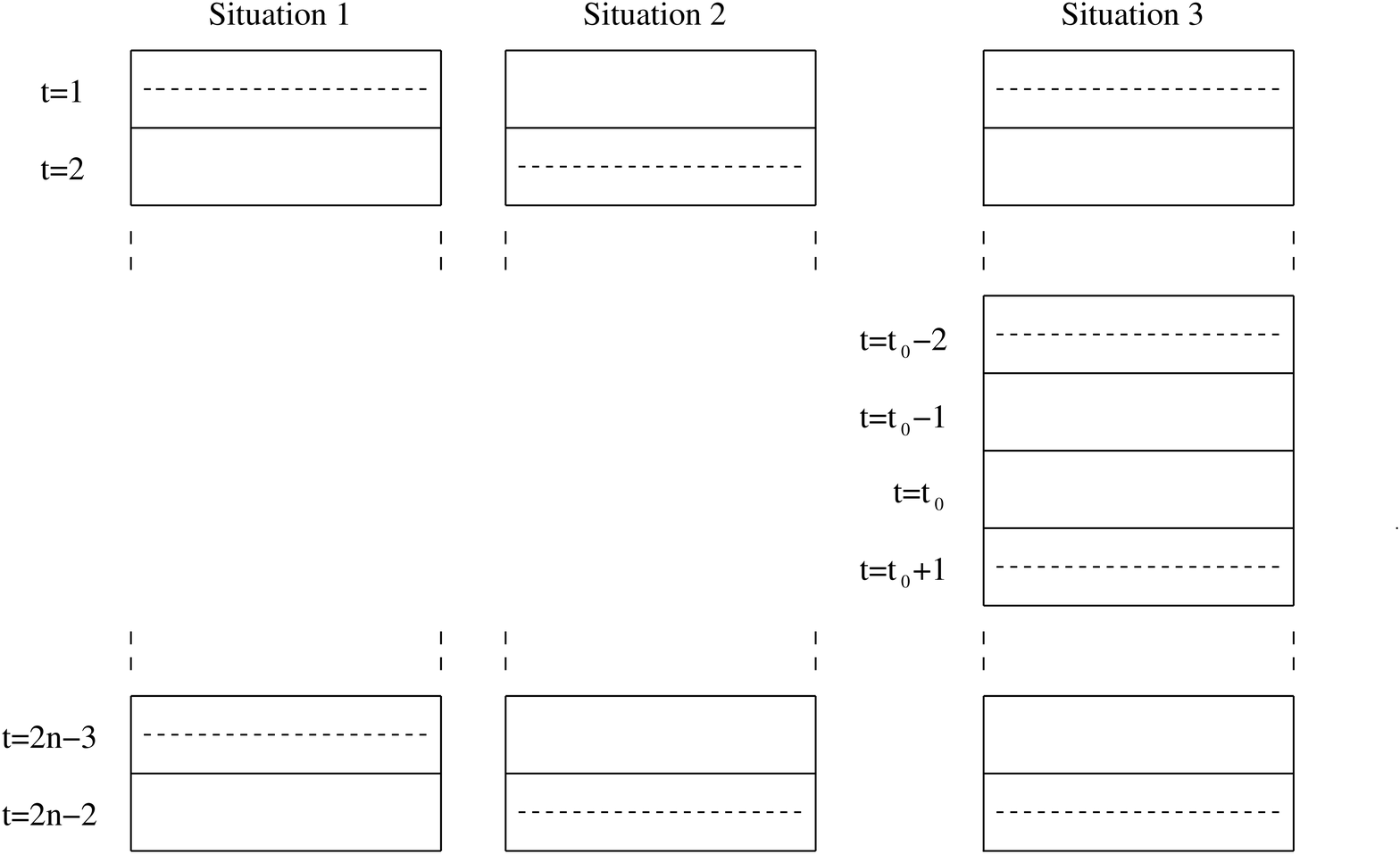}{An illustration of the 
three different situations in our final case distinction. The crossed
out fields stand for time steps in the encoding at which a
$move\mbox{-}L^0\mbox{-}L_{i}^1$ variable is set to $1$, and thus
all other move variables are forced out by UP over EC
clauses.}{-0.0cm}

{\bf Situation 1, $t_0 = 2n-1$.} {\em At all $t \in \{1\} \cup T$,
regression step A is used.} This means that
$move\mbox{-}L^0\mbox{-}L_{1}^1$ is forced to $0$ at {\em all} time
steps of the CNF, which obviously implies that the GC clause for
branch $1$ becomes empty in UP.

{\bf Situation 2, $t_0 = 1$.} {\em At all $t \in \{1\} \cup T$,
regression step B is used.} Just like above, this means that
$move\mbox{-}L^0\mbox{-}L_{1}^1$ is forced to $0$ at all time steps
of the CNF. This is obvious for $t>1$; for $t=1$ it holds because, in
this situation, $NOOP\mbox{-}at\mbox{-}L^0(1)$ is set to $1$ by $a$.

\begin{sloppypar}
{\bf Situation 3, $t \in T$.} {\em Regression step A is used below
$t_0$, and regression step B is used at and above $t_0$.} This case is
a little more tricky; it is the only point where the $|T|-2$
$move\mbox{-}L^0\mbox{-}L_1^1$ variables in $MAP_n^1B$ come into the
play.

First, observe that $move\mbox{-}L^0\mbox{-}L_{1}^1$ is forced to
$0$ at all time steps except $t_0$. Regarding the situation at $t_0$,
observe that, if $t_0 \in T \setminus \{2n-5, 2n-3\}$, then the
variable $move\mbox{-}L^0\mbox{-}L_{1}^1(t_0)$ is contained in
$MAP_n^1B$. In that case, that variable must be set to $0$ by $a$
because otherwise UP gets a contradiction with the
$move\mbox{-}L^0\mbox{-}L_{i}^1$ variable already set to $1$ by
regression step B at step $t_0+1$. Thus, if $t_0 \in T \setminus
\{2n-3, 2n-5\}$, $move\mbox{-}L^0\mbox{-}L_{1}^1$ is forced to
$0$ at {\em all} time steps in the CNF and as above the GC clause for
branch $1$ becomes empty in UP. We need to consider the case where
$t_0 \in \{2n-5, 2n-3\}$. We show that
$move\mbox{-}L_1^2\mbox{-}L_{1}^1$ is forced out at all time steps;
below we use this to finish the proof. Clearly,
$move\mbox{-}L_1^2\mbox{-}L_{1}^1$ is forced out at all time steps $t
\leq t_0$. If $t_0 = 2n-3$, then the only thing we need to show is that 
$move\mbox{-}L_1^2\mbox{-}L_{1}^1$ is set to $0$ at step $2n-2$, which
it is, due to the $move\mbox{-}L^0\mbox{-}L_{i}^1(2n-2)$ variable
set to $1$ by regression step B. If $t_0 = 2n-5$, then at $2n-3$ the
action $move\mbox{-}L_1^1\mbox{-}L_{1}^2$, a precondition achiever of
$move\mbox{-}L_1^2\mbox{-}L_{1}^1$, is {\em not} set to $0$. However,
by the respective AC clause, $move\mbox{-}L_1^2\mbox{-}L_{1}^1$ is
still set to $0$, because its precondition achievers were forced out
at all earlier time steps.

We now know that $move\mbox{-}L_1^2\mbox{-}L_{1}^1$ is set to $0$ at
all time steps in the CNF. In effect, we get that the GC clause for
branch $1$ simplifies to $\{NOOP\mbox{-}visited\mbox{-}L_1^1(2n-2),$
$move\mbox{-}L^0\mbox{-}L_1^1(2n-2)\}$. But
$move\mbox{-}L^0\mbox{-}L_1^1(2n-2)$ is forced to $0$ so we get the
unit clause $\{NOOP\mbox{-}visited\mbox{-}L_1^1(2n-2)\}$. This clause
gives us, by the respective AC clause, the same goal constraint one
time step below, $\{NOOP\mbox{-}visited\mbox{-}L_1^1(2n-1),$
$move\mbox{-}L^0\mbox{-}L_1^1(2n-1)\}$. Iterating the argument, we
get the goal constraint $\{NOOP\mbox{-}visited\mbox{-}L_1^1(t_0),$
$move\mbox{-}L^0\mbox{-}L_1^1(t_0)\}$. But, since
$move\mbox{-}L^0\mbox{-}L_1^1$ is forced to $0$ at all steps $t <
t_0$, by AC clause $NOOP\mbox{-}visited\mbox{-}L_1^1(t_0)$ is set to
$0$, and this clause simplifies to
$\{move\mbox{-}L^0\mbox{-}L_1^1(t_0)\}$. So this move is forced in
at time $t_0$. But, since regression step B is used at $t_0$, a
different $move\mbox{-}L^0\mbox{-}L_i^1$ is forced in at step
$t_0+1$. As before, this yields a contradiction by UP. This concludes
our argument.
\end{sloppypar}
\end{proof}

\bigskip

\noindent {\bf Theorem~\ref{mapbottom:upcons}.} {\em Let $B'$ be a
  subset of $MAP_n^1B$ obtained by removing one variable.  Then the
  number of UP-consistent assignments to the variables in $B'$ is
  always greater than $0$, and at least $(n-3)!$ for $n \geq 3$.}

\begin{proof}
We use the terminology introduced in the proof to
Theorem~\ref{mapbottom}. Basically, the idea behind the proof is to do
at least $n-3$ regression steps A or B, at $n-3$ time step $t$. This
gives us the desired lower bound since, at every regression step, we
will have a choice of what branch $x^t$ to ``regress on'', i.e. at
each step we can choose the branch whose goal location is visited: we
distribute the branches over the time steps. The details are a bit
involved since we have to specify exactly what values to assign to all
variables in $B'$, depending on what variable is missing.

First, if $n=2$ then $MAP_n^1B$ contains the single variable
$NOOP\mbox{-}at\mbox{-}L^0(1)$, and $B'$ is empty. The claim follows
then simply because $MAP_2^1$ is not solved by UP. We henceforth
assume $n \geq 3$. Say $v$ is the variable in $MAP_n^1B
\setminus B'$. We distinguish between four cases, regarding the type
of $v$.

  {\bf Case 1, $v = NOOP\mbox{-}at\mbox{-}L^0(1)$.} Then we
  distribute the $n-2$ branches $3, \dots, n$ over the $n-2$ time
  steps $T = \{3, 5, \dots, 2n-3\}$, using exclusively regression step
  A. Precisely, we set the variables in the following way. For each $i
  \in \{3, \dots, n\}$: there is exactly one $t \in T$ where
  $move\mbox{-}L^0\mbox{-}L^1_i(t)$ is set to $1$; for all other
  $t'$, $move\mbox{-}L^0\mbox{-}L^1_i(t')$ is set to $0$; for $t' >
  t$, $NOOP\mbox{-}visited\mbox{-}L_i^1(t')$ is set to $1$; for $t'
  \leq t$, $NOOP\mbox{-}visited\mbox{-}L_i^1(t')$ is set to $0$. The
  $move\mbox{-}L^0\mbox{-}L^1_1(t)$ variables, and the
  $move\mbox{-}L^0\mbox{-}L^1_2(t)$ variables, are all set to
  $0$. In effect, the goal constraint for branch $2$ gets transported
  to step $2$ ($move\mbox{-}L^0\mbox{-}L^1_2$ is set to $0$ at all
  later steps). However, the goal constraint for branch $1$ does not
  get transported to step $2$. This is because
  $move\mbox{-}L^1_1\mbox{-}L^2_1$ is not set to $0$ at step $2$, and
  thus $move\mbox{-}L^2_1\mbox{-}L^1_1$ is not set to $0$ at steps $4,
  \dots, 2n-2$, which implies that the GC clause for branch $1$
  remains binary after UP: $\{move\mbox{-}L^2_1\mbox{-}L^1_1(2n-2),$
  $NOOP\mbox{-}visited\mbox{-}L_{1}^1(2n-2)\}$. So UP does not yield
  an empty clause, which concludes this argument. (We remark that,
  even if the goal constraint for branch $1$ {\em was} transported to
  step $2$, UP would not yield an empty clause without branching over
  $NOOP\mbox{-}at\mbox{-}L^0(1)$.)

  {\bf Case 2, $v = move\mbox{-}L^0\mbox{-}L^1_{1}(t_0)$.} We have
  $t_0 \in T \setminus \{2n-5,2n-3\}$. We distribute the $n-1$
  branches $\{2, \dots, n\}$ over the $n-1$ time steps $\{1, 3, \dots,
  t_0-2, t_0+1, t_0+3, \dots, 2n-2\}$. This corresponds to situation 3
  in Figure~\ref{XFIG/map-base-3cases.eps}, i.e., we use regression
  step A at all steps below $t_0$, and we use regression step B at all
  steps at and above $t_0$. Obviously, this can be done by setting all
  the variables, in a similar fashion as above in case 1, except for
  the differences necessary to produce regression step B (namely, set
  all move variables at $t$ to $0$, and set all NOOP variables at $t$,
  except one, to $1$), and for having to set
  $NOOP\mbox{-}at\mbox{-}L^0(1)$ to $0$. We do not get an empty
  clause in UP because $move\mbox{-}L^0\mbox{-}L^1_{1}$ is not set
  to $0$ at step $t_0$: in effect, $move\mbox{-}L^1_1\mbox{-}L^2_{1}$
  is not set to $0$ at steps $t_0+2, t_0+4, \dots 2n-3$, and
  $move\mbox{-}L^2_1\mbox{-}L^1_{1}$ is not set to $0$ at steps
  $t_0+4, t_0+6, \dots 2n-3$. Note that $t_0+4 \leq 2n-3$ because $t_0
  \in T \setminus \{2n-5,2n-3\}$. Hence the goal constraint for branch
  $1$ remains binary at step $2n-3$ after UP:
  $\{move\mbox{-}L^2_1\mbox{-}L^1_1(2n-3),$
  $NOOP\mbox{-}visited\mbox{-}L_{1}^1(2n-3)\}$, which concludes this
  argument.

  {\bf Case 3, $v = move\mbox{-}L^0\mbox{-}L^1_{i_0}(t_0)$, $i_0
  \neq 1$.} Then we distribute the $n-2$ branches $\{2, \dots, n\}
  \setminus \{i_0\}$ over the $n-2$ time steps $\{1, 3, \dots, t_0-2,
  t_0+3, \dots, 2n-2\}$. This corresponds to situation 3 in
  Figure~\ref{XFIG/map-base-3cases.eps}, except that no regression
  step is used at step $t_0$ (i.e., no move variable gets set to $1$
  at $t_0+1$). Precisely, at $t_0$ we set all present move variables
  to $0$, and we set the NOOP variables for $i \neq i_0$ according to
  the distribution (i.e., to $1$ if a move to $i$ is assigned earlier,
  and to $0$ if it is assigned later). We set the NOOP for $i_0$, if
  present (i.e. if $i_0 \neq 2$), to $0$. With this, the goal
  constraint for $i_0$ is transported to $t_0+1$. If $t_0 \in \{2n-3,
  2n-5\}$, then there is no $move\mbox{-}L^0\mbox{-}L^1_{1}(t_0)$
  variable in $MAP_n^1$, so both branch $1$ and branch $i_0$ can move
  to their goal locations in either $t_0$ or $t_0+1$ (the respective
  move variables are not set nor forced to $0$), which implies that no
  unit clause arises in UP: the goal constraints for both branches are
  binary in $t_0+1$. If $t_0 \not \in \{2n-3, 2n-5\}$, then a similar
  argument as above in case 2 holds. We have that
  $move\mbox{-}L^0\mbox{-}L^1_{1}$ is not set to $0$ at step
  $t_0+1$. In effect, $move\mbox{-}L^1_1\mbox{-}L^2_{1}$ is not set to
  $0$ at steps $t_0+2, t_0+4, \dots 2n-3$, and
  $move\mbox{-}L^2_1\mbox{-}L^1_{1}$ is not set to $0$ at steps
  $t_0+4, t_0+6, \dots 2n-3$. The goal constraint for branch $1$
  remains binary at step $2n-3$ after UP, concluding this argument.

  {\bf Case 4, $v = NOOP\mbox{-}visited\mbox{-}L_{i_0}^1(t_0)$.}  We
  have $t_0 \in T = \{3, 5, \dots, 2n-3\}$. In this case we end up
  distributing only $n-3$ branches, over $n-3$ time steps. This
  results in the somewhat weaker lower bound $(n-3)!$ rather than
  $(n-2)!$ or $(n-1)!$ as above. Precisely, we distribute the branches
  $\{3, \dots, n\} \setminus \{i_0\}$ over the time steps $T \setminus
  \{t_0\}$. That is, we omit steps $1$ and $t_0$, and use regression
  step A at all other steps (in $T$). For branches $1$, $2$, and
  $i_0$, we set all move variables to $0$. For branch $i_0$, we set
  all NOOP variables above $t_0$ to $1$ and all NOOP variables below
  $t_0$ to $0$. We set $NOOP\mbox{-}at\mbox{-}L^0(1)$ to $0$. What
  happens under UP is this. Setting $NOOP\mbox{-}at\mbox{-}L^0(1)$
  to $0$ forces all $move\mbox{-}L^0\mbox{-}L^1_{i}$ actions out at
  time step $2$, but leaves them unaffected at time step $1$. Since
  the $move\mbox{-}L^0\mbox{-}L^1_{i}$ actions at time steps above
  $t_0+1$ are all forced out, the goal constraints for branches $2$
  and $i_0$ are transported to step $t_0+1$. In that step, both moves
  to these branches are still possible. Since the moves are also
  possible in step $1$, the goal constraints are binary, i.e., include
  the move variable and the respective NOOP variable, and so no
  further propagation is triggered. As for branch $1$, since
  $move\mbox{-}L^0\mbox{-}L^1_{1}$ is possible at step $1$,
  $move\mbox{-}L^1_1\mbox{-}L^2_{1}$ is possible at step $2$, and
  $move\mbox{-}L^2_1\mbox{-}L^1_{1}$ is possible at steps $4, 6, \dots
  2n-2$. So the goal constraint for branch $1$ is binary at step
  $2n-2$, and does not get transported even to step $2n-3$. This
  concludes our argument.
\end{proof}

\subsubsection{Asymmetrical Case}
\label{proofs:map:top}

  The backdoor we identify in this case, named $MAP_n^{2n-3}$, is
defined as: {\small
\medskip
\begin{tabbing}
$MAP_n^{2n-3}B$ := \=
$\{move\mbox{-}L_1^{2^i-2}\mbox{-}L_1^{2^i-1}(2^i-1)
\mid 1 \leq i \leq \lceil log_2 n \rceil\}$
\end{tabbing}}

\bigskip

\noindent {\bf Theorem~\ref{maptop}.} {\em $MAP_n^{2n-3}B$ is a
  backdoor for $MAP_n^{2n-3}$.}

\begin{proof}
\begin{sloppypar}
For simplicity of notation, we use the convention that $L_1^0$ stands
for $L^0$. We start with two observations:
\begin{itemize}
\item {\bf Observation 1.} {\em If $move\mbox{-}L_1^i\mbox{-}L_1^{i+1}(i+1)$
    is set to $1$, then UP sets
    $move\mbox{-}L_1^j\mbox{-}L_1^{j+1}(j+1)$ to 1 for all $1 \leq j <
    i$.}

\medskip

    This is because, from $i$ downwards, the only precondition
    achiever for $move\mbox{-}L_1^j\mbox{-}L_1^{j+1}(j+1)$ in time
    step $j$ is $move\mbox{-}L_1^{j-1}\mbox{-}L_1^{j}(j)$. So the
    settings occur via UP over the respective AC clauses.

\medskip

\item {\bf Observation 2.} {\em If $move\mbox{-}L_1^i\mbox{-}L_1^{i+1}(i+1)$
    is set to, then UP sets $move\mbox{-}L_1^j\mbox{-}L_1^{j+1}(j+2)$
    to $1$ for all $i \leq j \leq 2n-4$.}  

\medskip

  First, observe that $move\mbox{-}L_1^j\mbox{-}L_1^{j+1}(j+1)$ is set
  to $0$ by UP for $i < j \leq 2n-4$. This happens by chaining over AC
  clauses just as observed above. In effect, also by chaining over AC
  clauses, $NOOP\mbox{-}at\mbox{-}L_1^{j+1}(j+2)$ and
  $NOOP\mbox{-}visited\mbox{-}L_1^{j+1}(j+2)$ are set to $0$ by UP for
  $i < j \leq 2n-4$: the only precondition supporter for these actions
  at step $j+1$ is $move\mbox{-}L_1^j\mbox{-}L_1^{j+1}(j+1)$, which is
  set to $0$. For $j=2n-4$, we get that
  $NOOP\mbox{-}visited\mbox{-}L_1^{2n-3}(2n-2)$ is set to $0$. Thus
  the goal constraint for branch $1$ becomes unit, and
  $move\mbox{-}L_1^{2n-4}\mbox{-}L_1^{2n-3}(2n-2)$ is set to $1$. The
  respective AC clause simplifies to the unit clause containing only
  $move\mbox{-}L_1^{2n-5}\mbox{-}L_1^{2n-4}(2n-3)$, because
  $NOOP\mbox{-}at\mbox{-}L_1^{2n-4}(2n-3)$ is already set to $0$. In a
  similar fashion, all the $move\mbox{-}L_1^j\mbox{-}L_1^{j+1}(j+2)$
  variables are set to $1$ down to step $i+2$.
\end{itemize}

\noindent
We prove below that:
\begin{itemize}
\item {\bf Observation 3.} {\em For all $1 \leq i \leq \lceil log_2 n \rceil$,
  $move\mbox{-}L_1^{2^i-2}\mbox{-}L_1^{2^i-1}(2^i-1)$ must be set to
  $1$ or else UP yields a contradiction.}
\end{itemize}

This suffices because setting $move\mbox{-}L_1^{2^{\lceil log_2 n
\rceil}-2}\mbox{-}L_1^{2^{\lceil log_2 n \rceil}-1}(2^{\lceil log_2 n
\rceil}-1)$ to $1$ yields a contradiction by UP.  Let us first show the
latter. With observation 1, below $2^{\lceil log_2 n \rceil}$, in
every time step a $move$ variable is set to $1$ so all other $move$
variables are forced out; in particular,
$move\mbox{-}L^0\mbox{-}L_2^1$. To get to a point where
$move\mbox{-}L^0\mbox{-}L_2^1$ can be re-inserted, i.e., to get to a
time step at which that action variable is not set to $0$, one needs
to ``go back from $L_1^{2^{\lceil log_2 n \rceil}-1}$ to
$L^0$''. Which means, by sequencing UP over AC clauses of the actions
moving back from $L_1^{2^{\lceil log_2 n \rceil}-1}$ to $L^0$'', UP
shows that it takes $2^{\lceil log_2 n \rceil}-1$ steps before one
gets to a step where $move\mbox{-}L^0\mbox{-}L_2^1$ can be applied
again. So the goal for branch $2$ can be first achieved $2^{\lceil
log_2 n \rceil}-1$ steps after step $2^{\lceil log_2 n
\rceil}-1)$ (where we just set a move variable to $1$). Note that $2^{\lceil log_2
n \rceil} \geq n$. Now, after time step $2^{\lceil log_2 n \rceil}$,
there are only $2n-2^{\lceil log_2 n \rceil}-1$ time steps left in the
encoding. With $2^{\lceil log_2 n \rceil} \geq n$, we have at most
$n-1$ layers left, which is less than the number of layers, again
$2^{\lceil log_2 n \rceil} \geq n$, before the branch $2$ goal becomes
re-achievable. So the branch $2$ goal constraint becomes empty at step
$2n-2$.

We now prove observation 3 by induction over $i$. Base case, $i = 1$,
$move\mbox{-}L^0\mbox{-}L_1^1(1)$. If we set that variable to $0$,
then, by observation 2, all $move\mbox{-}L_1^j\mbox{-}L_1^{j+1}(j+1)$
variables are forced to $1$ for $1 \leq j
\leq 2n - 3$.  This means, in particular, that
$move\mbox{-}L^0\mbox{-}L_2^1(t)$ is forced to $0$ at all layers $t >
1$. So the branch $2$ goal constraint gets transported, by chaining
over the AC clauses of the respective NOOP actions, down to time step
$1$. Consequently, $move\mbox{-}L^0\mbox{-}L_2^1(1)$ gets forced to
$1$, in contradiction to $move\mbox{-}L^0\mbox{-}L_1^1(2)$ which was
already set to $1$ earlier, and now loses its precondition support,
yielding an empty AC clause.

Inductive case, $i \rightarrow i+1$. We assume that
$move\mbox{-}L_1^{2^i-2}\mbox{-}L_1^{2^i-1}(2^i-1)$ is set to 1.  By
observation 1, we get that $move\mbox{-}L_1^j\mbox{-}L_1^{j+1}(j+1)$
is set to $1$ for all $1 \leq j < 2^i-1$.  So
$move\mbox{-}L^0\mbox{-}L_2^1(t)$ is forced to $0$ for all time steps $t
\leq 2^i-1$. Now, say we set
$move\mbox{-}L_1^{2^{i+1}-2}\mbox{-}L_1^{2^{i+1}-1}(2^{i+1}-1)$ to 0.
Then, by observation 2, $move$ actions are forced in at all time steps
$t > 2^{i+1}-1$. So $move\mbox{-}L^0\mbox{-}L_2^1(t)$ is also forced
to $0$ for all layers $t > 2^{i+1}-1$. With that, the branch $2$ goal
constraint is transported to time step $2^{i+1}-1$. Observe that,
between step $2^i-1$ and step $2^{i+1}-1$, there are $2^i$ layers. But
the number of steps we need to make the precondition of
$move\mbox{-}L^0\mbox{-}L_2^1$ achievable, i.e. to ``go back'' from
$L_1^{2^i-1}$ to $L^0$, is $2^i-1$. So $move\mbox{-}L^0\mbox{-}L_2^1$
first re-appears (is not set to $0$ by UP over AC clauses) in time
step $2^{i+1}-1$. As observed before, we have the goal constraint for
branch $2$ at that time step. Since the time step is now the first one
in the CNF where $move\mbox{-}L^0\mbox{-}L_2^1$ is not set to $0$,
$NOOP\mbox{-}visited\mbox{-}L_2^{1}$ is not available (forced to $0$),
and the goal constraint is unit. Thus
$move\mbox{-}L^0\mbox{-}L_2^1(2^{i+1}-1)$ is set to $1$. By chaining
over AC clauses, also all the move actions that ``go back'' from
$L_1^{2^i-1}$ to $L^0$, within the time steps $2^i, \dots, 2^{i+1}-2$,
are set to $1$ by UP. But then, the precondition clause of
$move\mbox{-}L_1^{2^{i+1}-2}\mbox{-}L_1^{2^{i+1}-1}(2^{i+1})$, which
was set to $1$ earlier, becomes empty. This concludes the argument.
\end{sloppypar}
\end{proof}

The reader who is confused by all the $(2^{x}-y)$ indexing in the
proof to Theorem~\ref{maptop} is advised to have a look at
Figure~\ref{XFIG/map-top-BD.eps}, and instantiate the indices in the
proof with the numbers 1, 3, and 7.

\bigskip

\noindent {\bf Theorem~\ref{maptop:upcons}} {\em Let $B'$ be a subset
  of $MAP_n^{2n-3}B$ obtained by removing one variable.  Then there is
  exactly one UP-consistent assignment to the variables in $B'$.}

\begin{proof}
  The claim follows quite easily from the proof to
  Theorem~\ref{maptop}. We use the notations from that proof. We
  observe the following:
\begin{sloppypar}
\begin{enumerate}
\item If a $MAP_n^{2n-3}B$ variable $v$ at time $t$ is set to $1$,
  then all the $MAP_n^{2n-3}B$ variables at times $t' < t$ are set to
  $1$ by UP. This follows from observation 1 in the proof to
  Theorem~\ref{maptop}.
\item If the topmost variable in $MAP_n^{2n-3}B$ (the variable at 
   the latest time step) is set to $1$, then we get an empty clause in
   UP. This is shown below observation 3 in the proof to
   Theorem~\ref{maptop}.
\item If a $MAP_n^{2n-3}B$ variable $v$ at time $t$ is set to $0$,
  then all the $MAP_n^{2n-3}B$ variables at times $t' > t$ are set to
  $0$ by UP. This follows from observation 2 in the proof to
  Theorem~\ref{maptop}.
\item If the lower most variable in $MAP_n^{2n-3}B$ (the variable at 
   time step $1$) is set to $0$, then we get an empty clause in
   UP. This corresponds to the base case in the induction in the proof
   to Theorem~\ref{maptop}.
\item If we have two consecutive $MAP_n^{2n-3}B$ variables $v$ and $v'$, i.e. $v =
  move\mbox{-}L_1^{2^i-2}\mbox{-}L_1^{2^i-1}(2^i-1)$ for some $1 \leq
  i < \lceil log_2 n \rceil$, and $v' =
  move\mbox{-}L_1^{2^{i+1}-2}\mbox{-}L_1^{2^{i+1}-1}(2^{i+1}-1)$, and
  $v$ is set to $1$ and $v'$ is set to $0$, then we get an empty
  clause during UP. This is because there are not enough time steps
  between $2^i-1$ and $2^{i+1}$ to ``go back'' from $L_1^{2^i-1}$ to
  $L^0$, and to accommodate the necessary move to $L_2^1$ at time step
  $2^{i+1}-1$. This corresponds to the inductive case in the induction
  in the proof to Theorem~\ref{maptop}.
\end{enumerate}
\end{sloppypar}
Now, say $v = move\mbox{-}L_1^{2^i-2}\mbox{-}L_1^{2^i-1}(2^i-1)$, for
some $1 \leq i \leq \lceil log_2 n \rceil$, is the left-out variable,
i.e., $v \in MAP_n^{2n-3}B \setminus B'$. With the above, the only
chance we have to avoid a contradiction in UP is to set all variables
at times above $2^i-1$ to $0$, and all variables at times below
$2^i-1$ to $1$. Indeed, in such a variable assignment we don't get a
contradiction with UP because, between $2^{i-1}-1$ and $2^{i+1}$,
there are enough time steps to ``go back'' from $L_1^{2^i-1}$ to
$L^0$, and accommodate the move to $L_2^1$. This concludes the
argument.
\end{proof}

%% file: SBW-proofs.tex
\subsection{SBW  Backdoors}
\label{proofs:sbw}

We first consider the symmetrical case, then the asymmetrical case. Remember that,
to simplify notation, we use the convention that $g_0$ denotes $t_2$,
i.e. $t_2$ is the ``lower most good block''.

\subsubsection{Symmetrical Case}
\label{proofs:sbw:bottom}

In the case $k=0$, there are only ``good'' blocks, with no
restrictions whatsoever on the possible stacking order. With $n=2$
this is inconsistent under UP, so we assume $n > 2$.

As stated in Section~\ref{sd:sbw}, the backdoor we identify for
$SBW_n^0$, denoted by $SBW_n^0B$, is defined as follows (compare
Figure~\ref{XFIG/sbw-base-newBD.eps}):
{\small
\medskip
\begin{tabbing}
  $SBW_n^0B$ := \=  $\{movetot2\mbox{-}g_i\mbox{-}g_j(t) \mid 1 \leq i \leq n-2, 
  0 \leq j \leq n, j \neq i, 2 \leq t \leq n-1\} \; \; \setminus$\\ 
  \> $\{movetot2\mbox{-}g_i\mbox{-}g_j(i+1) \mid max(2,n-4) \leq i \leq n-2, 0 \leq
  j \leq n-2\}$
\end{tabbing}}

\bigskip

\noindent {\bf Theorem~\ref{sbwbottom}.} {\em $SBW_n^0B$ is a backdoor
  for $SBW_n^0$.}

\begin{proof}
We assume we are given any truth value assignment $a$ to the variables
in $SBW_n^0B$. Our task is to show that an empty clause arises under
UP. For simplicity of notation, we identify blocks $g_i$ with their
indices $i$. We will also use indices $i$ interchangeably as indices
into the blocks and as indices into the time steps. We refer to the
blocks $2, \dots, n-2$ as the {\em cut-affected} blocks. For a
cut-affected block $i$ we refer to the move variables for $i$ at time
$i+1$, and to time step $i+1$ itself, as the {\em cut-field} of
$i$. We say that a time step $t$ is {\em occupied} if some move
variable is set to $1$ at $t$, or if all move variables are set to $0$
at $t$ (for some other reason). We say that a set $B$ of blocks {\em
occupies} a set $T$ of time steps if all $t \in T$ are occupied, and
the only move variables set to $1$ at any $t$ have the form
$movetot2\mbox{-}i\mbox{-}j(t)$, with $i \in B$. We say that a
block $i$ is {\em assigned} if there is a time step $t$ where some
$movetot2\mbox{-}i\mbox{-}j(t)$ variable is set to $1$.

We use the following notations. By $A$ we denote the set of blocks for
which we have all variables, i.e., the set of blocks that are not
cut-affected: the set $\{1\} \cup \{2, \dots, n-4\}$. By $CA$ we
denote the set of cut-affected blocks whose cut-field is occupied
after UP. By $\overline{CA}$ we denote the set of cut-affected blocks
whose cut-field is {\em not} occupied after UP. For a block $b \in
\overline{CA}$, it may be that $b$ is assigned at some $t$ other than 
its cut-field. By $\overline{CA}_{|a}$ we denote these blocks, and by
$\overline{CA}_{|\neg a}$ we denote the members of $\overline{CA}$
that are not assigned. The main proof argument will be to show that
$\overline{CA}|_{\neg a}$ is, in fact, empty. For better flow of
language we will sometimes refer to the blocks $A \cup CA \cup
\overline{CA}_{|a}$ as ``the assigned'' blocks, and to $\overline{CA}_{|\neg
  a}$ as ``the open'' blocks. If we have a set $I$ of blocks/block
indices, by $I\pp$ we denote the set of time steps $\{i+1 \mid i \in
I\}$ -- a shorthand to refer to the cut-fields of the blocks in $I$.
For example, $CA\pp$ denotes the set of occupied cut-fields. Note that
$CA\pp$ and $\overline{CA}\pp$ partition the set $\{max(3, n-3) \dots,
n-1\}$.

We start the proof with a basic observation about the restrictions
imposed by sequences of block moves.

\begin{sloppypar}
\begin{itemize}
\item {\bf Observation 1a.} {\em Say all time steps up to a time $t$
    are occupied by the blocks $B$, and block $b$ is moved last
    ($b=t_2$ if $B$ is empty). Then at time $t+1$, all
    $movetot2\mbox{-}i\mbox{-}{b'}$ variables, $b' \neq b$, are
    forced to $0$ by UP.}

\medskip

This is due to a simple UP chaining over precondition (AC) clauses: no
block other than $b$ can be clear and above $t_2$ at step $t+1$. More
formally, the argument is this. For $b' \not \in B$, all actions of
the form $movetot2\mbox{-}{b'}\mbox{-}{j}$ are forced to $0$ at
all steps $t' \leq t$, so UP detects that $b'$ can not be above $t_2$
at step $t+1$. For $b' \in B \setminus \{b\}$, moved to $t_2$ at step
$t'$, $NOOP\mbox{-}clear\mbox{-}b'$ is forced to $0$ at step $t'+1$,
and all actions achieving $clear\mbox{-}b'$, namely the actions that
move the block on $b'$ back to $t_1$, are forced to $0$ up to step
$t$. So UP detects that $b'$ can not be clear at step $t+1$.

\item {\bf Observation 1b.} {\em Say all time steps up to a time $t$
    are occupied by the blocks $B$, and block $b \in B$ is moved
    last. Say further that $NOOP\mbox{-}abovet_2\mbox{-}b$ is set to
    1 at all times $t' > t$. Then at all times $t' > t+1$, all
    $movetot2\mbox{-}i\mbox{-}{b'}$ variables, $b' \in B \cup
    \{0\} \setminus \{b\}$, are forced to $0$ by UP.}

\medskip

This is also due to UP chaining over AC clauses. Since
$NOOP\mbox{-}abovet_2\mbox{-}b$ is set to $1$ at all times $t'$, $b$
can not be moved away from where it is, so by UP over EC clauses we
get that the block $b'$ directly below $b$ can not be made clear. By
chaining over AC clauses, we get that the block directly below $b'$
can not be made clear either, and so forth until $t_2 = g_0$.
\end{itemize}
\end{sloppypar}

\noindent
We now observe that we get a contradiction in UP if only a single time
step is open (not occupied).

\begin{sloppypar}
\begin{itemize}
\item {\bf Observation 2.} {\em If at least $n-2$ time steps are occupied, 
then UP yields a contradiction.}

\medskip

If all $n-1$ time steps are occupied, then the goal constraints for
$g_{n-1}$ and $g_n$ become empty at step $n-1$. Otherwise, these goal
constraints are transported to the open step $t_0$. Let $b$ be the
block moved at step $t_0-1$ (or $b = t_2$ if $t_0 = 1$). With
observation 1a, the only available move actions for $g_{n-1}$ and
$g_n$ at step $t_0$ are $movetot2\mbox{-}{n-1}\mbox{-}b$ and
$movetot2\mbox{-}n\mbox{-}b$. They will both be set to $1$, yielding
an empty EC clause.
\end{itemize}
\end{sloppypar}

\noindent
We now observe that, with an occupied cut-field, any block occupies at
least one time step.

\begin{sloppypar}
\begin{itemize}
\item {\bf Observation 3.} {\em For $i \in A \cup CA$, there exist
    a time step $t$ and block $j$ such that, after UP,
    $movetot2\mbox{-}i\mbox{-}j(t)$ is set to $1$.}

\medskip

  If any variable $movetot2\mbox{-}i\mbox{-}j(t)$, $t>1$, is set
  to $1$, there is nothing to show. Otherwise, we know that all these
  variables are set to $0$ (either by $a$ or by UP over EC
  clauses). Thus the goal constraint for $i$ at step $n-1$, i.e. the
  clause $\{NOOP\mbox{-}abovet_2\mbox{-}i\} \cup$
  $\{movetot2\mbox{-}i\mbox{-}j(n-1) \mid 0 \leq j \leq n, j \neq
  i\}$, simplifies to the unit clause
  $\{NOOP\mbox{-}abovet_2\mbox{-}i\}$. Then the AC clause of that
  NOOP variable simplifies to the same goal constraint at step $n-2$.
  That is, the goal constraint for $i$ is transported from $n-1$ to
  $n-2$. The same argument applies downwards over the time steps until
  the goal constraint for $1$ becomes present at step $1$. In that
  step, the only present action -- the only action that can bring $i$
  above $t_2$ -- is, of course, $movetot2\mbox{-}i\mbox{-}t_2$, so
  the respective variable is set to $1$, which concludes this
  argument.
\end{itemize}
\end{sloppypar}

Note that we have now already shown that $SBW_n^0B$ {\em without} the
cut-fields is a backdoor: if $A$ was the set of all blocks $\{1,
\dots, n-2\}$, then by observation 3 we would have $n-2$ occupied time 
steps, and by observation 2 we would have an empty clause in UP. While
this was relatively easy to prove, it is rather tricky to prove that
the backdoor property gets preserved when actually removing the
cut-field variables. The following observation delivers the case
distinction that will make this proof possible. By $t_a$ we denote the
highest step in $A\pp$, i.e., the highest time step with no cut-field
($t_a = max(2, n-4)$).

\begin{sloppypar}
\begin{itemize}
\item {\bf Observation 4.} {\em $|\overline{CA}_{|a}| \leq 1|$. If 
$|\overline{CA}_{|a}| = 0$ then the blocks $A \cup CA$ occupy all time
    steps $\{1\} \cup A\pp \cup CA\pp$, except at most $t_a$. If
    $|\overline{CA}_{|a}| = 1|$ then the blocks $A \cup CA \cup
    \overline{CA}_{|a}$ occupy all time steps $\{1\} \cup A\pp \cup
    CA\pp$.}

\medskip

Observe the following. First, with observation 3 and the EC clauses,
the blocks $A \cup CA \cup \overline{CA}_{|a}$ occupy at least $|A| +
|CA| + |\overline{CA}_{|a}$ time steps. Second, clearly the occupied
time steps must be taken from the set $\{1\} \cup A\pp \cup CA\pp$
(namely, from the time steps with no or with occupied
cut-fields). From this it follows that $|\overline{CA}_{|a}|
\leq 1|$, simply because $|\{1\} \cup A\pp \cup CA\pp| = |A \cup CA| + 1$. 
If $\overline{CA}_{|a}| = 1|$, then by the same argument all time
steps $t \in \{1\} \cup A\pp \cup CA\pp$ are occupied. If
$\overline{CA}_{|a}| = 0|$, then at most one step in $\{1\} \cup A\pp
\cup CA\pp$ can remain free. The free step can not be step $1$
because, if a $movetot2\mbox{-}i\mbox{-}j$ variable is set to $1$
at step $2$, then UP over a respective AC clause forces
$movetot2\mbox{-}j\mbox{-}t_2$ to $1$ at step $1$. The free step
can't be any of $CA\pp$ since these are all occupied by
construction. Finally, the free step can not be any step $t$ so that
$t+1 \in A\pp$. Assume that this was the case. We then have a sequence
of time steps $\{1, \dots, t-1, t, t+1,
\dots, t_a\}$ where at all these steps except $t$ a move variable is
set to $1$. But, by chaining over EC and AC clauses, this implies that
either a move variable is set to $1$ at $t$ (if $t-1$ moves B onto A
and $t+1$ moves D onto C), or all move variables are set to $0$ at $t$
(if $t-1$ moves B onto A and $t+1$ moves C onto A). In either case,
$t$ is occupied, which is a contradiction.
\end{itemize}
\end{sloppypar}

We are now ready for the final step of the proof. As stated, we will
show that $\overline{CA}_{|\neg a}$ is empty. This suffices with
observations 2 and 3 since it gives us $n-2$ blocks that occupy some
time step. The proof is by contradiction. Assume $x$ is the smallest
(index of a) block in $\overline{AC}_{|\neg a}$. We distinguish three
cases.

{\bf Case 1, $\overline{CA}_{|a} = \emptyset$ and $t_a$ is occupied.}
In this case, since $x$ is the {\em smallest} element of
$\overline{AC}_{|\neg a}$, obviously all time steps below step $x+1$
are occupied. Say step $x$ is occupied by a move for block $b$ (an
occupation through forcing out all moves can only happen at a step $t$
if step $t+1$ is also occupied by an assigned block, which isn't the
case here). That is, $b$ is the ``top'' block. With observation 1a,
$movetot2\mbox{-}x\mbox{-}b$ is the only $movetot2$ action for $x$
not forced to $0$ at time $x+1$. Now, observe that
$NOOP\mbox{-}abovet_2\mbox{-}x$ is set to $0$ at all times $t \leq
x+1$, simply because all the time steps below $x+1$ are occupied.
Also, observe that $NOOP\mbox{-}abovet_2\mbox{-}x$ is set to $1$ at
all times $t > x+1$, due to the goal constraint for $x$ and the fact
that all $movetot2$ actions for $x$ at times greater than $x$ belong
to $SBW_n^0B$ and are set to $0$ (as $x$ does not occupy any time
step). That is, the goal constraint for $x$ is transported down to
step $x+1$, and simplifies to the unit clause setting
$movetot2\mbox{-}x\mbox{-}b$ to $1$. We get $x \in AC$, which is a
contradiction to our assumption.

{\bf Case 2, $\overline{CA}_{|a} = \emptyset$ and $t_a$ is not
occupied.} Here, we can conclude that $x=t_a$, i.e. the cut-field of
$x$ is directly above $t_a$. If this was not the case, then the block
$t_a$ would be an element of $CA$, which would mean that a
$movetot2\mbox{-}i\mbox{-}j$ action was set to $1$ at $t_a+1$, and
another $movetot2\mbox{-}i\mbox{-}j$ action was set to $1$ at
$t_a-1$. But that would imply that $t_a$ is occupied, with the same
argument as at the end of the proof to observation 4.

We have $x=t_a$. All steps below $t_a$ are occupied, say by the blocks
$B$. Say the top block, moved at $t_a-1$, is $b \in B$. We first show
that $NOOP\mbox{-}abovet_2\mbox{-}b$ is set to $1$ at all time steps
$t > t_a-1$. Note that all assigned blocks occupy {\em exactly} one
hole, because $|A \cup CA| + 1 = |\{1\} \cup A\pp \cup O\pp|$. So, in
particular, all $SBW_n^0B$ variables for $b$ at other time steps are
set to $0$. If $b \in A$, then these are {\em all} move variables for
$b$ above $t_a-1$. If $b \in CA$, then the cut-field variables are set
to $0$ because the cut-field is occupied. In both cases, the goal
constraint for $b$ gets transported down to $t_a-1$, and the claim
follows.

With observation 1b, we can now conclude that, at all times $t > t_0$,
in particular at time $t_0+1$, the only $movetot2$ variables that are
not forced to $0$ are of the form
$movetot2\mbox{-}i\mbox{-}{b'}(t)$ where either $b'=b$ or $b' \not
\in B$. We now proceed to show that, for $i=x$ and $t=t_0+1$, the only
available option, i.e. the only such move variable not forced to $0$,
is $movetot2\mbox{-}x\mbox{-}{b}(t_0+1)$. As for
$movetot2\mbox{-}x\mbox{-}(n-1)(t_0+1)$ and
$movetot2\mbox{-}x\mbox{-}{n}(t_0+1)$, these are contained in
$SBW_n^0B$ and set to $0$ by construction. Consider the variables of
the form $movetot2\mbox{-}{b'}\mbox{-}{b}(t_0)$, $b' \not \in
\{n-1, n\}$. All these variables are contained in $SBW_n^0B$ and 
are set to $0$ by construction ($t_0$ is not occupied). That is, the
AC clause of $movetot2\mbox{-}x\mbox{-}{b'}(t_0+1)$ becomes empty
because all precondition achievers are set to $0$ at all steps $t \leq
t_0$. We finally note that, just as argued in case 1 above, the goal
constraint for $x$ is transported down to step $t_0+1$. As a result,
$movetot2\mbox{-}x\mbox{-}{b}(t_0+1)$ gets set to $1$, and $x \in
AC$ in contradiction to our assumption.

{\bf Case 3, $\overline{CA}_{|a} = \{b_a\}$.} Then $t_a$ is
occupied. If $x < b_a$ then all time steps below $x+1$ are occupied,
and we can argue exactly as in case 1. So we can assume that $b_a$ is
the lower most block in $\overline{CA}$. It follows that all time
steps below $b_a+1$ are occupied. It also follows that $x = b_a+1$,
since otherwise the time steps $b_a$ and $b_a+2$ would both be
occupied, which would enforce a move at $b_a+1$ in
contradiction. Thus, like above in case 2, we have that all time steps
below the step $x$ are occupied, i.e., the time step directly below
$x$'s cut-field is open but all time steps further below are occupied,
by a set $B$ of blocks with top block $b$ moved at time $x-1$. We can
now proceed just like in case 2. $NOOP\mbox{-}abovet_2\mbox{-}b$ is
set to $1$ at all time steps $t > x-1$. With observation 1b, at all
times $t > x$, in particular at time $x+1$, the only $movetot2$
variables that are not forced to $0$ are of the form
$movetot2\mbox{-}i\mbox{-}{b'}(t)$ where either $b'=b$ or $b' \not
\in B$. We now proceed to show that, for $i=x$ and $t=x+1$, the only
available option, i.e. the only such move variable not forced to $0$,
is $movetot2\mbox{-}x\mbox{-}{b}(x+1)$. Consider the variables
$movetot2\mbox{-}x\mbox{-}{b'}(x+1)$, $b' \neq b$. For $b' \in
\{n-1, n\}$ these are contained in $SBW_n^0B$ and are set to $0$. For 
$b' \in \{1, \dots, n-2\} \setminus \{b_a\}$, the variable
$movetot2\mbox{-}{b'}\mbox{-}{b}(x)$ is contained in $SBW_n^0B$
and set to $0$, implying that $movetot2\mbox{-}x\mbox{-}{b'}(x+1)$
is set to $0$ by AC clauses. The only thing left to show is that, if
$b_a \neq b$, then $movetot2\mbox{-}{x}\mbox{-}{b_a}(x+1)$ is set
to $0$. If $b_a \in B$, this follows from observation 1b. If $b_a
\not \in B$ (and $b_a \neq b$), then $b_a$ is assigned to some time 
step $t > x+1$. Due to the definition of $SBW_n^0B$, this can only be
time step $x+2$ ($= n-1$), i.e.,
$movetot2\mbox{-}{b_a}\mbox{-}{j}(x+2)$ is set to $1$ for some
$j$. The theorem now follows because one of the AC clauses of this
variable simplifies to the unit clause
$\{NOOP\mbox{-}clear\mbox{-}b_a(x+1)\}$. The other achievers of that
precondition are of the form $movefromt2\mbox{-}i\mbox{-}{b_a}$, which
can not be done at time step $x+1$ since time step $x$ is the first
one where $movetot2\mbox{-}{b_a}\mbox{-}{j}$ is not forced to $0$ (for
some $j$): in words, we don't have the time to stack $i$ into $b_a$ in
the first place. Now, setting $NOOP\mbox{-}clear\mbox{-}b_a(x+1)$ to
$1$ forces, by UP over the respective AC clause,
$movetot2\mbox{-}{x}\mbox{-}{b_a}(x+1)$ to be set to $0$. This
concludes the argument.
\end{proof}

\subsubsection{Asymmetrical Case}
\label{proofs:sbw:top}

We prove that $SBW_n^{n-2}B$, see its definition in
Section~\ref{sd:sbw}, and an example in Table~\ref{tab:map-bd-top}, is
a minimal backdoor in $SBW_n^{n-2}$.

\bigskip

\noindent {\bf Theorem~\ref{sbwtop}.} {\em $SBW_n^{n-2}B$ is a
  backdoor for $SBW_n^{n-2}$.}

\begin{proof}
\begin{sloppypar}
  First, observe that the claim is true for $n = 3$. There,
  $SBW_n^{n-2}B$ contains only $movetot2\mbox{-}g_1\mbox{-}t_2(1)$ and
  $movetot2\mbox{-}g_2\mbox{-}t_2(1)$. If we set one of these to $1$,
  then the move actions for the other good block, as well as for the
  single bad block, get forced to $0$ at step $1$, so they get forced
  to $1$ (by their goal constraints) at the only other time step,
  number $2$. This yields an inconsistency with the respective EC
  clause. On the other hand, if we set both
  $movetot2\mbox{-}g_1\mbox{-}t_2(1)$ and
  $movetot2\mbox{-}g_2\mbox{-}t_2(1)$ to $0$, then the NOOP variables
  in the respective goal constraints are forced to $0$ at step $2$, so
  both $move$ actions get forced to $1$ at step $2$, yielding again an
  empty EC clause.

  For the rest of the proof, we assume $n > 3$. We next make two
  observations:
\begin{itemize}
\item {\bf Observation 1.} {\em If $movetot2\mbox{-}b_{i+1}\mbox{-}b_i(i+1)$, 
   $i \geq 1$, is set to $1$, then UP sets
    $movetot2\mbox{-}b_{j+1}\mbox{-}b_j(j+1)$ to $1$ for all $0 \leq j
    < i$.}

    \medskip

   This happens by chaining over AC clauses. In time step $i$, the
   only present precondition achiever of
   $movetot2\mbox{-}b_{i+1}\mbox{-}b_i$, i.e. the only achiever of the
   precondition $abovet_2\mbox{-}b_i$, is
   $movetot2\mbox{-}b_{i}\mbox{-}b_{i-1}$. So the latter action is
   forced to $1$ at step $i$. The same argument applies iteratively
   downward to step $1$.

\item {\bf Observation 2.} {\em If $movetot2\mbox{-}b_{i+1}\mbox{-}b_i(i+1)$
    is set to $0$, then UP sets $movetot2\mbox{-}b_{j+1}\mbox{-}b_j(j+2)$
    to $1$ for all $i \leq j \leq n-3$.} 

    \medskip

  This happens as follows. With the lost precondition support of
  $movetot2\mbox{-}b_{i+1}\mbox{-}b_i(i+1)$,
  $movetot2\mbox{-}b_{i+2}\mbox{-}b_{i+1}(i+2)$ and
  $NOOP\mbox{-}abovet_2\mbox{-}b_{i+1}(i+2)$ are forced to $0$. The
  same happens iteratively upwards until
  $movetot2\mbox{-}b_{n-2}\mbox{-}b_{n-3}(n-2)$ and
  $NOOP\mbox{-}abovet_2\mbox{-}b_{n-3}(n-2)$ are set to $0$. Then,
  also $NOOP\mbox{-}abovet_2\mbox{-}b_{n-2}(n-1)$ is set to $0$. This
  means that the goal constraint for $b_{n-2}$, the upper most bad
  block, becomes unary. Thus,
  $movetot2\mbox{-}b_{n-2}\mbox{-}b_{n-3}(n-1)$ is forced to
  $1$. Since $NOOP\mbox{-}abovet_2\mbox{-}b_{n-3}(n-2)$ is already set
  to $0$, this forces $movetot2\mbox{-}b_{n-3}\mbox{-}b_{n-4}(n-2)$ to
  $1$. The same happens iteratively downwards until step $i+2$, i.e.,
  until $movetot2\mbox{-}b_{i+1}\mbox{-}b_i(i+2)$ is set to $1$.
\end{itemize}

Let $u$ be the number so that $3*2^{u-1} \leq n \leq 3*2^u$, i.e. $u$
is the $i$-index of $n$'s $BD_{top}$ equivalence class as illustrated
in Table~\ref{tab:map-bd-top} ($u \geq 1$ since $n > 3$). Note that $u
= \lceil log_2(n/3) \rceil$. Keep in mind that $3*2^u$ is the largest
$n$ in the equivalence class, and that this largest $n$ has a CNF with
$n-1$ layers, numbered $1, \dots, n-1 = 3*2^u-1$. We prove below that:
\begin{itemize}
\item {\bf Observation 3.} {\em For all $1 \leq i \leq u$,
$movetot2\mbox{-}b_{3*2^{i-1}-1}\mbox{-}b_{3*2^{i-1}-2}(3*2^{i-1}-1)$
  must be set to $1$ or else UP yields a contradiction.}
\end{itemize}

This suffices because setting
$movetot2\mbox{-}b_{3*2^{u-1}-1}\mbox{-}b_{3*2^{u-1}-2}(3*2^{u-1}-1)$
to $1$ yields a contradiction by UP.  Let us first show the
latter. With observation 1, below $3*2^{u-1}-1$ every time step has a
$movetot2$ set to $1$, so all other $movetot2$ actions are out, in
particular $movetot2\mbox{-}g_1\mbox{-}t_2$ and
$movetot2\mbox{-}g_2\mbox{-}t_2$. To re-insert these actions, i.e., to
reach a time step where they are not set to $0$ by UP, one needs
$3*2^{u-1}-1$ steps: by chaining over AC clauses, UP finds that all of
the $b_j$ that are already above $t_2$ (i.e., $1 \leq j \leq
3*2^{u-1}-1$) have to be moved back to $t_1$ again.  So
$movetot2\mbox{-}g_1\mbox{-}t_2$ and $movetot2\mbox{-}g_2\mbox{-}t_2$
re-appear first in time step $3*2^{u-1} + 3*2^{u-1}-1 =
3*2^{u}-1$. But either there aren't that many time steps in the CNF,
namely if $n < 3*2^u$, yielding the $g_1$ and $g_2$ goal constraints
empty; or we get a contradiction in time step $3*2^{u}-1$ since it is
the only unoccupied (by a movetot2 action) time step left, and both
$movetot2\mbox{-}g_1\mbox{-}t_2$ and $movetot2\mbox{-}g_2\mbox{-}t_2$
are forced to $1$ by their respective goal constraint.

We now prove observation 3 by induction over $i$. Base case, $i = 1$,
$movetot2\mbox{-}b_2\mbox{-}b_1(2)$. If we set that to $0$, then, by
observation 2, all $movetot2\mbox{-}b_{j+1}\mbox{-}b_j(j+2)$ are
forced to $1$ for $1 \leq j \leq n-3$.  This means, in particular,
that all layers $t \geq 3$ are occupied by $movetot2$ actions, and
only layers $1$ and $2$ are left to achieve the goals for $g_1$ and
$g_2$. This reduces to exactly the same situation as with $n=3$,
discussed at the start of the proof: the goal constraints for $g_1$
and $g_2$ get transported down to time step $2$, and the bad block
$b_1$ must be moved either in step $1$ or in step $2$ due to the AC
clause of $movetot2\mbox{-}b_2\mbox{-}b_1(3)$, which is set to $1$.

Inductive case, $i \rightarrow i+1$. We assume that
$movetot2\mbox{-}b_{3*2^{i-1}-1}\mbox{-}b_{3*2^{i-1}-2}(3*2^{i-1}-1)$
is set to $1$. By observation 1 we get that
$movetot2\mbox{-}b_{j+1}\mbox{-}b_j(j+1)$ is set to $1$ for all $0
\leq j < 3*2^{i-1}-2$. So $movetot2\mbox{-}g_1\mbox{-}t_2$ and
$movetot2\mbox{-}g_2\mbox{-}t_2$ are forced to $0$ at all time steps
$t \leq 3*2^{i-1}-1$. Say we set
$movetot2\mbox{-}b_{3*2^{i}-1}\mbox{-}b_{3*2^{i}-2}(3*2^{i}-1)$ to
$0$.  Then, by observation 2, $movetot2$ actions are forced to $1$ at
all time steps $t > 3*2^{i}-1$. So $movetot2\mbox{-}g_1\mbox{-}t_2$
and $movetot2\mbox{-}g_2\mbox{-}t_2$ are also forced to $0$ for all
time steps $t > 3*2^{i}-1$. After applying
$movetot2\mbox{-}b_{3*2^{i-1}-1}\mbox{-}b_{3*2^{i-1}-2}(3*2^{i-1}-1)$,
i.e., starting from time step $3*2^{i-1}$, the number of steps we need
to make the preconditions of $movetot2\mbox{-}g_{1}\mbox{-}t_2$ and
$movetot2\mbox{-}g_{2}\mbox{-}t_2$ achievable is $3*2^{i-1}-1$: we
must remove all the already moved bad blocks from $t_2$ again (this is
basically the same argument as used further above). So
$movetot2\mbox{-}g_1\mbox{-}t_2$ and $movetot2\mbox{-}g_2\mbox{-}t_2$
first re-appear in time step $3*2^{i-1} + 3*2^{i-1}-1 =
3*2^{i}-1$. But, as observed above, all time steps $t' > 3*2^{i}-1$
are already occupied, by observation 2, and since we decided to set
$movetot2\mbox{-}b_{3*2^{i}-1}\mbox{-}b_{3*2^{i}-2}(3*2^{i}-1)$ to
$0$. So both $movetot2\mbox{-}g_1\mbox{-}t_2$ and
$movetot2\mbox{-}g_2\mbox{-}t_2$ get forced to $1$ at time step
$3*2^{i}-1$ due to their goal constraints, which are transported down
to that time step. We get an empty EC clause. This concludes the
argument.
\end{sloppypar}
\end{proof}

The reader who is confused by all the $(3*2^{x}-y)$ indexing in the
proof to Theorem~\ref{sbwtop} is advised to have a look at
Table~\ref{tab:map-bd-top}, and instantiate the indices in the proof
with the numbers $1$ for $3*2^{1-1}-2$, $4$ for $3*2^{2-1}-2$, and
$10$ for $3*2^{3-1}-2$.

\bigskip

\noindent {\bf Theorem~\ref{sbwtop:upcons}} {\em Let $B'$ be a subset
  of $SBW_n^{n-2}B$ obtained by removing one variable.  Then there is
  exactly one UP-consistent assignment to the variables in $B'$.}

\begin{sloppypar}
\begin{proof}
  The claim follows, with some additional thoughts, from the proof to
  Theorem~\ref{sbwtop}. We use the notations from that proof. We
  observe the following:
\begin{enumerate}
\item If a $SBW_n^{n-2}B$ variable $v$ at time $t$ is set to $1$,
  then all the $SBW_n^{n-2}B$ variables at times $t' < t$ are set to
  $1$ by UP. This follows from observation 1 in the proof to
  Theorem~\ref{sbwtop}.
\item If the topmost $SBW_n^{n-2}B$ variable (the variable at 
   the latest time step) is set to $1$, then we get an empty clause in
  UP. This is shown below observation 3 in the proof to
  Theorem~\ref{sbwtop}.
\item If a $SBW_n^{n-2}B$ variable $v$ at time $t$ is set to $0$,
  then all the $SBW_n^{n-2}B$ variables at times $t' > t$ are set to
  $0$ by UP. This follows from observation 2 in the proof to
  Theorem~\ref{sbwtop}.
\item If the lower most variable regarding a bad block in $SBW_n^{n-2}B$ 
   (the variable at time step $2$) is set to $0$, then we get an empty
   clause in UP. This corresponds to the base case in the induction in
   the proof to Theorem~\ref{sbwtop}.
\item  If we have two consecutive $SBW_n^{n-2}B$ variables $v$ and $v'$,
  i.e. $v =
  movetot2\mbox{-}b_{3*2^{i-1}-1}\mbox{-}b_{3*2^{i-1}-2}(3*2^{i-1}-1)$
  for some $1 \leq i < \lceil log_2(n/3) \rceil$, and $v' =
  movetot2\mbox{-}b_{3*2^{i}-1}\mbox{-}b_{3*2^{i}-2}(3*2^{i}-1)$, and
  $v$ is set to $1$ and $v'$ is set to $0$, then we get an empty
  clause during UP. This is because there are not enough time steps
  between $3*2^{i-1}-1$ and $3*2^{i}-1$, to move all the bad blocks
  back to $t_1$, and to accommodate the necessary moves for the two
  good blocks. This corresponds to the inductive case in the induction
  in the proof to Theorem~\ref{sbwtop}.
\end{enumerate}
We now prove the claim with a case distinction over the type of the
left out variable $v$.

{\bf Case 1, $v =
movetot2\mbox{-}b_{3*2^{i-1}-1}\mbox{-}b_{3*2^{i-1}-2}(3*2^{i-1}-1)$,
for some $1 < i \leq \lceil log_2(n/3) \rceil$.} That is, we leave out
one of the upper moves regarding bad blocks. With the observations
above, the only chance we have to avoid an empty clause in UP is to
set all $SBW_n^{n-2}B$ variables at time steps $t > 3*2^{i-1}-1$ (all
moves of bad blocks above the left-out variable) to $0$, and all
$SBW_n^{n-2}B$ variables at time steps $2 \leq t < 3*2^{i-1}-1$ (all
moves of bad blocks below the left-out variable) to $1$. The
$SBW_n^{n-2}B$ variables at time step $1$, i.e., the moves of the two
good blocks, are set to $0$ by UP. Since between step $3*2^{i-2}-1$
and step $3*2^{i}-1$ there is enough time to accommodate the moves for
the good blocks, we don't get an empty clause in UP.

{\bf Case 2, $v = movetot2\mbox{-}b_2\mbox{-}b_1(2)$.} With the
observations above, we have to set all $SBW_n^{n-2}B$ variables at
time steps $t > 2$ (all moves of bad blocks above the left-out
variable) to $0$. Then all time steps are occupied except steps $1,
\dots, 5$. If we set one of the $SBW_n^{n-2}B$ variables at step $1$
to $1$, then we get an empty clause because both the move for the
remaining good block, and the move for the lower most bad block, are
forced to $1$ in time step $2$. However, if we set both $SBW_n^{n-2}B$
variables at step $1$ to $0$, then no empty clause arises -- the same
moves are still available (not forced to $0$) in all steps $2, \dots,
5$.

{\bf Case 3, $v = movetot2\mbox{-}g_i\mbox{-}t_2(1)$, $i \in \{1,
2\}$.} That is, $v$ is one of the two moves regarding good
blocks. With the observations above, we have to set all $SBW_n^{n-2}B$
variables at time steps $t > 1$ (all moves of bad blocks) to $0$. So
all time steps are occupied except steps $1$ and $2$. If we set
$movetot2\mbox{-}g_{i'}\mbox{-}t_2(1)$, $i' \neq i$ (the variable not
left out at step $1$) to $1$, then all other moves are forced out at
that step, in particular the move $movetot2\mbox{-}g_i\mbox{-}t_2(1)$,
i.e. the left-out variable. But then all $SBW_n^{n-2}B$ variables are
instantiated, and we get an empty clause because of
Theorem~\ref{sbwtop}. If we set $movetot2\mbox{-}g_{i'}\mbox{-}t_2(1)$
to $0$, however, the only thing that happens is that
$movetot2\mbox{-}g_{i}\mbox{-}g_{i'}$ is forced to $0$ at step $2$. In
particular, the goal constraints for both $g_i$ and $g_{i'}$ are still
binary in step $2$: for $g_{i'}$, the constraint is
$\{movetot2\mbox{-}g_{i'}\mbox{-}g_{i}(2),$
$movetot2\mbox{-}g_{i'}\mbox{-}t_2(2)\}$; for $g_{i}$, the constraint
is $\{movetot2\mbox{-}g_{i}\mbox{-}t_2(2),$
$NOOP\mbox{-}abovet_2\mbox{-}g_{i}(2)\}$.  This concludes the
argument.
\end{proof}
\end{sloppypar}

%% file: SPH-proofs.tex
\subsection{SPH  Backdoors}
\label{proofs:sph}

As stated in Section~\ref{sd:sph}, the backdoor we identify for
$SPH_n^k$, denoted by $SPH_n^kB$, is defined as follows (compare
Figure~\ref{XFIG/sph-BD.eps}):
 {\small
\medskip
\begin{tabbing}
$SPH_n^kB$ := \= $\{0\_y \mid 1 \leq y \leq n-1\} \; \; \cup$\\
\> $\{x\_y \mid k+2 \leq x \leq n, 1 \leq y \leq n-1\}$
\end{tabbing}}

\medskip
\noindent {\bf Theorem~\ref{sph}} {\em $SPH_n^kB$ is a backdoor in
  $SPH_n^k$.}

\medskip

\begin{proof}
  We prove a more general claim where the subsets of holes and normal
  pigeons underlying the backdoor are selected arbitrarily, of the
  appropriate size. Denote by $B$ any var set such that there exist $X
  \subseteq \{k, \dots, n\}$, and $Y \subseteq \{1, \dots, n\}$, $|X|
  = n-k-1$, $|Y| = n-1$, with $B = \{x\_y \mid x \in X \cup \{0\}, y
  \in Y\}$.  In words, $B$ talks about $n-1$ (i.e. all but one of the)
  holes, for the bad pigeon, and for $n-k-1$ (i.e. all but two of the)
  normal pigeons. Obviously, $SPH_n^kB$ can be constructed this way.
  
  \begin{sloppypar} Before we prove that $B$ is a backdoor, observe
  that the naive encoding of $GEQ(\{0\_1, \dots, 0\_n\}, k)$ has the
  following efficiency property. For any subset $Y'$, $|Y'| = n-k$, of
  holes, if $0\_y'$ is set to $0$ for all $y' \in Y'$, then $0\_y$ is
  set to 1 by UP for all $y \not \in Y'$.  The reason is that, in the
  respective clause $\{0\_y' \mid y' \in Y' \cup \{y\}\}$ requiring
  holes for the bad pigeon, only the single open literal $0\_y$ is
  left. If $|Y'| > n-k$, then one of these clauses is contained in
  $Y'$ and becomes empty.  \end{sloppypar}
  
  Now, let $a$ be any value assignment to the variables in $B$.
  Consider the $n-k-1$ pigeons $x$ in $X$. For all these $x$, $B$
  talks about all but one hole, so there is at least one hole $y$ so
  that $x\_y = 1$ after UP. Since no two $x$ can occupy the same hole,
  it follows that, when all $X$ variables are set, at least $n-k-1$
  $0\_y'$ variables are set to $0$ after UP.  If more than $n-k$
  $0\_y'$ variables are set to $0$, we get an inconsistency with the
  efficiency property observed above. If exactly $n-k$ $0\_y'$
  variables are set to $0$, then with the efficiency property the
  remaining $k$ $0\_y$ variables are set to $1$. The holes are then
  all occupied and the two normal pigeons not in $X$ yield a
  contradiction. So we can assume that the $n-k-1$ pigeons in $X$
  occupy only one hole each.  The respective $n-k-1$ $0\_y'$ variables
  are set to $0$ after UP.  Consider the remaining unset $0\_y$
  variables in $B$. If a single one of them was set to $0$, with what
  was stated above all others would be set to $1$.  This would yield
  $n-1$ occupied holes, and in effect by UP we would get a
  contradiction with the two normal pigeons not in $X$. So $a$ must
  set all remaining unset $0\_y$ variables in $B$ to $1$.  But there
  are at least $k$ of these unset $0\_y$ variables in $B$: if all
  holes occupied by the $X$ pigeons are in $Y$, then exactly $k$
  ($=n-1-(n-k-1)$) variables $0\_y \in B$ are unset; if one of the
  holes occupied by the $X$ pigeons is the single hole not in $Y$,
  then $k+1$ variables $0\_y \in B$ are unset. So, when $a$ sets all
  unset $0\_y$ variables in $B$ to $1$, at least $n-k-1+k=n-1$ holes
  are occupied, and again we get an inconsistency with the two normal
  pigeons not in $X$. This concludes the argument.
\end{proof}

\medskip

\noindent {\bf Theorem~\ref{sph:upcons}.} {\em Let $B'$ be a subset of
  $SPH_n^kB$ obtained by removing one variable $x\_y$. Then, if $x
  \neq 0$, to the variables in $B'$ there are exactly $(n-2) * \dots *
  (k+1)$ UP-consistent assignments; if $x=0$ then there are exactly
  $(n-2) * \dots * k$ UP-consistent assignments.}

\begin{proof}
  As in the proof to Theorem~\ref{sph}, denote by $X$ the normal
  pigeons underlying the backdoor, and by $Y$ the holes underlying the
  backdoor. Observe that the UP-consistent assignments are exactly
  those where the $n-k-1$ pigeons $(X \cup \{0\}) \setminus \{x\}$
  occupy all $n-1$ holes $Y \setminus \{y\}$. If an assignment does
  have this property, then it is UP-consistent because both the holes
  $y$ and $n$ remain empty, with the three pigeons $x$, $k$, and $k+1$
  being un-assigned. If an assignment does not have this property,
  then the variable $x\_y$ is assigned a value by UP, which yields an
  empty clause due to Theorem~\ref{sph}. From here, the claim follows
  with some simple counting arguments, i.e. by counting how many such
  assignments there are. If $x \neq 0$, then the $n-k-2$ pigeons in $X
  \setminus \{x\}$ can be distributed freely over the $n-2$ holes $Y
  \setminus \{y\}$, giving us $(n-2) * \dots * (k+1)$
  possibilities. After that, $k$ holes (in $Y$) are left open to
  accommodate the $k$ holes needed for the bad pigeon, adding no more
  possibilities.

  If $x=0$, then the $n-k-1$ pigeons in $X$ can be distributed freely
  over the $n-2$ holes $Y \setminus \{y\}$, giving us $(n-2) * \dots *
  k$ possibilities. After that, $k-1$ holes (in $Y$) are left open to
  accommodate the $k-1$ holes needed to be set to $1$ for the bad
  pigeon (if less than $k-1$ of the present variables are set to $1$,
  we get a contradiction by UP), giving us no more possibilities.
\end{proof}

%% file: cutsets.tex
\section{Cutsets}
\label{cutsets}

We discuss cutsets in our synthetic domains, showing that there are
large lower bounds in all cases. We consider MAP, SBW, and SPH in
turn. The different kinds of subsets are defined and explained in the
MAP section; the later sections use the same terminology.

\subsection{MAP}
\label{cutsets:map}

The notion of cutsets is defined in
\cite{dechter:ai-90,rish:dechter:jar-00,dechter03}. The simplest form
of a cutset is a {\em cycle-cutset}, defined as follows. For a CNF
formula $\phi$ with variable set $V$ and clause set $C$, the {\em
  constraint graph} of $\phi$ is the undirected graph $(V,E)$, where
$\{v,v'\} \in E$ iff $\exists c \in C: v, v' \in c$. A cycle-cutset is
a subset of nodes $V'$ in the graph so that, when removing $V'$ and
the incident edges from $(V,E)$, the resulting graph is cycle-free.

It is easy to see that there are no small cycle-cutsets in $MAP_n^k$,
independently of $k$. This is due to the many {\em cliques} of nodes
in the constraint graph. At each time step $1 \leq t
\leq 2n-2$, we have large cliques of variables, linear in $n$, due to
EC clauses between $move$ actions. At each time step $2 \leq t
\leq 2n-2$, we have large cliques of variables, linear in $n$, due to
the AC clauses of various $move$ and $NOOP$ actions. At time step $t =
2n-2$, for each branch $1 \leq i \leq n$, we have at most one edge
between a $NOOP\mbox{-}visited$ action, and a $move$ action. The
latter are the only edges that change, over $k$. For example, at any
$1 \leq t \leq 2n-2$, the variable set
$\{move\mbox{-}L^0\mbox{-}L^1_i(t),$
$move\mbox{-}L^1_i\mbox{-}L^0(t),$ $move\mbox{-}L^1_1\mbox{-}L^2_1(t),
\dots,$ $move\mbox{-}L^{t-1}_1\mbox{-}L^t_1(t) \mid 1 \leq i \leq n\}$
forms a clique due to EC clauses. Also, for any $2 \leq t \leq 2n-2$
and $1 \leq i \leq n$, the variable set
$\{move\mbox{-}L^0\mbox{-}L^1_i(t),$ $NOOP\mbox{-}at\mbox{-}L^0(t-1),$
$move\mbox{-}L^1_j\mbox{-}L^0(t-1) \mid 1 \leq j \leq n\}$ forms a
clique due to the AC clause for
$move\mbox{-}L^0\mbox{-}L^1_i(t)$. Obviously, for a clique of size
$m$, the smallest cycle-cutset has size $m-2$. Observe that the first
kind of example cliques are disjoint, and that the second kind of
example cliques are disjoint if we choose only one $i$. So, just from
these example cliques, we get lower bounds of
\[
\sum_{t=1}^{2n-2} [(2n + t - 1)-2] = 6n^2 - 13n + 7
\]
and
\[
\sum_{t=2}^{2n-2} [(n+2)-2] = 2n^2 - 3n  
\]
on the cycle-cutset size, independently of $k$. One can derive several
similar lower bounds in a similar manner. Remember that the total
number of variables is $16n^2 - 33n + 14$, i.e., also a square
function in $n$.

A refinement of cycle-cutsets are {\em $b$-cutsets}. These are
variable subsets so that, once removed from the graph with their
incident edges, the resulting graph has an {\em induced width} of at
most $b$. In a nutshell, this means that there is an ordering of the
graph nodes so that, for every node $v$, there are at most $b$ {\em
parents}: nodes $v'$ that are ordered before $b$, and that are
connected to $v$ by an edge.\footnote{In fact, the nodes $v$ are
processed in reverse order, and, every time a node is processed, all
its parents are being connected by, possibly, additional edges. This
can lead to a higher width encountered at later points during the
process.} If $b=1$ then the $b$-cutset is a cycle-cutset, because only
trees have induced width $1$. Now, in any clique of size $m$, the
induced width is $m-1$. To reduce it to $1 \leq b \leq m-1$, $m-1-b$
nodes must be removed. With the above, this shows that $b$-cutsets
aren't suitable either to capture what changes on the $k$ scale: we
don't gain much in the trade-off between $b$ and cutset size. In more
detail, since solution algorithms are exponential in $b$, normally one
wants to set $b$ as a constant. For any constant $b$, with the above
the $b$-cutset size is still a square function in $n$, irrespective of
$k$. Even if we set $b$ to a linear function in $n$ only a constant
number smaller than $n$ (the size of the second kind of cliques above,
minus $2$), irrespectively of $k$ we still get a linear-size
$b$-cutset.

Matters can change if we move to what one could refer to as {\em
conditional} cutsets. These can be defined based on the definition of
conditional constraint graphs \cite{dechter03}. A conditional
constraint graph depends on the cutset variables $V'$ {\em and a value
assignment $a$ to $V'$}. The graph results from the constraint graph
by removing the cutset variables $V'$ and their edges, {\em plus}
removing all edges corresponding to clauses that are satisfied by
$a$. For example, if $A$ is a cutset variable and there is a clause
$\{A, B, C\}$, then the edge $\{B, C\}$ is removed under the
assignment $A \mapsto 1$, but not under the assignment $A \mapsto
0$. Say we further strengthen this by saying that the conditional
constraint graph results from the constraint graph by executing all
unit propagations, i.e., we switch to a notion of ``conditional
cutsets with unit propagation''. Then, obviously, the cutsets we get
are at least as small as the backdoors we investigate here. For every
value assignment, UP will report that the conditional constraint graph
has collapsed, so any backdoor for unit propagation will also be a
cycle-cutset, in that sense. However, with this definition, both
cutsets and backdoors make use of UP, and the small cutsets for large
values of $k$ are caused by the same phenomenon as the small
backdoors. It is a topic for future work to investigate whether there
exist even smaller cutsets than backdoors, in that context. Note that
conditional cutsets, just like our backdoors here, can not be tested
statically, i.e., one can not test whether $V'$ is such a cutset
without enumerating the value assignments.

It is important to note that conditional cutsets {\em without} UP are
still not strong enough to capture the behavior of the $MAP_n^k$
formulas. Consider, for example, the above first kind of example
cliques. Every edge $\{v, v'\}$ in such a clique comes from a clause
$\{\neg v, \neg v'\}$, so the clique edges have a 1-to-1
correspondence to clauses. If we set a variable $v$ in the clique to
$1$, then {\em no} clause (edge) in the clique becomes satisfied. If
we set $v$ to $0$, then the clique clauses that become satisfied
correspond precisely to the clique edges that would be removed anyway
since they are adjacent to $v$. So the above stated first lower bound,
$6n^2 - 13n + 7$, is still valid for conditional cycle-cutsets
(without UP), independently of $k$.



\subsection{SBW}
\label{cutsets:SBW}

We can derive lower bounds on the size of cutsets from the cliques in
the constraint graph formed by the pairwise incompatible $move$
actions at each time step. What we have to think about is how many
$move$ actions are present in a time step $t$, depending on $n$, $k$,
and $t$. We do not spell this out in detail (which would involve
several case distinctions), but only derive a simple lower bound as
follows:
\begin{itemize}
\item In every time step $2 \leq t \leq k$, we have $(n-k)*(n-k)$ actions
that $move$ a good block above $t_2$ (move any good block onto any
other good block or onto $t_2$). We further have at least $t$ actions
that $move$ a bad block above $t_2$ (move $b_1$ onto $t_2$, move any
$b_i$, $i \leq t$, onto $b_{i-1}$). Note that there is no such time
step $t$ if $k < 2$.

\item In every time step $k+2 \leq t \leq n-1$, we have $(n-k)*(n-k)$ actions
that $move$ a good block above $t_2$, and we have at least $k$ actions
that $move$ a bad block above $t_2$. Note that we start at time step
$2$ if $k=0$ (so we have all the $(n-k)*(n-k)$ actions for the good
blocks).
\end{itemize}
At each $t$, the listed $move$ variables form a clique in the
constraint graph due to the EC clauses. Further, for distinct $t$,
clearly the cliques are disjoint. So a cycle-cutset must remove $m-2$
variables from each of these cliques of size $m$. We get a lower bound
of:
\[
\sum_{t=2}^{k} [(n-k)^2 + t - 2]\; \; +
\]
\[
\sum_{t=k+2}^{n-1} [(n-k)^2 + k - 2]
\]
variables to be removed. Inserting $k=0$, we get $\sum_{t=2}^{n-1}
[n^2 - 2] = n^3 - 2n^2 - 2n + 4$. Inserting $k=n-2$, we get
$\sum_{t=2}^{n-2} [4 + t - 2] = \frac{1}{2}n^2 + \frac{1}{2}n -
6$. So, at both extreme ends of the $k$ scale, the obtained lower
bound on cycle-cutset size is a linear function in the total number of
variables. Note that these lower bounds are, actually, quite generous
since we haven't considered, for example, the actions that move a
block back onto $t_1$.

Regarding $b$-cutsets, remember that, from any clique of size $m$,
$m-1-b$ nodes must be removed in order to reduce the induced width to
$1 \leq b \leq m-1$. With $k=0$, each of the $n-2$ disjoint cliques we
identified has $n^2$ nodes. So, for any constant $b$, the $b$-cutset
size is in $\Omega(n^3)$. The $b$-cutset size is linear in $n$ even if
we set $b$ to a square function in $n$ only a constant number smaller
than the clique size. With $k=n-2$, matters are a little more
complicated since the identified disjoint cliques grow with the time
step. If we set $b$ to a constant, then the $b$-cutset size is a
square function in $n$. If we set $b$ to be a large enough linear
function in $n$, only a constant number smaller than the largest
identified clique, then the lower bound becomes constant. Still, due
to the need for a $b$ that scales linearly with $n$, this does not
indicate that the $b$-cutsets are suitable to identify the particular
structure of the $SBW_n^{n-2}$ formulas.

Regarding conditional cutsets, as in MAP we observe that, without the
use of unit propagation in the definition of the conditional
constraint graph, the lower bounds do no change at all. The reason is,
as in MAP, that no setting of the clique variables satisfies/removes
any clique clauses/edges that would not be removed anyway. If we do
use UP in the definition of the conditional constraint graph, then, as
before, the cutsets trivially become as small as the backdoors. It is
an open question whether there exist even smaller cutsets in this
context.


\subsection{SPH}
\label{cutsets:sph}

As before we can identify large cliques of variables, that prevent the
existence of small cutsets, except when using unit propagation in the
definition of the conditional constraint graph. Precisely, in the
constraint graph for $SPH_n^k$, the cliques we get are the following:
\begin{enumerate}
\item $\{x\_1, \dots, x\_n\}$ for all pigeons $0 \leq x \leq n$, due 
to the clause(s) requiring $x$ to be put into a (sufficient number of)
hole(s). Note that, in the constraint graph, in this respect there is
no difference between the bad pigeon and the other pigeons, i.e.,
irrespective of $k$, the clique we get for $x=0$ is the same.
\item $\{1\_y, \dots, n\_y\}$ for all holes $1 \leq y \leq n$, due 
to the clauses requiring that in each hole there is at most one good
or normal pigeon.
\item $\{0\_y, k\_y, \dots, n\_y\}$ for all holes $1 \leq y \leq n$, due 
to the clauses requiring that in each hole there is at most one bad or
normal pigeon.
\end{enumerate}
It is easy to see that these three kinds of cliques actually form {\em
all} edges in the constraint graph. Observe that the only cliques
whose size or number depends on $k$ is kind 3. From kind 2, we get the
desired lower bounds on various kinds of cutsets. These cliques are
all disjoint; they have size $n$, and there are $n$ of them. This
implies that, irrespectively of $k$, any cycle-cutset has at least
size $(n-2)*n = n^2 -2n$, i.e., just $3n$ variables less than the
entire number of variables in the formula. Likewise, for $b$-cutsets
with constant $b$, we get a lower bound square in $n$. Similarly as in
MAP and SBW, conditional cutsets without unit propagation are not any
smaller, because no setting of the clique (kind 2) variables
satisfies/removes any clique clauses/edges that would not be removed
anyway. Conditional cutsets {\em with} unit propagation are just as
small as the backdoors, and it is an open question whether they can be
even smaller.
